%% file: main.tex
\documentclass{article}

\input{sections/packages.tex}
\input{sections/macros.tex}

\title{\input{sections/title.tex}}

\author[1]{Tam\'as P. Papp\thanks{\href{mailto:t.papp@lancaster.ac.uk}{\texttt{t.papp@lancaster.ac.uk}}}}
\author[2]{Chris Sherlock}
\affil[1]{\small{STOR-i Centre for Doctoral Training, Lancaster University, UK}}
\affil[2]{\small{School of Mathematical Sciences, Lancaster University, UK}}
\date{}

\begin{document}

\maketitle 

\begin{abstract}
   \input{sections/abstract.tex}
\end{abstract}

\emph{Keywords:} \input{sections/keywords.tex}

\begin{bibunit}

\input{sections/1-intro.tex}
\input{sections/2-empirical.tex}

\input{sections/3-debias.tex}
\input{sections/4-mcmc-bias.tex}
\input{sections/5-mcmc-convergence}
\input{sections/6-discussion.tex}

\section*{Acknowledgements} 
\input{sections/acknowledgements.tex}

\putbib
\end{bibunit}

\appendix

\begin{bibunit}

\input{sections/appendix.tex}

\putbib
\end{bibunit}

\end{document}

%% file: sections/packages.tex
\usepackage[utf8]{inputenc}
\usepackage{graphicx}

\usepackage[a4paper, margin = 2.5cm]{geometry}

\usepackage{amsmath}
    \allowdisplaybreaks
    
\usepackage{amsthm}
        
        \newtheorem{theorem}{Theorem}
        \newtheorem{proposition}{Proposition}
        \newtheorem{lemma}{Lemma}
        \newtheorem{corollary}{Corollary}
    \theoremstyle{remark}
        \newtheorem{remark}{Remark}
        \newtheorem*{remark*}{Remark}
 \theoremstyle{definition}
        \newtheorem{definition}{Definition}
        \newtheorem{example}{Example}
\usepackage{amssymb}
\usepackage{amsfonts}
\usepackage{bbm}
\usepackage{mathtools}

\usepackage{circledsteps}
\usepackage[ruled,vlined,noalgohanging]{algorithm2e}

\usepackage[shortlabels]{enumitem}
    \setlist[enumerate]{itemsep=0.5ex}
    \setlist[itemize]{itemsep=0.5ex}

\PassOptionsToPackage{rgb}{xcolor}
\usepackage{xcolor}
    \definecolor{wongblue}{HTML}{0072B2}
    \definecolor{wongorange}{HTML}{D55E00}
    
\usepackage{hyperref}
    \hypersetup{
        pdfencoding=auto,
        colorlinks=true,
        citecolor=wongblue,
        linkcolor=wongorange,
        urlcolor=wongorange
    }

\usepackage[round]{natbib}
\usepackage{bibunits}
    \defaultbibliographystyle{abbrvnat}
    \defaultbibliography{wasserstein}

\usepackage{authblk}

\usepackage{cancel}

%% file: sections/macros.tex
    \DeclareMathOperator{\sgn}{sgn} 
    \DeclareMathOperator{\var}{Var} 
    \DeclareMathOperator{\cov}{Cov} 
    \DeclareMathOperator{\unif}{Unif}
    \DeclareMathOperator{\invgamma}{InvGamma}
    
    \DeclareMathOperator{\iid}{\overset{\smash{\textup{iid}}}{\sim}}
\DeclareMathOperator{\wass}{\mathcal{W}} 
\DeclareMathOperator{\kl}{KL} 
\DeclareMathOperator{\tr}{{Tr}} 
\DeclareMathOperator{\gtrunidisp}{\ge_{disp}} 
\DeclareMathOperator{\gtrcvx}{\ge_{cvx}} 
\DeclareMathOperator{\diag}{diag}
\DeclareMathOperator{\weakly}{\Rightarrow}    
\newcommand{\lip}[1]{\|#1\|_\textsc{\textup{Lip}}}

\newcommand{\sgnpow}[2]{\left[ #2 \right]_\pm^{#1}}

\DeclareMathOperator{\cotr}{\overset{\smash{\scriptscriptstyle{\textsc{cot}}}}{\rightsquigarrow}}

\DeclareMathOperator{\pcar}{\overset{\smash{\scriptscriptstyle{\textsc{pca}}}}{\rightsquigarrow}}

\newcommand{\X}[1]{X^{\smash{(#1)}}}
\newcommand{\Xbar}[1]{\bar{X}^{\smash{(#1)}}}

\newcommand{\Z}[1]{Z^{\smash{(#1)}}}
\newcommand{\M}[1]{M^{\smash{(#1)}}}
\newcommand{\Sig}[1]{\Sigma^{\smash{(#1)}}}
\newcommand{\Mu}[1]{\mu^{\smash{(#1)}}}
\newcommand{\marg}[1]{\pi^{\smash{(#1)}}}
\newcommand{\margbar}[1]{\bar\pi^{\smash{(#1)}}}

%% file: sections/title.tex
Centered plug-in estimation of Wasserstein distances

%% file: sections/abstract.tex
The plug-in estimator of the squared Euclidean 2-Wasserstein distance is conservative, however due to its large positive bias it is often uninformative. We eliminate most of this bias using a simple centering procedure based on linear combinations. We construct a pair of centered plug-in estimators that decrease with the true Wasserstein distance, and are therefore guaranteed to be informative, for any finite sample size. Crucially, we demonstrate that these estimators can often be viewed as complementary upper and lower bounds on the squared Wasserstein distance. Finally, we apply the estimators to Bayesian computation, developing methods for estimating (i) the bias of approximate inference methods and (ii) the convergence of MCMC algorithms.

%% file: sections/keywords.tex
Wasserstein distance, plug-in estimation, approximate inference, Markov chain Monte Carlo, optimal transport

%% file: sections/1-intro.tex
\section{Introduction} \label{sec:intro}

Wasserstein distances are a class of probability metrics rooted in the theory of optimal transport \citep{villani2003topics, villani2009optimal} that increasingly underpin methodological developments in statistics \citep{panaretos2019statistical} and machine learning \citep{peyre2019computational}. 

We are motivated by two important problems from Bayesian computation: (i) assessing the quality of approximate inference methods, and (ii) assessing the convergence of Markov chain Monte Carlo (MCMC) algorithms to their limiting distributions. 
The former is one of the present \emph{grand challenges} in Bayesian computation \citep{bhattacharya2024grand}, whereas the latter has been challenging practitioners for over thirty years \citep{gelman1992inference}. 
Assessing the accuracy in terms of Wasserstein distances is particularly appealing in these contexts, because bounds on Wasserstein distances guarantee the accuracy of various downstream inferential tasks \citep{huggins2020validated}. 
At the same time, because we want to recognize when an approximation is accurate, one key requirement for Wasserstein distance estimators in these contexts is that they decrease with the Wasserstein distance itself.

Standard plug-in estimators of the Wasserstein distance have substantial positive biases that are particularly apparent when the Wasserstein distance is small. Furthermore, due to fundamental statistical challenges related to estimating Wasserstein distances \citep{hutter2021minimax}, this bias can often not be meaningfully reduced by increasing the sample size. To obtain informative estimators of the Wasserstein distance, we must therefore resolve the issue of bias by a different approach.

We eliminate most of the bias using a simple centering procedure based on linear combinations. 
Because this centering ensures that the bias decreases with the true Wasserstein distance for any finite sample size, it allows us to circumvent statistical challenges and obtain informative estimates, at moderate sample sizes, even in high dimensions. 
In a nutshell, we construct a pair of complementary estimators: 
$U$, which is often an approximate \emph{upper bound} on the squared Wasserstein distance, and
$L$, which is always an approximate \emph{lower bound}. 
Formal sufficient conditions for $U$ to be conservative may be interpreted as a form of overdispersion between the two distributions, which aligns naturally with our motivating problems from Bayesian computation.

The paper is organized as follows. 
In Section~\ref{sec:empirical} we review key aspects of Wasserstein distances and their estimation. 
In Section~\ref{sec:main-results} we introduce the new centered estimators, analyze their finite-sample statistical properties, and discuss how to efficiently quantify their uncertainties. 
In Section~\ref{sec:bias} we develop a methodology for assessing the quality of approximate inference methods, and in Section~\ref{sec:convergence} we develop a methodology for assessing the convergence of MCMC algorithms; both of these are based on post-processing the output of multiple replicate Markov chains using the centered estimators. 
We summarize our findings and outline directions for further research in Section~\ref{sec:discussion}. 
R \citep{r} code is available on \href{https://github.com/tamaspapp/wassersteinbound}{GitHub}.

%% file: sections/2-empirical.tex
\section{Plug-in estimation of Wasserstein distances} \label{sec:empirical}

We review here selected aspects of Wasserstein distances and their estimation. We refer the reader to the works \cite{villani2009optimal,peyre2019computational,panaretos2019statistical} for further theoretical, computational, and statistical details, respectively.

Let $(\mathcal{X},c)$ be a metric space and let $\mu, \nu \in \mathcal{P}(\mathcal{X})$ be probability distributions on $\mathcal{X}$. The $p$-Wasserstein distance is defined, through its $p$-th power, as the solution to the optimal transportation problem
\begin{equation} \label{eqn:k-primal}
    \wass_p^p(\mu, \nu) = \inf_{\pi \in \Gamma(\mu, \nu)} \int c(x,y)^p \mathrm{d}\pi(x,y) = \inf_{X\sim \mu, Y\sim \nu} \mathbb{E} \left[c(X,Y)^p \right],
\end{equation}
where $\Gamma(\mu,\nu)$ is the set of all joint distributions $\pi \in \mathcal{P}(\mathcal{X} \times \mathcal{X})$ with marginals ($\mu, \nu$). The primal problem~\eqref{eqn:k-primal} admits the Kantorovich dual formulation
\begin{equation*}
\begin{gathered}
    \wass_p^p(\mu, \nu) = \sup_{(\phi, \psi) \in \Phi(\mu, \nu)} \int \phi(x) \mathrm{d}\mu(x) + \int \psi(y) \mathrm{d}\nu(y),\\
    \Phi(\mu, \nu) = \{(\phi, \psi) \in L_1(\mu) \times L_1(\nu) \mid \phi(x) + \psi(y) \le c(x,y)^p ,\enskip \forall x, y \}.
\end{gathered}
\end{equation*}
We use $(\phi_{\mu, \nu}, \psi_{\mu, \nu})$ to denote a pair of optimal potentials for the Kantorovich dual. Properties of Wasserstein distances include \citep{villani2009optimal}: $\wass_p$ defines a metric on the set of distributions with finite $p$-th moments, it induces an intuitive geometry and controls weak convergence on this set, and it controls the discrepancy between certain moments of Lipschitz functions.

In this paper, we are interested in estimating Wasserstein distances in practice. Since the behavior of Wasserstein distance estimators is extremely rich from a statistical perspective, and depends on the features of the distributions of interest as well as of the distance itself, we must make some assumptions. In this paper, we specialize to continuous distributions in $\mathcal{X} = \mathbb{R}^d$, we fix the ground metric to be Euclidean $c(x,y) =\|x - y\|,$ and we fix the exponent to $p = 2$. Throughout the entire sequel, we impose the regularity assumption:
\begin{enumerate}[label=\textbf{(A0)}]
    \item\label{assumption:cts} The distributions $\mu,\nu \in \mathcal{P}(\mathbb{R}^d)$ are absolutely continuous with respect to the Lebesgue measure on $\mathbb{R}^d$ and satisfy $\mathbb{E}_{\mu}\left[\|X\|^2\right], \mathbb{E}_{\nu}\left[\|Y\|^2\right] < \infty.$
\end{enumerate}
Brenier's theorem (\citeyear{brenier1991polar}) then provides the unique solution $\wass_2^2(\mu, \nu) = \mathbb{E}_{\nu} [\|T_{\nu, \mu}(Y) - Y\|^2]$ in terms of an optimal transport map $T_{\nu, \mu}$ that pushes $\nu$ forward to $\mu$.

We focus on the case where independent samples $X_{1:n} \sim \mu$ and $Y_{1:n} \sim \nu$ are available from each distribution. We define the empirical measures $\mu_n = \frac{1}{n}\sum_{i=1}^n \delta_{X_i}$ and $\nu_n = \frac{1}{n} \sum_{i=1}^n \delta_{Y_i}$ and we call $\wass_2^2(\mu_n,\nu_n)$ the \emph{plug-in estimator} of the squared Wasserstein distance $\wass_2^2(\mu,\nu)$, which we now review from a computational and statistical perspective.

\subsection{Computational aspects}

Exact computational methods treat the plug-in estimator $\wass_2^2(\mu_n,\nu_n)$ as the solution to a linear assignment problem.
Although the worst-case theoretical complexity of exact assignment problem solvers is $O(n^3),$ particularly efficient solvers \citep{bonneel2011displacement, guthe2021toms1015} have complexities closer to $O(n^2)$ in practice, see the benchmark of Appendix~\ref{app:bench}.

Among approximate methods, the most popular is that of \cite{cuturi2013sinkhorn}, which solves for an entropy-regularized version of $\wass_2^2(\mu_n,\nu_n)$ using Sinkhorn's algorithm. This has complexity $O(n^2 /\varepsilon^2)$ \citep{dvurechensky2018computational} depending on the size of the regularization parameter $\varepsilon$, but is well-suited to vectorized hardware such as GPUs.

In this paper, we use the exact solver of \cite{guthe2021toms1015}. This allows us to compute plug-in estimators at relatively large sample sizes $n = \Theta(10^4)$ and dimensions $d = \Theta(10^3)$ in a matter of seconds, \emph{even while only using a single CPU core}. These sample sizes suffice for our all applications. Scaling to larger $n$ would require caching \citep{guthe2021toms1015} or batching \citep{charlier2021kernel} to overcome memory limitations, and would benefit from parallelism to reduce the computing time.

\subsection{Statistical aspects}

We turn to the statistical properties of Wasserstein distance estimators. The plug-in estimator $\wass_2^2(\mu_n,\nu_n)$ is consistent \citep{villani2009optimal} and has a positive bias which decreases with the sample size $n$ (see Appendix~\ref{app:pf-plugin}):
\begin{equation*}
\begin{aligned}
    \lim_{n \to \infty}\wass_2^2(\mu_n,\nu_n) &= \wass_2^2(\mu, \nu) \enskip \textup{almost surely},\\
   \forall n: \enskip \mathbb{E}\left[ \wass_2^2(\mu_n,\nu_n) \right] &\ge \mathbb{E}\left[ \wass_2^2(\mu_{n+1},\nu_{n+1}) \right] \ge \wass_2^2(\mu, \nu).
\end{aligned}
\end{equation*}

To make further progress, we separately impose two standard assumptions from the literature:
\begin{enumerate}[label=\textbf{(A\arabic*)}]
  \setcounter{enumi}{0}
    \item\label{assumption:compact} The distributions $\mu, \nu$ are supported in the same compact set of diameter at most 1.
    \item\label{assumption:4th-moment} The distributions $\mu, \nu$ have connected support with negligible boundary. Additionally, there exists a $\delta > 0$ such that $\mathbb{E}_{\mu}\left[\|X\|^{4+\delta}\right]< \infty$ and $\mathbb{E}_{\nu}\left[\|Y\|^{4+\delta}\right] < \infty.$
\end{enumerate}
Under Assumption~\ref{assumption:compact}, the plug-in estimator $\wass_2^2(\mu_n,\nu_n)$ concentrates around its mean exponentially, and has an $L_1$ rate of convergence that decays with the dimension $d$ \citep{fournier2015rate, weed2019sharp, chizat2020faster}:
\begin{equation*}
\begin{aligned}
    &\forall \varepsilon \ge 0: \enskip \mathbb{P}\left( \left|\wass_2^2(\mu_n,\nu_n) - \mathbb{E}\left[ \wass_2^2(\mu_n,\nu_n) \right] \right| \ge \varepsilon \right) \le 2\exp(-n\varepsilon^2), \\
    &\forall d \ge 5: \enskip \mathbb{E}\left[ \left|\wass_2^2(\mu_n,\nu_n) - \wass_2^2(\mu, \nu)\right|\right] \lesssim n^{-2/d},
\end{aligned}
\end{equation*}
where $\lesssim$ hides constants that do not depend on $n$. The rate of convergence also holds in the unbounded setting \citep{staudt2024convergence}, and is furthermore minimax optimal \citep{hutter2021minimax}. Although smoother estimators can achieve better rates under stronger assumptions, they also require much greater computational expense \citep{hutter2021minimax, deb2021rates}.

Under Assumption~\ref{assumption:4th-moment}, the plug-in estimator $\wass_2^2(\mu_n,\nu_n)$ satisfies a central limit theorem \citep[CLT;][]{delbarrio2019central, delbarrio2024central}. As $n \to \infty$,
\begin{equation} \label{eqn:plugin-clt}
    \sqrt{n}\left\{ \wass_2^2(\mu_n,\nu_n) - \mathbb{E}\left[\wass_2^2(\mu_n,\nu_n)\right] \right\} 
    \Longrightarrow
    \mathcal{N}_1\left(0, \var\left\{\phi_{\mu, \nu}\left(X\right) + \psi_{\mu, \nu}\left(Y\right)\right\} \right),
\end{equation}
where $X\sim \mu$ and $Y\sim \nu$ are independent. 
We can therefore view $\wass_2^2(\mu_n,\nu_n)$ as estimating $\mathbb{E}[\wass_2^2(\mu_n,\nu_n)]$ up to Gaussian error. We now benchmark several variance estimators that could be used to construct Gaussian confidence intervals for this quantity.

\begin{figure}[ht]
    \centering
    \includegraphics[width=\textwidth]{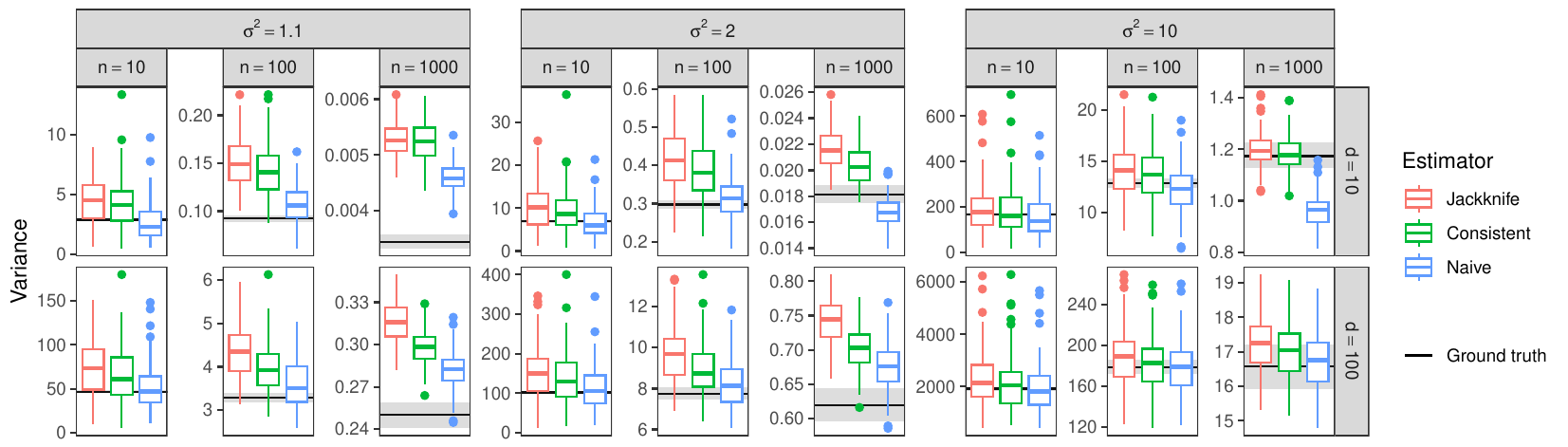}
    \caption{Variance estimation for $\wass_2^2(\mu_n, \nu_n)$ with $\mu = \mathcal{N}_d(0_d,I_d)$, $\nu = \mathcal{N}_d(0_d,\sigma^2 I_d)$ and various methods and values of $(\sigma^2,n,d)$. Unbiased estimates of the ground truth from 5000 replicates are shown with 95\% bootstrap confidence intervals.}
    \label{fig:prelim-var}
\end{figure}

\paragraph{Variance estimation.} We consider several ways of estimating the variance of the plug-in estimator $\wass_2^2(\mu_n,\nu_n)$: (i) the jackknife \citep{efron1981jackknife}, (ii) a consistent estimator based on the Kantorovich potentials \citep{delbarrio2024central} and (iii) a naive estimator.

Firstly, jackknife variance estimates are known to be conservative; in our context due to algorithmic considerations, the jackknife can be computed in $O(n^3)$ operations. (See Appendix~\ref{app:flapjack} for our procedure ``Flapjack" based on \citealp{mills-tettey2007dynamic}.) Secondly, the central limit theorem~\eqref{eqn:plugin-clt} suggests the consistent variance estimator
\begin{equation*}
    \var \left( \wass_2^2(\mu_n, \nu_n) \right) \approx \frac{1}{n}\var \left( \left\{\phi_{\mu_n, \nu_n}(X_i) + \psi_{\mu_n, \nu_n}(Y_i) \right\}_{i=1}^n \right),
\end{equation*}
where $\var(\{x_i\}_{i=1}^n) = \frac{1}{n-1}\sum_{i=1}^n (x_i - \frac{1}{n}\sum_{i=1}^n x_i)^2$  is the sample variance. This estimator is appealing as optimal potentials $(\phi_{\mu_n, \nu_n}, \psi_{\mu_n, \nu_n})$ are available without additional computation with many solvers, including that of \cite{guthe2021toms1015}. Finally, since $\wass_2^2(\mu_n,\nu_n) = \frac{1}{n}\sum_{i=1}^n \|X_i - Y_{\sigma(i)}\|^2$ for an optimal permutation $\sigma$, one might naively consider the sample variance of the preceding average, implicitly assuming that all terms are independent. This estimator is also available for little added cost, but is inconsistent.

We compare the three methods in Figure~\ref{fig:prelim-var}: we prefer the consistent estimator (ii) as it is slightly conservative and quick to compute. All variance estimators have substantial positive biases as $\nu \weakly \mu,$ because in this regime $\phi_{\mu, \nu}, \psi_{\mu, \nu} \to 0$ and therefore the asymptotics~\eqref{eqn:plugin-clt} break down to a point mass $\delta_0.$

\subsection{Tractable scenarios} \label{sec:empirical-tractable}

Certain structural conditions ease the computational and statistical challenges in estimating Wasserstein distances. For Gaussians, it holds that
\begin{equation*}
    \wass_2^2(\mathcal{N}_d(m_\mu, \Sigma_\mu), \mathcal{N}_d(m_\nu, \Sigma_\nu)) = \|m_\mu - m_\nu\|^2 + \tr\big(\Sigma_\mu + \Sigma_\nu - 2(\Sigma_\mu^{1/2} \Sigma_\nu \Sigma_\mu^{1/2})^{1/2}\big),
\end{equation*}
where $\Sigma^{1/2}$ denotes the principal square-root of $\Sigma$. An estimator of $\wass_2^2$ with favorable statistical properties \citep{rippl2016limit} can be obtained by plugging in estimated means and covariances, for $\Theta(n^2d + d^3)$ overall cost. Similar considerations apply to compatible elliptical distributions, see \citet[Remarks 2.31-32]{peyre2019computational}.

For one-dimensional measures, it holds that $\wass_2^2(\mu, \nu) = \smallint_0^1 | F_\mu^{-1}(u) - F_\nu^{-1}(u) |^2  \mathrm{d} u$ where $(F_\mu^{-1}, F_\nu^{-1})$ are the inverse-CDFs of $(\mu, \nu)$ which need not be continuous. In this case, the plug-in estimator $\wass_2^2(\mu_n, \nu_n)$ has favorable statistical properties \citep{bobkov2019one-dimensional}. It is also fast to compute, requiring the $O(n \log n)$ sorting of the two samples; the Kantorovich potentials can be recovered in $\Theta(n)$ operations \citep[Algorithm~3]{sejourne2022faster}. Similar considerations apply to product measures, due to tensorization: $\wass_2^2( \otimes_{i=1}^d \mu^{i}, \otimes_{i=1}^d \nu^{i}) = \sum_{i=1}^d \wass_2^2(\mu^i, \nu^i).$

\cite{gelbrich1990formula} and the tensorization of the squared Euclidean metric provide the tractable lower bound
\begin{equation}\label{eqn:computable-lb}
	\wass_2^2\left( \mathcal{N}_d(m_\mu, \Sigma_\mu), \mathcal{N}_d(m_\nu, \Sigma_\nu) \right) \lor \wass_2^2( \otimes_{i=1}^d \mu^{i},  \otimes_{i=1}^d \nu^{i}) \le \wass_2^2(\mu, \nu),
\end{equation}
where now $(m, \Sigma)$ denote expectations and covariances, and where superscripts denote coordinate-wise marginals. In Section~\ref{sec:bias}, we make use of this lower bound; since its finite-sample estimators are positively biased and noisy, we use the jackknife to correct the bias \citep{miller1974jackknife} and to quantify the additional noise \citep{efron1981jackknife}.

%% file: sections/3-debias.tex
\section{Centered plug-in estimators} \label{sec:main-results}

In applications, it is important for estimators of $\wass_2^2(\mu, \nu)$ to be informative in the regime $\nu \weakly \mu$: in addition to distinguishing between measures, we want to be able to recognize when they are similar. 
Even in low-dimensional scenarios, the plug-in estimator $\wass_2^2(\mu_n, \nu_n)$ does not satisfy this criterion, because it has a large bias that decays slowly with $n$ and becomes particularly apparent as $\nu \weakly \mu$. 
Since the bias cannot be meaningfully reduced by increasing the sample size, we must obtain informative estimators by different means.

We propose to render plug-in estimators of $\wass_2^2(\mu, \nu)$ informative by centering them. Formally, we assume that empirical measures $\mu_n = \frac{1}{n}\sum_{i=1}^n \delta_{X_i}$, $\bar\mu_n = \frac{1}{n}\sum_{i=1}^n \delta_{\bar{X}_i}$, $\nu_n = \frac{1}{n}\sum_{i=1}^n \delta_{Y_i}$, $\bar\nu_n = \frac{1}{n}\sum_{i=1}^n \delta_{\bar{Y}_i}$ are available, based on independent samples $X_{1:n}, \bar{X}_{1:n} \sim \mu$ and $Y_{1:n}, \bar{Y}_{1:n} \sim \nu$. The new centered estimators are:
\begin{equation*}
    \begin{aligned}
        U(\bar\mu_n, \mu_n,\nu_n) &= \wass_2^2(\bar\mu_n, \nu_n) - \wass_2^2(\bar\mu_n, \mu_n),\\
        L(\bar\mu_n, \mu_n,\nu_n) &= \sgnpow{2}{\wass_2(\bar\mu_n, \nu_n) - \wass_2(\bar\mu_n, \mu_n)},
    \end{aligned}
\end{equation*} 
where $\sgnpow{2}{x} = \sgn(x) x^2$, i.e. $L$ is the signed square of $\bar L(\bar\mu_n, \mu_n,\nu_n) = \wass_2(\bar\mu_n, \nu_n) - \wass_2(\bar\mu_n, \mu_n)$.

\begin{figure}
    \centering
    \includegraphics[width = \textwidth]{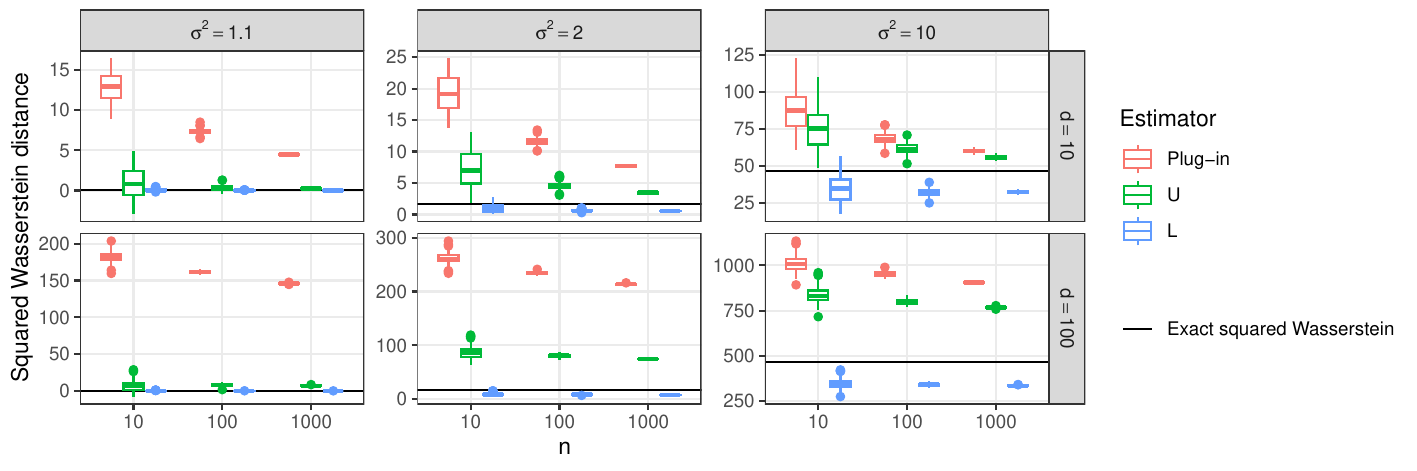}
    \caption{Comparison of plug-in estimator $\wass_2^2(\mu_n, \nu_n)$ and proposed estimators $U(\bar\mu_n, \mu_n, \nu_n)$ and $L(\bar\mu_n, \mu_n, \nu_n)$, with $\mu = \mathcal{N}_d(0_d,I_d)$, $\nu = \mathcal{N}_d(0_d,\sigma^2 I_d)$ and various values of $(\sigma^2,n,d)$.}
    \label{fig:prelim-comparison}
\end{figure}

The centering ensures that the proposed estimators are informative, as their expectations decrease to zero with $\wass_2^2(\mu, \nu)$ for any finite sample size. 
More importantly, the proposed estimators can be viewed as complementary bounds on $\wass_2^2(\mu, \nu)$: $U(\bar\mu_n, \mu_n,\nu_n)$ is an approximate upper bound when $\nu$ is overdispersed with respect to $\mu$; $L(\bar\mu_n, \mu_n,\nu_n)$ is an approximate lower bound in general. We establish these properties, and we discuss suitable notions of overdispersion, in Section~\ref{sec:bias-u-l}. Figure~\ref{fig:prelim-comparison} illustrates these properties: notably, centering reduces the bias without increasing the variance.

Increasing the sample size $n$ benefits the proposed estimators by decreasing the variance, reducing the bias and, as we shall see, further relaxing the conditions required for $U$ to be conservative. Trading some of these benefits off for faster computation, $\{U, L\}$ could be replaced by sample averages computed at a lower sample size. We establish the statistical properties of the proposed estimators in Section~\ref{sec:stat-u-l}, and we discuss uncertainty quantification in Section~\ref{sec:uq-u-l}.

We conclude this introduction with two practical refinements of our methodology.

\paragraph{Hedging.}

Taking the maximum of two estimators, we can obtain more generally applicable upper bounds and tighter lower bounds, as with the pair 
\begin{equation*}
    V(\mu_n,\nu_n, \bar\mu_n, \bar\nu_n) = U(\bar\mu_n, \mu_n, \nu_n) \lor U(\bar\nu_n, \nu_n, \mu_n) \enskip \text{and} \enskip L(\bar\mu_n, \mu_n,\nu_n) \lor L(\bar\nu_n, \nu_n, \mu_n).
\end{equation*}
The first hedging strategy is particularly useful when \emph{a priori} it is unclear which one of $\{\mu, \nu\}$ is more dispersed. Our experiments indicate that $V$ is often conservative, even when it is used naively.

\paragraph{Variance reduction using couplings.} 

When the sample generation can be controlled, positively correlating $(\mu_n, \nu_n)$ can reduce the variances of $U(\bar\mu_n, \mu_n, \nu_n),$ $L(\bar\mu_n, \mu_n,\nu_n)$ and $V(\mu_n,\nu_n, \bar\mu_n, \bar\nu_n)$ with little effect to their biases. This technique can reduce the variance substantially, particularly when $\wass_2^2(\mu, \nu)$ is small, see Section~\ref{sec:bias}.

\subsection{Analysis of the bias} \label{sec:bias-u-l}

We analyze the biases of the proposed estimators, showing that they are informative and providing conditions under which they can be viewed as approximate bounds. We recall that the minimal regularity Assumption~\ref{assumption:cts} applies.

Proposition~\ref{prop:l} establishes that $\bar L$ is not conservative.

\begin{proposition} \label{prop:l}
It holds that $\mathbb{E} \left[ \bar L(\bar\mu_n, \mu_n,\nu_n) \right]^2 = \mathbb{E} \left[ \wass_2(\bar\mu_n, \nu_n) - \wass_2(\bar\mu_n, \mu_n) \right]^2 \le \wass_2^2(\mu, \nu).$
\end{proposition}

Theorem~\ref{thm:u} establishes properties of $U$. We show that an appropriate condition on the optimal transport map $T_{\nu, \mu}$ ensures that $U$ is conservative, that $U$ remains informative as $\nu \weakly \mu$, and that $U$ is location-free.

\begin{definition}[Contractive optimal transport] \label{def:cot}  
    We write $\nu \cotr \mu,$ and say that $\nu$ is contractively optimally transported to $\mu,$ if the optimal transport map $T_{\nu, \mu}$ is a contraction, that is it has Lipschitz constant $\lip{T_{\nu, \mu}}\le 1.$
\end{definition}

\begin{theorem}\label{thm:u} 
The following assertions hold:
\begin{enumerate}[label=(\roman*)]
\item If $\nu \cotr \mu$, then $\mathbb{E} \left[ U(\bar\mu_n, \mu_n, \nu_n)\right] = \mathbb{E} \left[ \wass_2^2(\bar\mu_n, \nu_n) - \wass_2^2(\bar\mu_n, \mu_n) \right] \ge \wass_2^2(\mu, \nu).$
\item $\mathbb{E} \left[ U(\bar\mu_n, \mu_n, \nu_n)\right] \le K(\mu, \nu) \wass_2(\mu, \nu),$ where $K(\mu, \nu) = 3 \mathbb{E}_\mu[\lVert X \rVert^2]^{1/2} + \mathbb{E}_\nu[\lVert Y \rVert^2]^{1/2}$.
\item $\mathbb{E} \left[ U(\bar\mu_n, \mu_n, \nu_n) \right] - \wass_2^2(\mu, \nu)$ is invariant to shifting the expectation of either $\mu$ or $\nu$.
\end{enumerate}
\end{theorem}

We emphasize that the condition $\nu \cotr \mu$ of Theorem~\ref{thm:u}(i) is purely sufficient: it is what we use to formulate an otherwise generic result, which holds for all sample sizes $n,$ all dimensions $d,$ and does not impose structural assumptions on either measure $\mu$ or $\nu$.

We interpret the condition $\nu \cotr \mu$ in Section~\ref{sec:interpret-cot}; in Section~\ref{sec:robust-overdisp}, we demonstrate that the estimator~$U$ is in fact conservative much more generally. Since the inequality $\mathbb{E} \left[ U(\bar\mu_n, \mu_n, \nu_n) \right] \ge \wass_2^2(\mu, \nu)$ is location-free, its validity clearly only depends on how the dispersions of $\mu$ and $\nu$ are related. For $U$ to be an overestimate, the correct relation turns out to be that of overdispersion.

\begin{remark}
    Brenier's theorem states that $T_{\nu, \mu} = \nabla \varphi_{\nu, \mu}$ for a convex $\varphi_{\nu, \mu}$. The condition of Theorem~\ref{thm:u}(i) is the global Hessian bound $\nabla^2 \varphi_{\nu, \mu} \succeq I_d$ and resembles conditions used by recent computational \citep{paty2020regularity} and theoretical \citep{hutter2021minimax,deb2021rates,manole2024plugin} work. After finalizing a preliminary version of this manuscript, we became aware of an independently derived result from a \href{https://arxiv.org/abs/2107.12364v1}{preprint version} of \cite{manole2024plugin} that is similar to Theorem~\ref{thm:u}(i). We use our result for different purposes.
\end{remark}

\subsubsection{Interpreting contractive optimal transport} \label{sec:interpret-cot}

The condition $\nu \cotr \mu$ is location-free. This hints at a connection between $\cotr$ and stochastic orderings \citep{shaked2007stochastic}, which we now discuss.

For one-dimensional measures, the univariate dispersive ordering $\nu \gtrunidisp \mu$ \citep{shaked1982dispersive} requires the quantiles of $\nu$ to lie further apart than the corresponding quantiles of $\mu$. The condition $\nu \cotr \mu$ coincides with $\nu \gtrunidisp \mu$, because the optimal transport map $T_{\nu, \mu} = F^{-1}_\mu \circ F_\nu$ maps between the corresponding quantiles of $\nu$ and $\mu$. In general, $\nu\cotr\mu$ implies the SD-ordering of \cite{giovagnoli1995multivariate}, which requires the existence of a contractive map transporting $\nu$ to $\mu$. However, the SD-ordering does not provide a meaningful way of distinguishing between measures: for instance, $\mu$ and $\nu$ are equal under this ordering whenever they differ by a rotation, yet $\wass_2^2(\mu,\nu)$ could be arbitrarily large.

We draw further connections between $\nu \cotr \mu$ and stochastic orderings under structural assumptions.

\begin{proposition} \label{prop:cot-cases}
The following assertions hold:
\begin{enumerate}[label=(\roman*)]
    \item For Gaussians, $\mathcal{N}_d(m_\nu, \Sigma_\nu) \cotr \mathcal{N}_d(m_\mu, \Sigma_\mu)$ if and only if $\Sigma_\nu \succeq \Sigma_\mu$, where $\succeq$ is the Loewner order.
    \item For spherically symmetric measures, $\nu \cotr \mu$ if and only if the same relation holds between the distributions of their radial components.
    \item For product measures, $(\otimes_{i=1}^d \nu^i) \cotr (\otimes_{i=1}^d \mu^i)$ if and only if $\nu^i \cotr \mu^i$ for all $i$.
    \item If $\nu(x) \propto \exp(-N(x))$ and $\mu(x) \propto \exp(-M(x))$  with twice differentiable $N,M$ with convex support, and if $\nabla^2 N \preceq A \preceq \nabla^2 M$ holds point-wise for a fixed positive definite matrix $A$, then $\nu \cotr \mu.$
\end{enumerate}
\end{proposition}

Overall, we view $\nu \cotr \mu$ as a global overdispersion condition: $\nu$ must be a shifted version of $\mu$ that is more spread-out in all directions. In addition to providing key intuition, this condition suggests that the estimators could be useful to assess the quality of Bayesian computation methods, where over- and underdispersion is pervasive, see Sections~\ref{sec:bias} and~\ref{sec:convergence}.

We conjecture that $\cotr$ does not define a partial order in general, and leave this as an open problem.

\subsubsection{When is $U$ conservative in practice?} \label{sec:robust-overdisp}

We investigate the conditions required for $U$ to be conservative in practice. We begin with a sharp characterization of the small-$n$ case, see Appendix~\ref{app:debias-n=1}.

\begin{example}[$n=1$] \label{ex:debias-n=1}
The inequality $\mathbb{E} \left[ U(\bar\mu_1, \mu_1, \nu_1)\right] \ge \wass_2^2(\mu, \nu)$ is equivalent to
\begin{equation*}
     \sup_{(X,Y)\sim (\mu,\nu)} \tr(\cov(X,Y)) \ge \tr(\var(X)), \enskip \text{denoted by} \enskip \nu \pcar \mu.
\end{equation*}
Intuitively, \emph{$\nu$ is more dispersed than $\mu$, averaged along the principal components of $\mu$}.

In particular, $\pcar$ is partially closed under mixtures. Furthermore $\nu \pcar \mu$ holds under the convex order \citep{strassen1965existence}, which provides the intuition that it suffices for $\nu$ to be a diffuse version of $\mu$.
\end{example}

For large $n,$ one might expect consistency to weaken the conditions under which $U$ is conservative. However, the challenge in obtaining the exact expression of the bias to first order in $n$ precludes a general analysis. Instead, we derive a sharp result in the one-dimensional case, see Appendix~\ref{app:debias-d=1}.

\begin{example}[$d=1$] \label{ex:debias-d=1}
Under regularity conditions, in dimension $d=1$ it holds that 
\begin{equation*}
    \lim_{n \to \infty} n \left(\mathbb{E} \left[ U(\bar\mu_n, \mu_n, \nu_n)\right] - \wass_2^2(\mu, \nu)\right)
     \ge 0 \enskip \text{if and only if} \enskip J(\mu, \nu) \ge J(\mu, \mu),
\end{equation*}
where $J(\mu, \nu) = \int_0^1 u(1-u) (F_\mu^{-1})'(u)(F_\nu^{-1})'(u)\mathrm{d}u.$ This condition is significantly milder than $\nu \cotr \mu$, which asks for $(F_\nu^{-1})' \ge (F_\mu^{-1})'$ uniformly.
\end{example}

Examples~\ref{ex:debias-n=1} and~\ref{ex:debias-d=1} indicate that a partial overdispersion can also ensure that $U$ is conservative. This recommends the estimator $V$ for general use. Whether $V$ is conservative depends on the compatibility of the measures: the centering term of $V$ may over-correct when most of the masses of $\mu$ and $\nu$ lie in directions orthogonal to each other. In practice, the compatibility of the measures can be checked using a principal component analysis.

\begin{figure}[ht]
    \centering
    \includegraphics[width = \textwidth]{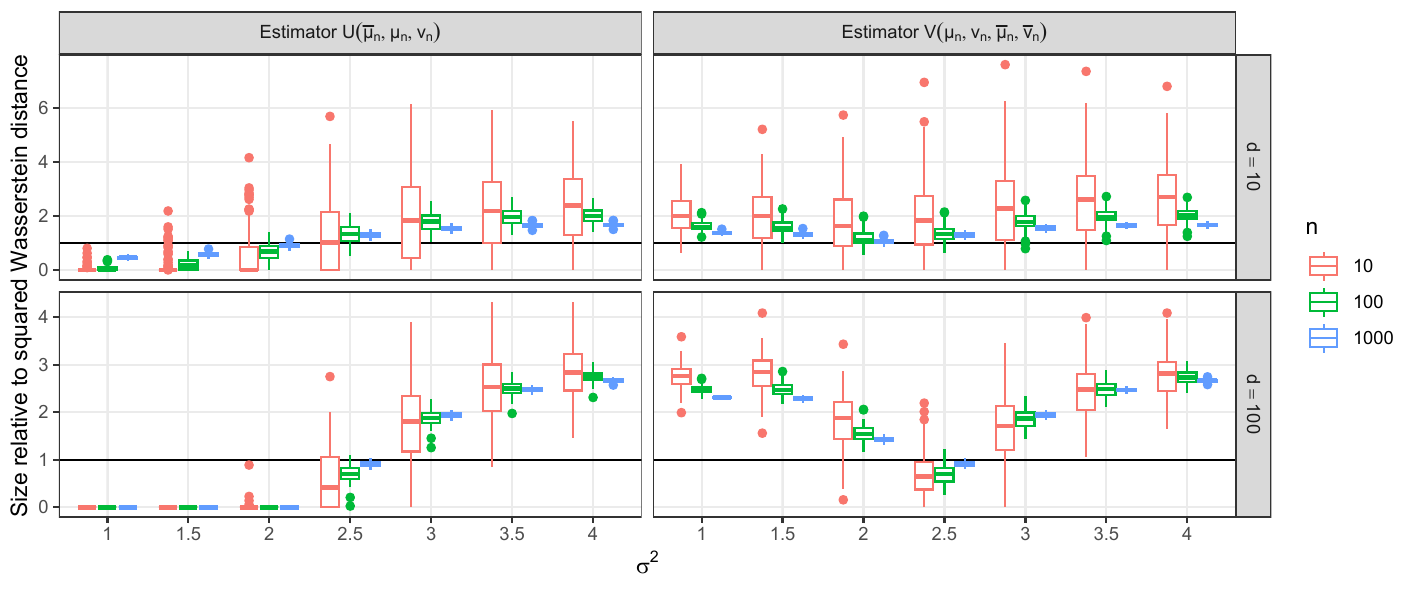}
    \caption{Robustness of proposed estimators $\{U,V\}$ to the degree of overdispersion, with $\mu = \mathcal{N}_d(0_d, \diag(1,4) \otimes I_{d/2})$ and $\nu = \mathcal{N}_d(0_d,\sigma^2 I_d)$ and various $(\sigma^2,n,d)$. The relation $\nu \cotr \mu$ holds for $\sigma^2 = 4$ 
    and $\nu \pcar \mu$ holds for $\sigma^2\ge 2.87$ (resp. $\mu \pcar \nu$ for $\sigma^2\le 2.25$ and $\mu \cotr \nu$ for $\sigma^2 = 1$). Negative estimates are set to zero.}
    \label{fig:prelim-overdisp}
\end{figure}

Figure~\ref{fig:prelim-overdisp} illustrates that the proposed estimators $\{U,V\}$ are robust: they are conservative under relatively weak forms of overdispersion. We see that $U(\bar\mu_n, \mu_n, \nu_n)$ is conservative as long as $\nu$ is more dispersed than $\mu$ on average. That it is not conservative when $\nu$ is significantly less dispersed than $\mu$ should not be concerning to the reader: in practice, one would have swapped the roles of the measures and used the estimator $U(\bar\nu_n,\nu_n, \mu_n)$ instead. This effectively amounts to using the estimator $V$, which at the largest sample size is sensible throughout the considered scenario.

\subsection{Statistical properties} \label{sec:stat-u-l}

We study the statistical properties of the proposed estimators. It is clear that they are consistent; they additionally inherit the concentration and near-minimax rate of convergence of the plug-in estimators they are composed of.

\begin{theorem} \label{thm:concentration-bounds}
Under Assumption~\ref{assumption:compact}, it holds that
\begin{align*}
    \forall \varepsilon \ge 0:& \enskip \mathbb{P}\left( \left|U(\bar\mu_n, \mu_n,\nu_n) - \mathbb{E} \left[U(\bar\mu_n, \mu_n,\nu_n)\right]\right| \ge \varepsilon \right) \le 2\exp \left(-n\varepsilon^2 / 3 \right),\\
    \forall \varepsilon \ge 0:& \enskip \mathbb{P}\left( \left|\bar L(\bar\mu_n, \mu_n,\nu_n) - \mathbb{E} \left[\bar L(\bar\mu_n, \mu_n,\nu_n) \right]\right| \ge \varepsilon \right) \le 2\exp \left(-n\varepsilon^4 / 32 \right).
\end{align*}
\end{theorem}

\begin{theorem} \label{thm:convergence-rates}
Let $\mu \ne \nu.$ Under Assumption~\ref{assumption:compact}, it holds that
\begin{equation*}
    \forall d \ge 5: \enskip \mathbb{E} \left[\left| U(\bar\mu_n, \mu_n,\nu_n) - \wass_2^2(\mu, \nu) \right|\right] \lesssim n^{-2/d}, \quad \mathbb{E}\left[\left| \wass_2(\mu, \nu) - \bar L(\bar\mu_n, \mu_n,\nu_n) \right|\right] \asymp n^{-1/d},
\end{equation*}
where $\asymp$ denotes decay at the exact rate.
\end{theorem}

As a consequence, the proposed estimators are high-probability bounds as soon as they are bounds in expectation. We emphasize that this does not require the sufficient condition $\nu \cotr \mu$: in Corollary~\ref{cor:bounds-high-prob}, we ask for $U(\bar\mu_n, \mu_n,\nu_n)$ to be positively biased by an amount which decays in $n$ at the rate of Theorem~\ref{thm:convergence-rates}.

\begin{corollary} \label{cor:bounds-high-prob}
Let $\mu \ne \nu.$ Under Assumption~\ref{assumption:compact}, it holds that
\begin{equation*}
    \forall d \ge 5: \enskip \mathbb{P}\left(L(\bar\mu_n, \mu_n,\nu_n) \le \wass_2^2(\mu, \nu)\right) = \mathbb{P}\left(\bar L(\bar\mu_n, \mu_n,\nu_n) \le \wass_2(\mu, \nu)\right) \ge 1 - \exp\big(- C_1 n^{1 - 4/d}\big).
\end{equation*}
If additionally $\mathbb{E} \left[ U(\bar\mu_n, \mu_n,\nu_n) \right] - \wass_2^2(\mu, \nu) \gtrsim n^{-2/d},$ it holds that
\begin{equation*}
    \forall d \ge 5: \enskip \mathbb{P}\left(U(\bar\mu_n, \mu_n,\nu_n) \ge \wass_2^2(\mu, \nu)\right) \ge 1 - \exp\big(- C_2 n^{1 - 4/d}\big).
\end{equation*}
The constants $C_1,C_2 > 0$ only depend on the measures $\mu, \nu$ and on the dimension $d.$
\end{corollary}

Similar properties can be shown for $V$, with the hedging ensuring that this estimator is a high-probability bound on $\wass_2^2(\mu,\nu)$ under weaker conditions. We avoid further technical details.

\subsection{Uncertainty quantification} \label{sec:uq-u-l}

We describe how to quantify the uncertainty of the proposed estimators based on their asymptotic distributions. The estimator $U$ obeys a Gaussian CLT, as a direct consequence of \citet[Theorem~4.10]{delbarrio2024central} and Slutsky's theorem, which we state without proof.

\begin{theorem} \label{thm:clt-u}
Under Assumption~\ref{assumption:4th-moment} it holds that, as $n \to \infty$,
\begin{equation*}
    \sqrt{n}\left( U(\bar\mu_n, \mu_n,\nu_n) - \mathbb{E}\left[ U(\bar\mu_n, \mu_n,\nu_n) \right] \right) 
    \Longrightarrow
    \mathcal{N}_1\left(0, \sigma^2 \right) \enskip \text{and} \enskip \lim_{n \to \infty}n \var\left\{ U(\bar\mu_n, \mu_n,\nu_n) \right\} = \sigma^2,
\end{equation*}
where $\sigma^2 = \var\{\phi_{\mu, \nu}(X) + \psi_{\mu, \nu}(Y)\}$ under independent $X\sim\mu$ and $Y\sim\nu.$
\end{theorem}

Formal results for $\bar L$ are more challenging because $\wass_2(\bar\mu_n, \mu_n)$ lacks a satisfactory limiting theory \citep{delbarrio2024central}, but experiments indicate that $\bar L$ is approximately Gaussian even for small $n$.

To quantify the variability of $\{U, \bar L\},$ we therefore use Gaussian confidence intervals. For $L = [\bar L]_{\pm}^2,$ we transform by $[\cdot]_{\pm}^2$ the confidence interval for $\bar L$. For estimators like $V$ that are formed as the maximum of two components, we use the confidence interval corresponding to the active component; the slight underestimation balances out with our conservative variance estimates, which we next describe.

The confidence intervals require variance estimates, we propose to use
\begin{equation*}
\begin{aligned}
    \var(U) &\approx \frac{1}{n}\var\left(\left\{ \phi_{\bar\mu_n, \nu_n}(\bar X_{i}) + \psi_{\bar\mu_n, \nu_n}(Y_{i}) - \phi_{\bar\mu_n, \mu_n}(\bar X_{i}) - \psi_{\bar\mu_n, \mu_n}(X_{i}) \right\}_{i=1}^n\right),\\
    \var(\bar L) &\approx \frac{1}{n}\var\left(\left\{ \frac{\phi_{\bar\mu_n, \nu_n}(\bar X_{i}) + \psi_{\bar\mu_n, \nu_n}(Y_{i})}{2\wass_2(\bar\mu_n, \nu_n)} - \frac{\phi_{\bar\mu_n, \mu_n}(\bar X_{i}) + \psi_{\bar\mu_n, \mu_n}(X_{i})}{2\wass_2(\bar\mu_n, \mu_n)} \right\}_{i=1}^n \right),
\end{aligned}
\end{equation*}
justified in turn by Theorem~\ref{thm:clt-u} and an approximate delta method for $\bar L$ (detailed in Appendix~\ref{app:uq-lbar}). Figure~\ref{fig:debiased-var} compares the proposed consistent variance estimator of $U$ with the jackknife, which is conservative and available with an additional $O(n^3)$ computation using the Flapjack algorithm (Appendix~\ref{app:flapjack}). We prefer the consistent estimator for its smaller positive bias and lower computing cost. Similar considerations hold for the variance estimator of $\bar L$. 

\begin{figure}[ht]
    \centering
    \includegraphics[width=\textwidth]{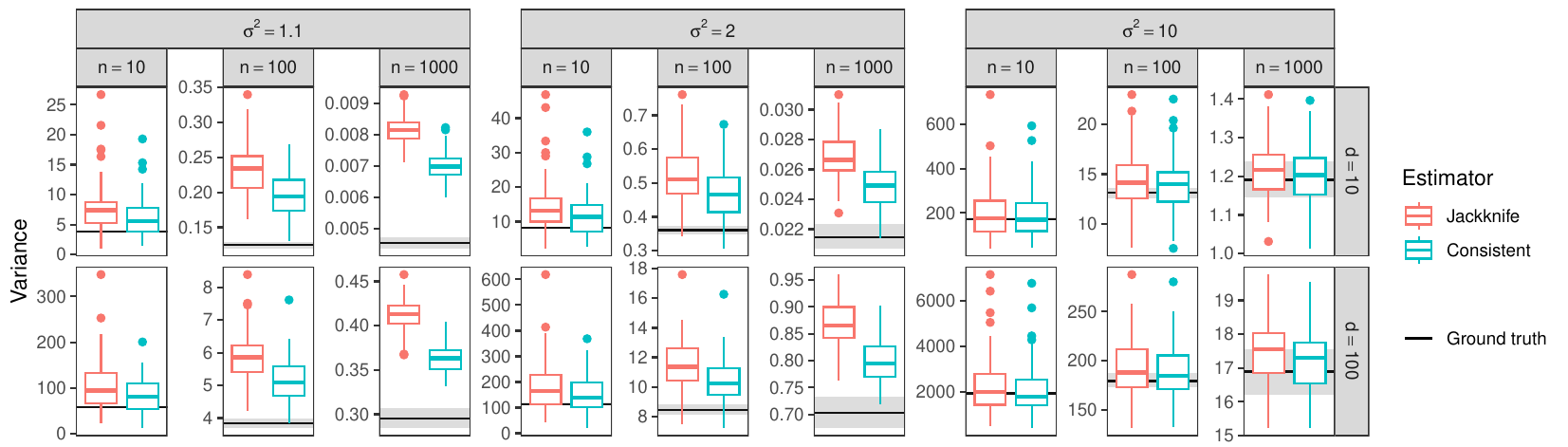}
    \caption{Variance estimates for $U(\bar\mu_n, \mu_n, \nu_n)$ with $\mu = \mathcal{N}_d(0_d,I_d)$, $\nu = \mathcal{N}_d(0_d,\sigma^2 I_d)$ and various methods and values of $(\sigma^2,n,d)$. Unbiased estimates of the ground truth from 5000 replicates are shown with 95\% bootstrap confidence intervals.}
    \label{fig:debiased-var}
\end{figure}

The variance estimates also remain valid when the pairs $(X_i,Y_i)$ are sampled i.i.d. from any coupling of $(\mu, \nu)$, so the confidence intervals correctly account for the variance reduction afforded by positively correlating $(\mu_n, \nu_n)$. As we explain in Appendix~\ref{app:uq-bias}, we can estimate this reduction in variance without additional simulation. Finally, these uncertainty quantification methods can be generalized to correlated samples and to averages of plug-in estimators, see Appendices~\ref{app:uq-bias} and~\ref{app:uq-conv}. These generalizations will prove useful in the applications of Sections~\ref{sec:bias} and~\ref{sec:convergence}.

%% file: sections/4-mcmc-bias.tex
\section{Assessing the quality of approximate inference methods} \label{sec:bias}

Reliably assessing the quality of approximate Bayesian inference methods is one of the grand challenges of Bayesian computation \citep{bhattacharya2024grand}, a problem that is of interest both to the researchers developing such methods, as well as to the practitioners using them. We propose here to estimate the squared Wasserstein distance $\wass_2^2(\mu, \nu)$ of approximations~$\nu$ to exact models~$\mu$ with the centered estimators of Section~\ref{sec:main-results}.

\subsection{Methodology} \label{sec:bias-sampling}

We advocate using MCMC to sample from the model $\mu$ and the approximation $\nu,$ in the following way. 
We sample i.i.d. $\mu$-invariant Markov chains $(\smash{X^{(t)}_k})_{t = 0}^{B_\mu+T_\mu(I-1)}$ and $\nu$-invariant chains $(\smash{Y^{(t)}_k})_{t = 0}^{B_\nu+T_\nu(I-1)}$ for $k \in [2K]$. 
We discard, respectively, $\{B_\mu, B_\nu\}$ iterations as burn in, and thin the remainder of each chain by factors of $\{T_\mu, T_\nu\}$ to provide the empirical measures
\begin{equation*}
\begin{alignedat}{3}
    \mu_n &=\frac{1}{KI} \sum_{k=1}^K \sum_{i=0}^{I-1} \delta_{X^{(B_\mu + T_\mu i)}_k}, 
     \quad
     \bar \mu_n &&=\frac{1}{KI}\sum_{k=K+1}^{2K} \sum_{i=0}^{I-1} \delta_{X^{(B_\mu + T_\mu i)}_{k}},\\
     \nu_n &=\frac{1}{KI} \sum_{k=1}^K \sum_{i=0}^{I-1} \delta_{Y^{(B_\nu + T_\nu i)}_k}, 
     \quad 
     \bar\nu_n &&=\frac{1}{KI} \sum_{k=K+1}^{2K} \sum_{i=0}^{I-1} \delta_{Y^{(B_\nu + T_\nu i)}_k}, 
\end{alignedat}
\end{equation*}
each with $n=KI$ samples. We modify the confidence intervals to account for within-chain sample dependence in Appendix~\ref{app:uq-bias}.

Our parameter guidelines are motivated by the insight that the biases of the proposed estimators primarily depend on the smallest of the effective sample sizes \citep[ESSes; e.g.][]{vats2019multivariate} within the empirical measures $\{\mu_n, \nu_n\}$. We therefore recommend setting the thinning factors $\{T_\mu, T_\nu\}$ such that the ESSes are roughly equal,\footnote{That is, we recommend performing more iterations with the slowest-mixing chain.} and increasing $\{K,I\}$ until a target ESS is attained. The burn-ins $\{B_\mu, B_\nu\}$ should be set based on estimates of the rate of convergence, see Section~\ref{sec:convergence}. Our experience is that the estimators are robust to small $\{T_\mu, T_\nu\}$.

To reduce the variance of estimators, we can induce a positive correlation between $(\mu_n, \nu_n)$ by coupling the pairs $\smash{(X_k^{(t)}, Y_k^{(t)})}$ and setting $(B_\mu, T_\mu) = (B_\nu, T_\nu)$. We consider various practical coupling strategies based on common random numbers (CRNs) in Section~\ref{sec:bias-numerical}.

\subsection{Approximate inference methods} \label{sec:bias-dispersion}

We discuss several common types of approximate inference methods $\nu$, focusing on how their variabilities relate to that of the exact model $\mu$. The general trend is that approximate inference methods tend to be over- or underdispersed versions of the exact model, and so we typically expect the estimators of Section~\ref{sec:main-results} to reliably bound the squared Wasserstein distance $\wass_2^2(\mu, \nu)$.

\paragraph{Laplace approximations.} A Laplace approximation $\nu$ is the best Gaussian fit around a mode of the density of the true model $\mu$. Since Laplace approximations are local, they typically underestimate the variability, particularly if $\mu$ has heavier-than-Gaussian tails or it has multiple modes. Other types of localized approximations can similarly be expected to underestimate the variability in the true model.

\paragraph{Variational approximations.} Variational inference \citep[VI;][]{blei2017variational} uses optimization to fit the approximation $\nu$. The approximating family is often Gaussian. The objective is typically the \emph{exclusive} (or \emph{reverse}) Kullback-Leibler (KL) divergence $\kl(\cdot \| \mu)$, which tends to produce local approximations which underestimate the true variability \citep{wang2005inadequacy}. Conversely, expectation propagation \citep[EP;][]{minka2001expectation} is an algorithm that optimizes for the \emph{inclusive} (or \emph{forward}) KL divergence $\kl(\mu \| \cdot)$. EP appears to have two regimes, either globally overestimating the true variability or globally underestimating it  \citep{dehaene2018expectation}.

\paragraph{Approximate MCMC algorithms.} Certain gradient-based unadjusted MCMC algorithms, such as the unadjusted Langevin algorithm \citep[ULA;][]{roberts1996exponential} and the OBABO discretization of the \emph{underdamped} (or \emph{kinetic}) Langevin diffusion \citep[e.g.][]{monmarche2021high}, tend to have stationary distributions $\nu$ that are overdispersed versions of the exact target $\mu$. We verify this analytically for Gaussian targets $\mu$.

\begin{proposition} \label{prop:ula-bias-overdisp}
    The stationary distribution $\nu$ of an ULA or OBABO chain targeting a Gaussian $\mu$ satisfies $\nu \cotr \mu$.
\end{proposition}

Stochastic gradient MCMC algorithms \citep{ma2015complete} are gradient-based unadjusted MCMC algorithms where the gradient is replaced by an unbiased estimate; they are popular in tall-data applications. The additional noise typically causes the stationary distribution $\nu$ of a stochastic gradient MCMC algorithm to be an overdispersed version of the target $\mu$ \citep{nemeth2021stochastic}.

Exact and approximate Gibbs samplers for high-dimensional linear regression models with horseshoe priors \citep{carvalho2010horseshoe} were developed in \cite{johndrow2020scalable}; these samplers were later extended to more general half-t priors in \cite{biswas2021coupling-based, biswas2024bounding}. We explain why these approximate Gibbs samplers generate overdispersed versions $\nu$ of the exact target $\mu$ in Appendix~\ref{app:half-t-overdisp}.

\paragraph{Approximate Bayesian computation.}  Approximate Bayesian computation (ABC) methods perform Bayesian inference using noisy surrogate versions of the likelihood. Due to this noise, the ABC posterior is typically more dispersed than the true posterior \citep{sisson2018handbook}.

\subsection{Related methods} \label{sec:bias-lit-review}

\cite{biswas2024bounding} use couplings to assess the quality of approximate sampling methods. The idea is to sample a pair of coupled Markov chains $\smash{(X^{(t)}, Y^{(t)})_{t \ge 0}}$ with kernels $(P, Q)$ and stationary distributions $(\mu, \nu)$. In the idealized setting where the chains are stationary, for all $(B,I)$ it holds that
\begin{equation} \label{eqn:bias-cpl-bound}
    \wass_2^2(\mu, \nu) \le \mathbb{E} \left[ \frac{1}{I}\sum_{t=B}^{B+I-1}\big\lVert X^{(t)} - Y^{(t)} \big\rVert^2 \right].
\end{equation}
In practice, we discard the first $B$ iterations as burn-in and we estimate the coupling bound by averaging over $K$ replicates.

The method of \cite{biswas2024bounding} can only perform well if $(P,Q)$ are similar in a uniform sense. It additionally requires the user to carefully design a contractive coupling of $(P, Q)$. As we demonstrate in Section~\ref{sec:bias-numerical}, sensible couplings of $(P, Q)$ can still produce loose bounds, whereas any coupling that positively correlates the chains can reduce the variance of our proposed estimators.

\cite{huggins2020validated} derive computable upper bounds on $\wass_2^2(\mu, \nu)$ based on a series of worst-case theoretical bounds and importance sampling using $\nu$ as a proposal. \cite{dobson2021using} propose a coupling-based upper bound that is similar to~\eqref{eqn:bias-cpl-bound}, but incurs an additional term related to the rate of contraction of the kernel $Q$. Because \citet[Section~3.4]{biswas2024bounding} demonstrates that the method of \cite{huggins2020validated} deteriorates rapidly with increasing dimension and that the method of \cite{dobson2021using} produces a looser bound than~\eqref{eqn:bias-cpl-bound}, we do not compare with these methods in the sequel.

\subsection{Numerical illustrations} \label{sec:bias-numerical}

We illustrate the proposed methodology with various applications, comparing our method with the coupling-based bound of \cite{biswas2024bounding} and assessing the sharpness of all estimates against the tractable lower bound~\eqref{eqn:computable-lb}. Because the squared Wasserstein distance does not have a global upper bound, we instead provide the trace of the covariance $\tr(\cov_\mu(X))$ as a measure of scale,\footnote{$\wass_2^2(\mu, \nu) = \tr(\cov_\mu(X))$ when $\nu$ is a Dirac mass centered at the mean of $\mu$.}  that intuitively indicates a poor approximation~$\nu$. We defer additional experimental details to Appendix~\ref{app:bias-experiments}.

\subsubsection{Asymptotic bias of unadjusted MCMC algorithms} \label{sec:bias-langevin}

We estimate the asymptotic bias of two unadjusted MCMC algorithms, ULA and the OBABO discretization of the underdamped Langevin diffusion. The algorithms target $\mu = \mathcal{N}_d(0_d, \Sigma_d)$ with $(\Sigma_d)_{ij} = 0.5^{|i-j|}$ and use spherical Gaussian proposals with standard deviation $h = d^{-1/6}$ in various dimensions~$d$. The underdamped algorithm uses critical damping. This synthetic Gaussian setting presents us with a dual advantage: it allows us to compare estimators against the true squared Wasserstein distance, as well as to assess the sensitivity of estimators to the dynamics of each approximate MCMC algorithm, since both algorithms have identical Gaussian stationary distributions $\mu_h$ at identical step sizes $h,$ see Appendix~\ref{app:obabo-bias}.

We follow \citet[Section~2.2]{biswas2024bounding} and couple each unadjusted algorithm with its Metropolis-adjusted counterpart by CRNs, ULA with the Metropolis-adjusted Langevin algorithm \citep[MALA;][]{besag1994mala} and OBABO with the method of \cite{horowitz1991generalized}. We do not use the couplings to reduce the variance of the proposed estimators.

\begin{figure}[ht]
    \centering
    \includegraphics[width=0.7\textwidth]{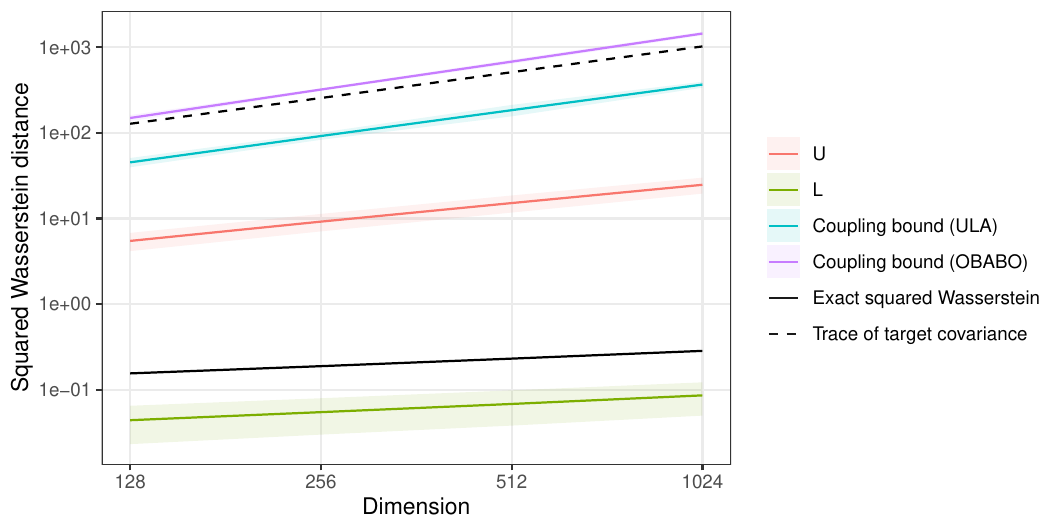}
    \caption{Asymptotic bias of unadjusted MCMC algorithms in increasing dimension, see Section~\ref{sec:bias-langevin} for details. The considered algorithms (ULA and OBABO) have identical stationary distributions. Solid lines represent empirical means, shaded areas represent two standard deviations.}
    \label{fig:bias-langevin}
\end{figure}

Figure~\ref{fig:bias-langevin} displays estimates of the asymptotic bias $\wass_2^2(\mu, \mu_h)$. The proposed estimators $\{U,L\}$ reveal that the asymptotic bias is small even in high dimensions and provide identical results for both approximate algorithms. In contrast, the coupling bound is at least an order of magnitude looser and performs significantly worse for OBABO than it does for ULA. We estimate that the coupling of ULA (resp. OBABO) could have reduced the variance of $U$ by a factor of~$2\times$ (resp. $1.1\times$).

This experiment highlights a limitation of the coupling bound. Although seemingly a reasonable default, coupling unadjusted MCMC algorithms with their Metropolized counterparts turns out to only be effective when the acceptance rate of the Metropolized algorithm is extremely high, i.e. the mixing is poor. For ULA coupled with MALA, we observe that the squared-distance between the chains increases by $\Theta(h^2 d)$ upon rejection in MALA, whereas the chains contract exponentially at rate $\Theta(h^2)$ upon acceptance. The equilibrium therefore lies at $\Theta(d)$ times the rejection rate, which is typically much larger than $\wass_2^2(\mu, \mu_h) = \Theta(h^2d)$ \citep{durmus2019high}. For OBABO coupled with the Horowitz method, the situation worsens because the Horowitz method reverses direction upon rejection; the persistent momentum then causes the chains to move away from each other for several iterations. In this experiment, the step size $h = d^{-1/6}$ ensures a small asymptotic bias and a relatively high acceptance rate of $\approx 70\%$, yet the coupling bound is still loose.

\subsubsection{Approximate inference for tall data} \label{sec:bias-talldata}

We assess the quality of various approximate inference methods for tall datasets \citep{bardenet2017tall}, where the number of observations is much larger than the number of covariates. We consider stochastic gradient Langevin dynamics \citep[SGLD;][]{welling2011sgld} subsampling $10\%$ of the data per iteration, SGLD with control variates \citep[SGLD-cv;][]{baker2019control} subsampling $1\%$ of the data per iteration, the Laplace approximation, and full-rank Gaussian VI \citep{kucukelbir2017automatic}. We compare these methods on Bayesian logistic regression models with the following datasets: Pima Indians (\citealp{smith1988using}; 768 observations, 8 covariates) and DS1 life sciences (\citealp{komarek2003fast}; 26733 observations, 10 covariates).

For parity across methods, and to reduce the variance, we compute all Wasserstein distance estimators based on the same coupled pairs of Markov chains targeting $(\mu, \nu).$ We target $\mu$ and optimization-based approximations $\nu$ with MALA and use CRN couplings, as in \citet[Section~4.1]{biswas2024bounding}. To make the implementation generic across different approximations, we use the proposed estimator $V$ based on splitting the available sample.

\begin{figure}[ht]
    \centering
    \includegraphics[width=\textwidth]{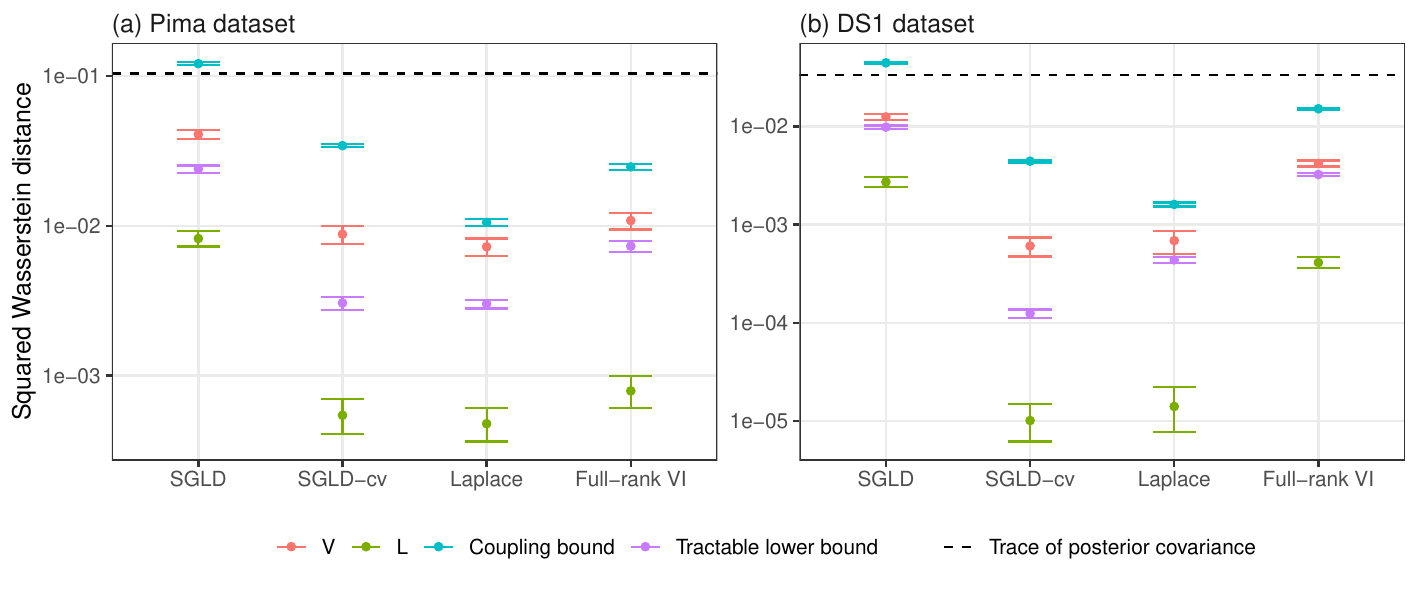}
    \caption{Quality of various approximate inference methods applied to Bayesian logistic regression models with various datasets, see Section~\ref{sec:bias-talldata} for details. Error bars represent approximate $95\%$ confidence intervals.}
    \label{fig:bias-logistic}
\end{figure}

Figure~\ref{fig:bias-logistic} displays estimates of the asymptotic bias of each approximate inference method. Consistent with the concentration of the posterior due to Bernstein-von-Mises limit, SGLD-cv and the Laplace approximation have the smallest biases. In contrast, SGLD overestimates the posterior variance due to noisy gradient estimates, whereas VI underestimates the posterior variance.

The proposed estimators accurately quantify the asymptotic bias: $V$ is often remarkably close to the tractable lower bound~\eqref{eqn:computable-lb}, which we expect to be tight due to the proximity of the model to its Bernstein-von-Mises limit. The coupling bound is uniformly looser: similarly to Section~\ref{sec:bias-langevin}, the issue is partly caused by the challenge in coupling MCMC algorithms that involve accept-reject decisions. We estimate that the coupling reduced the variance of $\{V,L\}$ by factors of up to $1.6\times$ for the Pima dataset and $2.2\times$ for the DS1 dataset.

Finally, sampling from the exact model $\mu$ with MALA becomes a significant bottleneck for datasets larger than the ones considered here. The proposed estimators can scale to larger datasets by amortizing the cost of sampling from $\mu$ using recent advances in exact MCMC algorithms based on subsampling \citep[e.g.][]{fearnhead2018piecewise, prado2024mhss}. However, because these algorithms have complex dynamics or parametrizations, it is less clear how one can couple them effectively.

\subsubsection{Approximate sampling for high-dimensional Bayesian linear regression} \label{sec:bias-half-t}

We consider a high-dimensional Bayesian linear regression model with half-t($\eta$) priors. \cite{johndrow2020scalable} developed exact and approximate Gibbs samplers for the case $\eta = 1$, corresponding to the horseshoe prior; \cite{biswas2021coupling-based} and \cite{biswas2024bounding} extended these samplers to degrees of freedom $\eta > 1$. We assess the asymptotic bias of such approximate Gibbs samplers with $\eta = 2$ on the Riboflavin dataset (\citealp{buhlmann2014high}; 71 observations, 4088 covariates).

This is a challenging scenario: the distributions we compare are high-dimensional, multimodal and heavy-tailed. This setting is also ideal for the coupling bound, since considerable effort has been spent on devising effective couplings for these samplers \citep{biswas2021coupling-based, biswas2024bounding}. We follow \cite{biswas2024bounding} and use CRN couplings between the approximate and exact Gibbs samplers. We also use the couplings to reduce the variance of our proposed estimators. Since we know that the exact model is the less dispersed distribution, we draw an additional set of samples from it to use throughout the experiment, and we use the estimator~$U$.

\begin{figure}[ht]
    \centering
    \includegraphics[width=0.7\textwidth]{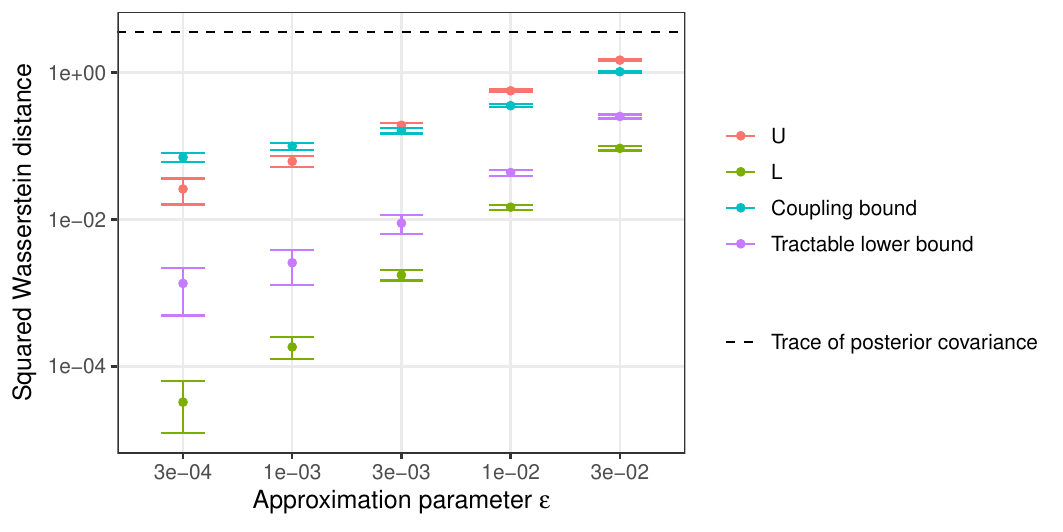}
    \caption{Asymptotic bias of approximate Gibbs sampler for high-dimensional linear regression model with half-t(2) prior, see Section~\ref{sec:bias-half-t}. Error bars represent approximate $95\%$ confidence intervals. The estimate of the tractable lower bound~\eqref{eqn:computable-lb} has a considerable positive bias for small $\varepsilon.$}
    \label{fig:bias-half-t}
\end{figure}

Figure~\ref{fig:bias-half-t} displays estimates of the asymptotic bias $\wass_2^2(\mu, \mu_\varepsilon)$ against the parameter $\varepsilon \ge 0$ that controls the quality of the approximation, where $\mu$ is the exact and $\mu_\varepsilon$ the approximate posterior marginal of the regression coefficients. The figure suggests that $\wass_2^2(\mu, \mu_\varepsilon) \approx \Theta(\varepsilon),$ which is consistent with the results of \cite{johndrow2020scalable} for the case $\eta=1$ and confirms that their recommended default of setting $\varepsilon$ as the reciprocal of the number of covariates ($\approx 2.5 \times 10^{-4}$ here) achieves a small asymptotic bias.

The experiment illustrates that the proposed estimators can be effective in complex problems of very high dimensionality. The estimator $U$ is competitive with the coupling bound and outperforms it for smaller values of $\varepsilon$. In fact, whereas our proposed estimators are guaranteed to be informative for all~$\varepsilon$, the coupling bound becomes uninformative as $\varepsilon \to 0$ because the CRN coupling is not uniformly contractive when $\varepsilon = 0$ \citep[Appendix B]{biswas2021coupling-based}. Nevertheless, because it reduces the variance of $\{U,L\}$ by a factor of $22\times$ for the smallest $\varepsilon$, the coupling appears crucial for controlling the variance of the proposed estimators when the true Wasserstein distance is small.

%% file: sections/5-mcmc-convergence.tex
\section{Assessing the convergence of MCMC algorithms} \label{sec:convergence}

MCMC algorithms undergo an initial warm-up phase wherein the time-marginals $(\marg{t})_{t\ge0}$ converge towards the stationary distribution $\marg{\infty}$. Assessing how quickly MCMC algorithms converge is of great importance to the researchers developing such methods, as well as to the practitioners using them. We propose here to estimate the convergence in squared Wasserstein distance $\wass_2^2(\marg{\infty}, \marg{t}),$ by post-processing the output of several parallel Markov chains using the estimators of Section~\ref{sec:main-results}.

\subsection{Methodology} \label{sec:convergence-method}

We simulate $2n$ replicate Markov chains up to a large time $T \gg 1$. We split the samples from $\marg{t}$ into equally weighted empirical measures $\{\marg{t}_n, \margbar{t}_n\}$ for all $t \ge 0.$ When $\marg{t}$ is more dispersed than $\marg{T}$, we estimate 
\begin{equation} \label{eqn:mcmc-simple-bounds}
   L(\margbar{T}_n, \marg{T}_n, \marg{t}_n) \lessapprox \wass_2^2(\marg{T}, \marg{t}) \lessapprox U(\margbar{T}_n, \marg{T}_n, \marg{t}_n).
\end{equation}
Conversely, we estimate $L(\margbar{t}_n, \marg{t}_n, \marg{T}_n) \lessapprox \wass_2^2(\marg{T}, \marg{t}) \lessapprox U(\margbar{t}_n, \marg{t}_n, \marg{T}_n)$ when $\marg{t}$ is less dispersed than $\marg{T}$.

The standard practice in MCMC is to use overdispersed initializations. Because the time-marginals $\marg{t}$ tend to gradually concentrate towards the stationary distribution $\marg{\infty}$ when the initialization is overdispersed, see Section~\ref{sec:convergence-overdisp}, in this setting we expect our estimators~\eqref{eqn:mcmc-simple-bounds} to reliably bound the convergence. We describe in Appendix~\ref{app:convergence-time-averaging} a reduced-variance methodology based on time-averaging that is tailored to overdispersed initializations.

We highlight three reasons why the proposed methodology is appealing. Firstly, the method closely approximates the true convergence rate, since by Theorem~\ref{thm:u}(ii) rates estimated by $U$ are loose by at most a factor of two. Secondly, the method is plug-in, so its performance is unaffected by how complex the implementation or dynamics of the MCMC kernel are. Finally, the method also applies to non-Markovian processes, so it can estimate the convergence of adaptive MCMC algorithms \citep{andrieu2008tutorial}.
Competing methods lack one or more of these properties, see Section~\ref{sec:convergence-couplings}.

The method can however be vulnerable to issues of pseudo-convergence, since it assumes that the replicate MCMC runs have converged and become stationary within the computing budget, so that $\wass_2^2(\marg{T}, \marg{t}) \approx \wass_2^2(\marg{\infty}, \marg{t})$. Convergence diagnostics \citep[e.g.][]{gorham2017measuring, margossian2024nestedRhat} can help check stationarity in practice. 

\subsection{On MCMC with an overdispersed initialization} \label{sec:convergence-overdisp}

The guideline of choosing overdispersed initializations dates back to the early days of parallel MCMC \citep{gelman1992inference}. Decades of experience suggest that overdispersed initializations facilitate both exploration and convergence diagnosis, with the intuition being that such initializations cause the time-marginals $\marg{t}$ to concentrate towards $\marg{\infty}$ over time. We verify this intuition in a stylized setting that is prototypical for many popular MCMC samplers.

\begin{proposition} \label{prop:cot-convergence-gauss}
    Let $(\marg{t})_{t\ge0}$ be the time-marginals of a Gaussian AR(1) process with a Gaussian initialization $\marg{0}$. If $\marg{0} \cotr \marg{\infty}$, then $\marg{t} \cotr \marg{\infty}$ for all $t \ge 0.$
\end{proposition}

\begin{remark} \label{rem:ar1}
Proposition~\ref{prop:cot-convergence-gauss} directly applies to discretizations of the overdamped Langevin diffusion. An extension of Proposition~\ref{prop:cot-convergence-gauss} holds for the position component of discretizations of the underdamped Langevin diffusion. In a small step-size asymptotic limit \citep{bou-rabee2010pathwise}, Proposition~\ref{prop:cot-convergence-gauss} applies to MALA and the method of \cite{horowitz1991generalized}, and similar insight \citep{roberts1997weak} can be expected to hold for the random walk Metropolis \citep[RWM;][]{tierney1994markov} algorithm. Finally, Proposition~\ref{prop:cot-convergence-gauss} applies to deterministic scan Gibbs samplers \citep{roberts1997updating}, and overdispersion persists in the sense of $\marg{t} \pcar \marg{\infty}$ for random scan Gibbs samplers. We provide verification in Appendix~\ref{app:checking-samplers-ar1}.
\end{remark}

For unimodal targets, Proposition~\ref{prop:cot-convergence-gauss} suggests that samplers initialized overdispersed should gradually concentrate towards their stationary distributions. Simulations with non-Gaussian unimodal targets in Appendix~\ref{app:mcmc-overdisp-start} support this insight. For multimodal targets, simulations in Appendix~\ref{app:mcmc-overdisp-start} suggest that the convergence happens in a similar way provided that the initialization is dispersed across all modes.

The choice of an appropriately overdispersed initialization should be guided by the target at hand. In Bayesian inference problems \citep[e.g.][]{gelman2013bayesian}, the prior is often a suitable initialization, because it tends to be less concentrated than the (target) posterior distribution. More generally, initializing from an overdispersed version of an approximation to the target is a sensible strategy: \cite{gelman1992inference} use heavy-tailed mixtures centered at the target modes; \cite{carpenter2017stan} use uniform distributions adapted to the length-scales of the target parameters.

\subsection{Related methods} \label{sec:convergence-couplings}

\cite{biswas2019estimating} use couplings to bound the convergence MCMC algorithms. Originally devised for 1-Wasserstein distances, we extend this method to $p$-Wasserstein distances of all orders $p \ge 1$ in Appendix~\ref{app:convergence-lagged-coupling}. With an appropriate choice of parameters, the method effectively amounts to repeatedly sampling coupled Markov chains $(\Xbar{t}, \X{t})_{t \ge 0}$ with initializations $(\Xbar{0}, \X{0}) \in \Gamma(\marg{\infty}, \marg{0})$ and marginal evolutions according to the Markov kernel of interest, then estimating the coupling inequality
\begin{equation} \label{eqn:convergence-coupling-bound}
    \wass_2^2 (\marg{\infty}, \marg{t}) \le \mathbb{E}\big[\| \Xbar{t} - \X{t}\|^2\big].
\end{equation}
using empirical averages. In our experiments, we estimate the idealized bound~\eqref{eqn:convergence-coupling-bound} based on independent initializations $(\Xbar{0}, \X{0})$.

\cite{johnson1996studying,sixta2024bounding} propose to estimate a looser version of the idealized bound~\eqref{eqn:convergence-coupling-bound} based on a rejection-sampling construction. Since this suffers from the curse of dimensionality, we do not compare with it in the sequel.

Coupling-based methods require the user to design and implement couplings that contract the chains $(\Xbar{t}, \X{t})$ quickly over time. As we demonstrate in Section~\ref{sec:convergence-numerics}, the availability of effective couplings is case-specific, and couplings can be sensitive to the dynamics and the parametrization of the MCMC algorithm at hand. In particular, we will see that Metropolis accept-reject steps, which are ubiquitously used to devise asymptotically exact MCMC algorithms, complicate the design of effective couplings in high dimensions \citep[see also][]{papp2024scalable}.

\subsection{Numerical illustrations} \label{sec:convergence-numerics}

We illustrate the proposed methodology with various moderate- to high-dimensional applications. We focus on the case of overdispersed initializations and use the reduced-variance method of Appendix~\ref{app:convergence-time-averaging}. We compare our method against the coupling bound of \cite{biswas2019estimating}, using state-of-the art couplings \citep[e.g.][]{heng2019unbiased, jacob2020unbiased, monmarche2021high} based on CRNs unless stated otherwise. As a default, we compute estimators based on $n = 1024$ replicates. We defer additional experimental details to Appendix~\ref{app:convergence-experiments}.

\subsubsection{Synthetic examples} \label{sec:conv-synthetic}

We consider synthetic examples with Gaussian target distributions. These allow us to directly assess the sharpness of our estimators against the exact squared Wasserstein distance $\wass_2^2(\marg{\infty}, \marg{t})$.

\begin{figure}[ht]
    \centering
    \includegraphics[width = \textwidth]{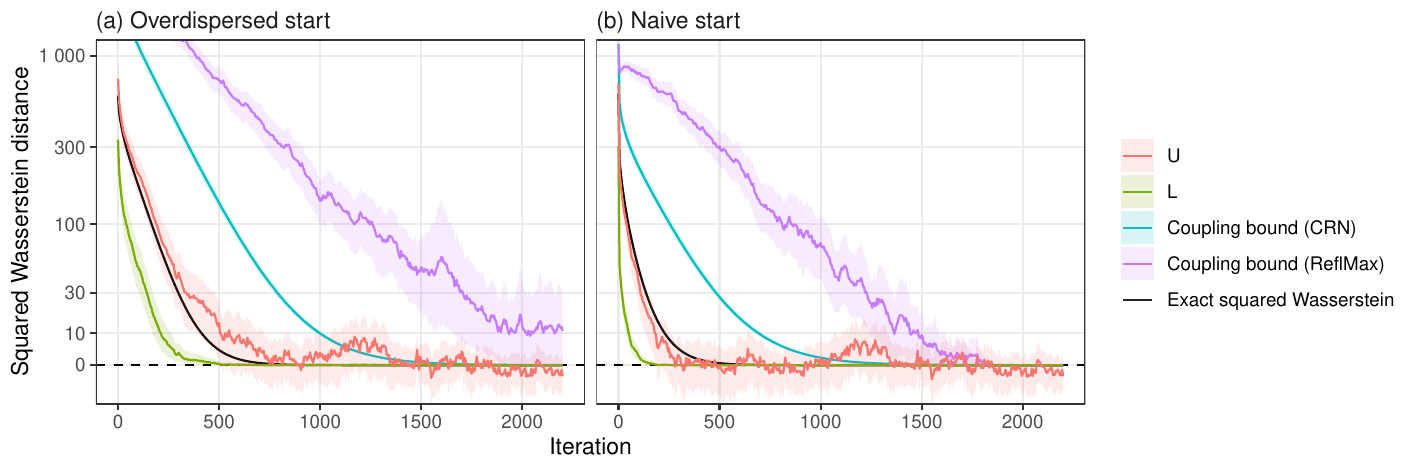}
    \caption{Convergence of a Gibbs sampler with various initializations, see Section~\ref{sec:conv-synthetic} for details. Shaded areas represent approximate $95\%$ confidence intervals.}
    \label{fig:gauss-gibbs}
\end{figure}

\paragraph{Gibbs sampler.} We target a periodic-boundary AR(1) process $\marg{\infty} = \mathcal{N}_d(0_d, \Sigma_d)$ with autocorrelation $\rho = 0.95$ in dimension $d = 50$ with a systematic scan Gibbs sampler. We consider two initializations: (a) a fully overdispersed start $\marg{0} = \mathcal{N}_d(0_d, 4\Sigma_d) \cotr \marg{\infty}$; (b) a naive start $\marg{0} = \mathcal{N}_d(0_d, \diag(\Sigma_d)) \cancel{\pcar} \marg{\infty}$ representing a mean-field approximation to $\marg{\infty}$.

Figure~\ref{fig:gauss-gibbs} displays estimates of the convergence of the Gibbs sampler with various methods. We see that the estimator~$U$ is conservative when the initialization is overdispersed and is robust to using naive initializations. Remarkably, in both cases, the true squared Wasserstein distance consistently falls within the confidence interval for~$U$; we speculate that this relates to the target having a few very large principal components which dominate the overall contribution to the Wasserstein distance. The proposed estimator~$L$ provides a sensible companion lower bound to~$U$. The sharpness of the coupling bound is highly dependent on the coupling used (coordinate-wise CRN or reflection-maximal, \citealp{jacob2020unbiased}), but even with the optimal Markovian CRN coupling this bound is relatively loose compared to the estimator~$U$.

\begin{figure}[ht]
    \centering
    \includegraphics[width=\textwidth]{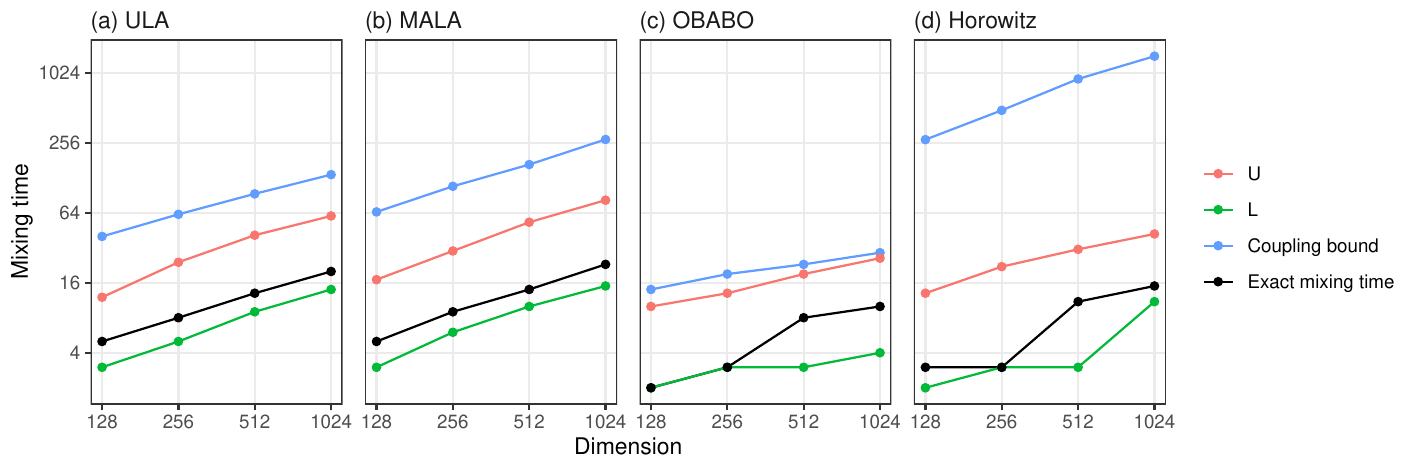}
    \caption{Mixing time of various adjusted and unadjusted MCMC algorithms, see Section~\ref{sec:conv-synthetic} for details.}
    \label{fig:langevin-mixing-time}
\end{figure}

\paragraph{Mixing time of Langevin algorithms.} We study the mixing time of MCMC algorithms based on the over- and underdamped Langevin diffusions. For each diffusion, we consider a discretization and its Metropolis-adjusted version: ULA and MALA in the overdamped case, the OBABO discretization and the \cite{horowitz1991generalized} method in the underdamped case.

We revisit the setting of Section~\ref{sec:bias-langevin}, targeting $\pi = \mathcal{N}_d(0_d, \Sigma_d)$ with $(\Sigma_d)_{ij} = 0.5^{|i-j|}$ and using spherical Gaussian proposals with standard deviation $h = d^{-1/6}$ in various dimensions~$d$. The target condition number is $\kappa \approx 9$ in all dimensions. The initialization $\marg{0} = \mathcal{N}_d(0_d, 3I_d)$ satisfies $\marg{0} \cotr \pi$.

While theoretical bounds must consider worst-case scenarios, the proposed estimators allow for comparisons to be drawn in the operational regime. Our scaling $h \sim d^{-1/6}$ is larger than ones suggested by non-asymptotic analyses \citep[e.g.][]{wu2022minimax}, but it is consistent with asymptotic analyses at stationarity \citep{roberts1998optimal} and at transience when converging ``inward" from the tails of the target \citep{christensen2005scaling}. The initialization ensures that we are in the latter regime.

Figure~\ref{fig:langevin-mixing-time} displays estimates of the mixing time $\tau_6 = \inf\{t : \wass_2^2(\marg{\infty}, \marg{t})\le 6\}$. The proposed estimators $\{U,L\}$ allow for meaningful comparisons to be drawn between algorithms: our findings are in line with the better scaling of the underdamped diffusion with the condition number of the target, as well as with the common belief that Metropolization slows down mixing. For the Horowitz method, the slow-down is due to the momentum reversals that occur whenever proposals are rejected, which cause the sampler to back-track. These momentum reversals are particularly problematic for the coupling bound, because they cause the coupled chains to drift apart when acceptances (resp. rejections) do not occur simultaneously. The coupling bound therefore wrongly suggests that the Horowitz method converges significantly slower than MALA.

\subsubsection{Stochastic volatility model} \label{sec:svm}

We consider the posterior distribution of a stochastic volatility model \citep[e.g.][]{liu2001monte} of dimension $d = 360$, a popular benchmark for MCMC algorithms. We target this model with various MCMC algorithms: the RWM algorithm with spherical Gaussian proposals and either (a) the optimal step size scaling (24\% acceptance rate; \citealp{roberts1997weak}) or (b) a smaller step size scaling (64\% acceptance rate); (c) MALA with spherical proposals and the optimal step size scaling (57\% acceptance rate; \citealp{roberts1998optimal}); (d) Fisher-MALA \citep{titsias2023optimal}, an adaptive MCMC algorithm that learns the proposal covariance structure together with the global scale parameter. The algorithms are initialized from the prior, which we verified to be substantially more dispersed than the target posterior distribution.

\begin{figure}[ht]
    \centering
    \includegraphics[width=\textwidth]{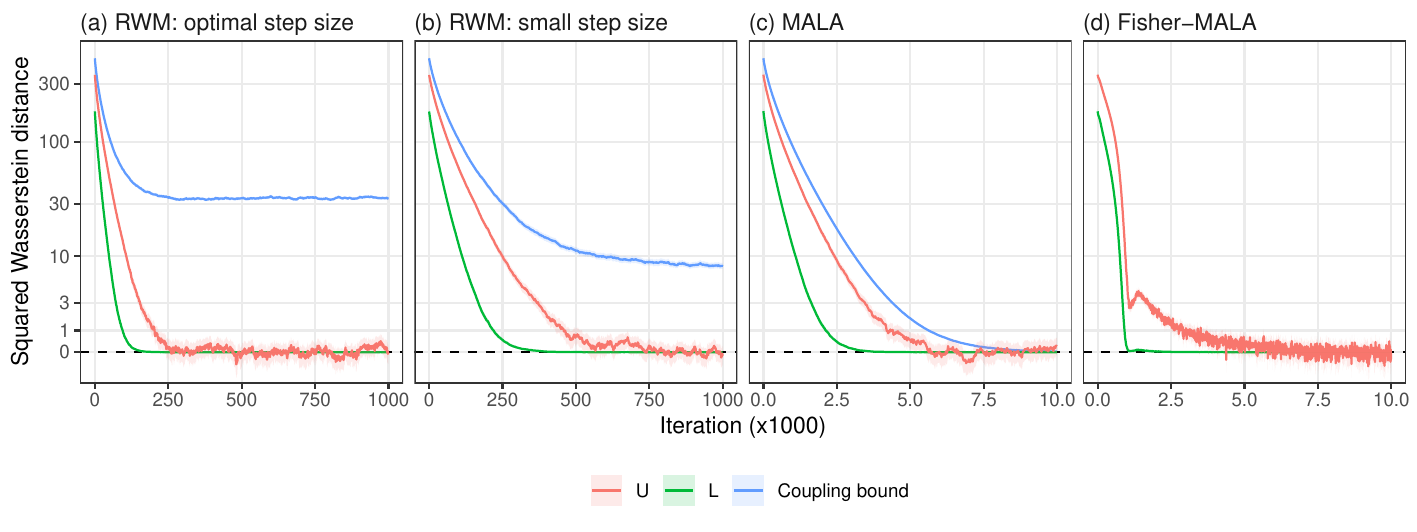}
    \caption{Convergence of various MCMC algorithms targeting a stochastic volatility model, see Section~\ref{sec:svm} for details. Shaded areas represent approximate $95\%$ confidence intervals. No coupling bound is computed for Fisher-MALA.}
    \label{fig:svm}
\end{figure}

Figure~\ref{fig:svm} displays estimates of the convergence rates of the considered algorithms. Based on the proposed estimators $\{U,L\}$, we see that the RWM converges faster with the larger step size, that MALA converges an order of magnitude faster than the RWM due to its use of more informative proposals, and that Fisher-MALA converges faster than MALA due to the adaptation. Notably, the initial convergence rate of Fisher-MALA is super-exponential due to rapid initial adaptation, which slows down to approximately exponential as the adaptation stabilizes. Consistent with the findings of \citet[Appendix~E]{titsias2023optimal}, Fisher-MALA appears not to converge monotonically in Wasserstein distance.

The proposed estimators provide such insights without requiring specific step sizes or that the underlying algorithm be Markovian. In contrast, the effectiveness of coupling-based estimators depends on the considered MCMC algorithm and its tuning parameters, with the considered reflection-maximal coupling of the RWM \citep{jacob2020unbiased} failing to produce informative bounds in this experiment. Furthermore, the coupling-based methodology of \cite{biswas2019estimating} is not yet applicable to non-Markovian adaptive algorithms.

In Appendix~\ref{app:svm}, we perform simulations with a more contractive but considerably more compute-intensive RWM coupling from \cite{papp2024scalable}, follow-up work from the \href{https://arxiv.org/abs/2203.11627v1}{initial version} of this manuscript, as well as with a Gaussian approximation to the model where the exact squared Wasserstein distance is available.

%% file: sections/6-discussion.tex
\section{Discussion} \label{sec:discussion}

Centering is a simple and effective strategy for obtaining informative estimates of the squared Euclidean 2-Wasserstein distance. We have demonstrated that our proposed centered estimators can often be viewed as approximate bounds on the squared Wasserstein distance, and have developed them into methodologies for assessing the quality approximate inference methods and the convergence of MCMC algorithms. The proposed methodologies compare favorably with coupling-based methods \citep{biswas2019estimating, biswas2024bounding}, while requiring considerably less expertise from the user.

We highlight a few methodological extensions that could be explored by further work.

\paragraph{Fast approximations.} Practitioners with access to GPUs could speed up the computation of the proposed estimators, at the cost of introducing a small degree of approximation, by using regularized versions of $\wass_2^2$ with a small regularization parameter \citep[e.g.][]{cuturi2013sinkhorn, genevay2018learning}. In settings like Section~\ref{sec:convergence} where multiple related optimal transport problems must be solved, progressive solvers based on successive warm starts \citep[e.g.][]{kassraie2024progressive} could speed up the computation further.

\paragraph{Importance-weighted empirical measures.} Importance sampling schemes \citep[e.g.][]{chopin2020introduction}, which approximate distributions by unequally weighted empirical measures, can provide an appealing alternative to MCMC in Bayesian computation applications such as those in  Section~\ref{sec:bias}. Exploring the use of importance-weighted empirical measures within our centered estimators is thus a promising direction for further work. We speculate that, as in Section~\ref{sec:bias}, the behavior of the proposed estimators would primarily depend on the effective sample sizes \citep{kong1992note} of the importance-weighted empirical measures.


%% file: sections/acknowledgements.tex
We gratefully acknowledge funding from the UK Engineering and Physical Sciences Research Council: TPP was supported by grants EP/S022252/1 (STOR-i Centre for Doctoral Training) and EP/R034710/1 (CoSInES) and CS was supported by grant EP/P033075/1. We are grateful to Aram-Alexandre Pooladian and Sinho Chewi for correcting an earlier proof of Proposition~\ref{prop:cot-cases}(i).

%% file: sections/appendix.tex
\section{Analysis for Sections~\ref{sec:empirical} and~\ref{sec:main-results}} \label{app:proofs}

It will be convenient to consider the Wasserstein distance of general order $p \ge 1$, defined through its $p$-th power as
\begin{equation*}
    \wass_p^p(\mu, \nu) = \inf_{\pi \in \Gamma(\mu, \nu)} \int \|x-y\|^p \mathrm{d}\pi(x,y) = \inf_{X\sim \mu, Y\sim \nu} \mathbb{E} \left[ \|X-Y\|^p \right],
\end{equation*}
where $\Gamma(\mu,\nu)$ is the set of all joint distributions $\pi$ with marginals ($\mu, \nu$). This has the Kantorovich dual
\begin{equation*}
\begin{gathered}
    \wass_p^p(\mu, \nu) = \sup_{(\phi, \psi) \in \Phi(\mu, \nu)} \int \phi(x) \mathrm{d}\mu(x) + \int \psi(y) \mathrm{d}\nu(y),\\
    \Phi(\mu, \nu) = \{(\phi, \psi) \in L_1(\mu) \times L_1(\nu) \mid \phi(x) + \psi(y) \le \|x - y\|^p ,\enskip \forall x, y \},
\end{gathered}
\end{equation*}
with an optimal solution $(\phi_{\mu, \nu}, \psi_{\mu, \nu}).$ We implicitly assume that $\mathbb{E}_\mu[\|X\|^p]< \infty$ and $\mathbb{E}_\nu[\|Y\|^p] < \infty$ whenever this distance is in use.

Recall that we have drawn independent samples $X_{1:n}, \bar{X}_{1:n} \iid \mu$ and $Y_{1:n}, \bar{Y}_{1:n} \iid \nu$ and defined the empirical measures 
\begin{equation*}
    \mu_n = \frac{1}{n}\sum_{i=1}^n \delta_{X_i},
    \enskip
    \bar\mu_n = \frac{1}{n}\sum_{i=1}^n \delta_{\bar X_i},
    \enskip
    \nu_n = \frac{1}{n}\sum_{i=1}^n \delta_{Y_i},
    \enskip
    \bar\nu_n = \frac{1}{n}\sum_{i=1}^n \delta_{\bar Y_i}.
\end{equation*}

\subsection{Bias of estimators} \label{app:pf-main-results}

\subsubsection{Plug-in estimator} \label{app:pf-plugin}

Lemma~\ref{lemma:plugin-bias} shows that the plug-in estimator of $\wass_p^p$ has a non-negative bias. 

\begin{lemma} \label{lemma:plugin-bias}
It holds that $\mathbb{E}\left[ \wass_p^p(\mu_n,\nu_n) \right] \ge \mathbb{E}\left[ \wass_p^p(\mu,\nu_n) \right] \ge \wass_p^p(\mu, \nu).$
\end{lemma}
\begin{proof}
   We prove that $\mathbb{E}\left[ \wass_p^p(\mu_n,\nu_n) \right] \ge \wass_p^p(\mu, \nu).$ Since $L_1(\nu) \subset L_1(\nu_n)$ it holds that $\Phi(\mu, \nu) \subset \Phi(\mu_n, \nu_n),$ therefore 
   \begin{equation*}
       \wass_p^p(\mu_n, \nu_n) = \sup_{(\phi, \psi) \in \Phi(\mu_n, \nu_n)} \int \phi \mathrm{d} \mu_n +  \int \psi \mathrm{d} \nu_n \ge \sup_{(\phi, \psi) \in \Phi(\mu, \nu)} \int \phi \mathrm{d} \mu_n +  \int \psi \mathrm{d} \nu_n \ge \int \phi_{\mu, \nu} \mathrm{d} \mu_n +  \int \psi_{\mu, \nu} \mathrm{d} \nu_n.
   \end{equation*}
   It follows that
   \begin{equation*}
       \mathbb{E}[\wass_p^p(\mu_n, \nu_n)] \ge \mathbb{E}\left[ \int \phi_{\mu, \nu} \mathrm{d} \mu_n +  \int \psi_{\mu, \nu} \mathrm{d} \nu_n \right] = \int \phi_{\mu, \nu} \mathrm{d} \mu +  \int \psi_{\mu, \nu} \mathrm{d} \nu = \wass_p^p(\mu, \nu),
   \end{equation*} 
   as claimed. The other inequalities follow by similar arguments, using in turn that $\mathbb{E}_{\mu_n}[\smallint \phi_{\mu, \nu_n} \mathrm{d} \mu_n] = \smallint \phi_{\mu, \nu_n} \mathrm{d} \mu$ and that $\mathbb{E}_{\nu_n}[\smallint \phi_{\mu, \nu} \mathrm{d} \nu_n] = \smallint \phi_{\mu, \nu} \mathrm{d} \nu$.
\end{proof}

Lemma~\ref{lemma:plugin-decay} shows that the bias of the plug-in estimator of $\wass_p^p$ decreases with the sample size.

\begin{lemma} \label{lemma:plugin-decay}
It holds that $\mathbb{E}\left[ \wass_p^p(\mu_{n-1},\nu_{n-1}) \right] \ge \mathbb{E}\left[ \wass_p^p(\mu_{n},\nu_{n}) \right].$
\end{lemma}
\begin{proof}
We define the leave-one-out empirical measures $\mu_{-i} = \frac{1}{n-1}\sum_{j\in[n]\setminus i} \delta_{X_j}$ and $\nu_{-i} = \frac{1}{n-1}\sum_{j\in[n]\setminus i}  \delta_{Y_j}.$ Using Kantorovich duality, 
    \begin{align*}
    \wass_p^p(\mu_{n}, \nu_{n}) 
    	&= \int \phi_{\mu_{n}, \nu_{n}}\mathrm{d}\mu_{n} + \int\psi_{\mu_{n}, \nu_{n}}\mathrm{d}\nu_{n}\\
    	&= \int \phi_{\mu_{n}, \nu_{n}} \left(\frac{1}{n}\sum_{i=1}^{n}\mathrm{d}\mu_{-i}\right) + \int\psi_{\mu_{n}, \nu_{n}} \left(\frac{1}{n}\sum_{i=1}^{n}\mathrm{d}\nu_{-i}\right)\\
    	&= \frac{1}{n}\sum_{i=1}^{n}\left(\int \phi_{\mu_{n}, \nu_{n}} \mathrm{d} \mu_{-i} + \int \psi_{\mu_{n}, \nu_{n}} \mathrm{d} \nu_{-i} \right)\\        
    	&\le \frac{1}{n}\sum_{i=1}^{n} \sup_{(\phi, \psi) \in \Phi(\mu_{-i}, \nu_{-i})} \int \phi \mathrm{d} \mu_{-i} + \int \psi \mathrm{d} \nu_{-i} = \frac{1}{n}\sum_{i=1}^{n}\wass_p^p(\mu_{-i}, \nu_{-i}),
    \end{align*}
where finally we used that $(\phi_{\mu_{n}, \nu_{n}}, \psi_{\mu_{n}, \nu_{n}}) \in \Phi(\mu_{n}, \nu_{n}) \subset \Phi(\mu_{-i}, \nu_{-i}),$ then Kantorovich duality. The claimed result follows by taking expectations and using that $\mathbb{E}[\wass_p^p(\mu_{-i}, \nu_{-i})] = \mathbb{E}[\wass_p^p(\mu_{n-1}, \nu_{n-1})]$ for all~$i$.
\end{proof}
\subsubsection{Proof of Theorem~\ref{thm:u}(i)}

The proof relies on a few standard results, which we recall without proof. For a convex $\varphi:\mathbb{R}^d \to \mathbb{R},$ we let $\varphi^*(x) = \sup_y \{x^\top y - \varphi(y)\}$ be its Legendre transform, which is convex and satisfies $\varphi^{**} = \varphi$. We say that $\varphi$ is $m$-strongly convex for $m>0$ if and only if $f(x) = \varphi(x) - m\|x\|^2/2$ is convex.

\begin{lemma}[{\citealp[Duality between smoothness and strong convexity;][]{zhou2018fenchel}}] \label{lemma:lip-grad-dual}
    Let $\varphi:\mathbb{R}^d \to \mathbb{R}$ be convex and let $L > 0$. Then, $\lip{\nabla \varphi} \le L$ if and only if $\varphi^*$ is $(1/L)$-strongly convex.
\end{lemma}

\begin{lemma}[{\citealp[Brenier's theorem;][]{mccann1995existence}}] \label{lemma:brenier}
     Let $\mu,\nu \in \mathcal{P}(\mathbb{R}^d)$ satisfy Assumption~\ref{assumption:cts}. Then,
    \begin{equation*}
        \wass_2^2(\mu, \nu) = \mathbb{E}_\mu[\|X - T_{\mu, \nu}(X)\|^2] = \mathbb{E}_\nu[\|T_{\nu, \mu}(Y) - Y\|^2],
    \end{equation*}
    where $T_{\mu, \nu}$ and $T_{\nu, \mu}$ are push-forward maps ($T_{\mu, \nu}\# \mu = \nu$, $T_{\nu, \mu}\# \nu = \mu$). Furthermore, the maps are uniquely determined by $T_{\mu, \nu} = \nabla \varphi_{\mu, \nu}$ and $T_{\nu, \mu} = \nabla \varphi_{\nu, \mu}$ where $\varphi_{\mu, \nu}, \varphi_{\nu, \mu}:\mathbb{R}^d \to \mathbb{R}$ are convex and conjugate ($\varphi_{\nu, \mu} = \varphi_{\mu, \nu}^*$).
\end{lemma}

\begin{lemma} $\wass_2^2(\mu_n, \nu_n) = \min_\sigma \frac{1}{n}\sum_{i=1}^n \| X_i - Y_{\sigma(i)} \|^2$ over all permutations $\sigma$.
\end{lemma}

We proceed with the proof of Theorem~\ref{thm:u}(i). Since we have assumed that $\lip{T_{\nu, \mu}} = \lip{\nabla \varphi_{\nu, \mu}} \le 1,$ it follows that $\varphi_{\nu, \mu}^* = \varphi_{\mu, \nu}$ is 1-strongly convex. Therefore, $T_{\mu, \nu} = \mathrm{id} + \nabla f$, where $f(x) = \varphi_{\mu, \nu}(x) - \|x\|^2/2$ is convex. For all $(x, \bar x),$ we therefore have that
\begin{equation} \label{eqn:thm-1-pf-1}
    \begin{aligned}
    \| T_{\mu, \nu}(x) - \bar x \|^2 
    &= \| T_{\mu, \nu}(x) - x\|^2 + 2\nabla f(x)^\top(x - \bar x) + \| x - \bar x \|^2 \\
    &\ge \| T_{\mu, \nu}(x) - x\|^2 + 2\{ f(x) - f(\bar x)\} + \| x - \bar x \|^2,
    \end{aligned}
\end{equation}
where finally we used the convexity of $f$.

Now, without loss of generality, we set $Y_i = T_{\mu, \nu}(X_i)$. By the primal formulation,
\begin{align*}
    \mathbb{E}\left[ \wass_2^2(\bar \mu_n, \nu_n)\right] &= \mathbb{E}\left[ \min_\sigma \frac{1}{n}\sum_{i=1}^n \| \bar X_i - T_{\mu, \nu}(X_{\sigma(i)}) \|^2 \right] \\
    &\ge \mathbb{E}\left[ \min_\sigma \frac{1}{n}\sum_{i=1}^n \left( \| \bar X_i - X_{\sigma(i)} \|^2 + 2\left\{f(\bar X_i) - f(X_{\sigma(i)})\right\} + \| X_{\sigma(i)} - T_{\mu, \nu}(X_{\sigma(i)}) \|^2\right) \right]\\
    &= \mathbb{E}\left[ \min_\sigma \frac{1}{n}\sum_{i=1}^n \| \bar X_i - X_{\sigma(i)} \|^2 + \frac{1}{n}\sum_{i=1}^n \| X_{i} - T_{\mu, \nu}(X_{i}) \|^2 \right]\\
    &= \mathbb{E}\left[ \wass_2^2(\bar \mu_n, \mu_n)\right] + \wass_2^2(\mu, \nu),
\end{align*}
where we used~\eqref{eqn:thm-1-pf-1} for the second line, that $\sigma$ is a permutation and that $X_i, \bar X_i \sim \mu$ for the third, and the primal formulation for the last. This concludes the proof.

\subsubsection{Proof of Theorem~\ref{thm:u}(ii)}

Using the primal formulation, we have that
\begin{equation*}
    \begin{aligned}
    \mathbb{E}\left[\wass_2^2(\bar\mu_n, \nu_n) - \wass_2^2(\bar\mu_n, \mu_n)\right]
    &= \mathbb{E}\big[\|Y\|^2-\|X\|^2\big] - 2\mathbb{E} \left[\max_{\sigma} \frac{1}{n}\sum_{i=1}^n Y_i^\top  \bar X_{\sigma(i)} - \max_{\sigma}\frac{1}{n}\sum_{i=1}^n X_i^\top \bar{X}_{\sigma(i)} \right] \nonumber \\
    & =: \Circled{1} - \Circled{2},
\end{aligned}
\end{equation*}
where $(X,Y) \sim (\mu,\nu)$ and where the maxima are over all permutations $\sigma$.

\textbf{Term \Circled{1}.} By the Minkowski inequality, $\big|\mathbb{E}[\|Y\|^2]^{1/2} - \mathbb{E}[\|X\|^2]^{1/2} \big| \le \inf_{(X,Y) \in \Gamma(\mu, \nu)}\mathbb{E}[\|Y - X\|^2]^{1/2} = \wass_2(\mu, \nu)$. It follows that 
\begin{equation*}
    \left| \mathbb{E}\big[\|Y\|^2-\|X\|^2\big] \right| \le  \wass_2(\mu, \nu) \left( \mathbb{E}\big[\|X\|^2\big]^{1/2} + \mathbb{E}\big[\|Y\|^2\big]^{1/2} \right).
\end{equation*}

\textbf{Term \Circled{2}.} Without loss of generality, we choose to sample the pairs $(X_i, Y_i) \sim (\mu, \nu)$ i.i.d. from the optimal coupling. We have that
\begin{align*}
   \frac{1}{2} \left|\Circled{2}\right|
    &\le \left| \mathbb{E} \left[ \max_{\sigma}\frac{1}{n}\sum_{i=1}^n (Y_i - X_i)^\top \bar{X}_{\sigma(i)} \right] \right| \tag{$\max$ is convex}\\
    &\le \mathbb{E} \left[ \max_{\sigma} \left(\frac{1}{n}\sum_{i=1}^n \|Y_i - X_i\|^2 \right)^{1/2} \left( \frac{1}{n}\sum_{i=1}^n\| \bar{X}_{\sigma(i)} \|^2 \right)^{1/2} \right] \tag{Cauchy-Schwarz}\\
    &= \mathbb{E} \left[ \left(\frac{1}{n}\sum_{i=1}^n \|Y_i - X_i \|^2 \right)^{1/2} \left( \frac{1}{n}\sum_{i=1}^n\| \bar{X}_i \|^2 \right)^{1/2} \right] \tag{$\sum_i \|\bar{X}_{\sigma(i)}\|^2 = \sum_i \|\bar{X}_i\|^2$ } \\
    &\le \mathbb{E} \left[\frac{1}{n}\sum_{i=1}^n \| Y_i - X_i \|^2 \right]^{1/2} \mathbb{E}\left[\frac{1}{n}\sum_{i=1}^n\| \bar{X}_i \|^2 \right]^{1/2} \tag{Cauchy-Schwarz}\\
    &= \wass_2(\mu, \nu) \mathbb{E} \big[\| X\|^2\big]^{1/2}. \tag{couplings $(X_i, Y_i)$ are optimal}
\end{align*}
Therefore,
\begin{equation*}
\left| \mathbb{E}\left[\wass_2^2(\bar\mu_n, \nu_n) - \wass_2^2(\bar\mu_n, \mu_n)\right] \right| \le \left|\Circled{1}\right| + \left|\Circled{2}\right| \le \wass_2(\mu, \nu) \left( 3\mathbb{E} \big[\| X\|^2\big]^{1/2} + \mathbb{E} \big[ \| Y \|^2 \big]^{1/2}\right),
\end{equation*}
which concludes the proof.

\subsubsection{Proof of Theorem~\ref{thm:u}(iii)}

Let $\{\mu^c, \nu^c\}$ be versions of $\{\mu, \nu\}$ with expectations $0$, and let $\{\bar\mu_n^c, \mu_n^c, \nu_n^c\}$ be the analogous transformations of $\{\bar\mu_n, \mu_n, \nu_n\}$. From \citet[Section~2]{panaretos2019statistical}, it holds that
\begin{align*}
    \wass_2^2(\mu, \nu) &= \lVert \mathbb{E}_\mu[X] - \mathbb{E}_\nu[Y] \rVert^2 + \wass_2^2(\mu^c, \nu^c),\\
    \mathbb{E}[\wass_2^2(\bar\mu_n, \nu_n)] &= \lVert \mathbb{E}_\mu[X] - \mathbb{E}_\nu[Y] \rVert^2 + \mathbb{E}[\wass_2^2(\bar\mu_n^c, \nu_n^c)],\\
    \mathbb{E}[\wass_2^2(\bar\mu_n, \mu_n)] &= \mathbb{E}[\wass_2^2(\bar\mu_n^c, \mu_n^c)].
\end{align*}
It follows that
\begin{equation*}
    \mathbb{E}[U(\bar\mu_n, \mu_n, \nu_n)] - \wass_2^2(\mu, \nu) = \mathbb{E}[U(\bar\mu_n^c, \mu_n^c, \nu_n^c)] - \wass_2^2(\mu^c, \nu^c),
\end{equation*}
hence the difference is location-free, as claimed.

\subsubsection{Proof of Proposition~\ref{prop:l}}

By the Jensen and triangle inequalities,
\begin{equation} \label{eq:lower-bound-proof-1}
    \left| \mathbb{E}\left[\wass_2(\bar\mu_n, \nu_n) - \wass_2(\bar\mu_n, \mu_n)\right] \right| \le \mathbb{E}\left[ \left|\wass_2(\bar\mu_n, \nu_n) - \wass_2(\bar\mu_n, \mu_n)\right| \right] \le \mathbb{E}\left[\wass_2(\mu_n, \nu_n)\right].
\end{equation}

Now, using the linearity of the expectation, without loss of generality (without changing the left-hand-side of \eqref{eq:lower-bound-proof-1}) we choose to instead sample $(X_i, Y_i) \sim (\mu,\nu)$ i.i.d. from the optimal coupling. By Jensen's inequality and the primal formulation,
\begin{equation} \label{eq:lower-bound-proof-2}
\mathbb{E}[\wass_2(\mu_n, \nu_n)] \le \mathbb{E}\left[\wass_2^2(\mu_n, \nu_n)\right]^{1/2}
    \le \mathbb{E}\left[\frac{1}{n}\sum_{i=1}^n \| X_{i} - Y_{i} \|^2 \right]^{1/2} 
    = \wass_2(\mu, \nu).
\end{equation}
Combining inequalities \eqref{eq:lower-bound-proof-1} and \eqref{eq:lower-bound-proof-2} completes the proof.

\subsection{Overdispersion conditions}

\subsubsection{Proof of Proposition~\ref{prop:cot-cases}}

We first require the following characterization of $\cotr$ and we recall an auxiliary result.

\begin{lemma} \label{lemma:interpret-cot}
The following claims are equivalent: $(i)$~$\nu \cotr \mu$; $(ii)$~$\| T_{\nu, \mu} \|_\textup{Lip} \le 1$; $(iii)$~$\varphi_{\mu, \nu}$ is 1-strongly convex; $(iv)$~$\nabla^2 \varphi_{\nu, \mu} \preceq I_d$ uniformly; $(v)$~$\nabla^2 \varphi_{\mu, \nu} \succeq I_d$ uniformly.
\end{lemma}
\begin{proof}
Since the Brenier potentials $(\varphi_{\mu, \nu}, \varphi_{\nu, \mu})$ are convex, by Alexandroff's theorem their gradients and Hessians exist almost-everywhere.

The equivalence $(i) \iff (ii)$ follows by definition. The equivalence $(ii) \iff (iii)$ follows from the duality of smoothness and strong convexity. The equivalence $(ii) \iff (iv)$ is shown in \citet[Theorem~2.1.6]{nesterov2004introductory}. The equivalence $(iii) \iff (v)$ is shown in \citet[Theorem~2.1.11]{nesterov2004introductory}. Therefore, all claims are equivalent.
\end{proof}

\begin{lemma}[{\citealp[Corollary~3.5]{lawson2001geometric}}] \label{lemma:geom-mean-lowener}
    Let $M,N\in \mathbb{R}^{d\times d}$ be positive definite matrices. Define $M^{-1} \# N := M^{-1/2} ( M^{1/2} N M^{1/2} )^{1/2} M^{-1/2}.$ Then, it holds that $ I \preceq M^{-1} \# N$ if and only if $M \preceq N$.
\end{lemma}

We proceed to the main proof. Since $\cotr$ is location-free, without loss of generality we let $\mathbb{E}_\mu[X] = \mathbb{E}_\nu[Y] = 0$.

\paragraph{Claim (i).} By \citet[Remark~2.31]{peyre2019computational}, the Brenier potential from $\mu = \mathcal{N}(0, \Sigma_\mu)$ to $\nu = \mathcal{N}(0, \Sigma_\nu)$ is $\varphi_{\mu, \nu}(x) =  x^\top  (\Sigma_\mu^{-1}\# \Sigma_\nu )x/2.$ By Lemma~\ref{lemma:interpret-cot}, we have that
\begin{equation*}
    \nu \cotr \mu \iff I \preceq \nabla^2 \varphi_{\mu, \nu} \text{ uniformly} \iff I \preceq \Sigma_\mu^{-1}\# \Sigma_\nu  \iff \Sigma_\mu \preceq \Sigma_\nu,
\end{equation*}
where finally we used Lemma~\ref{lemma:geom-mean-lowener}. This concludes the proof of the claim.

\paragraph{Claim (ii).} Let $\mu,\nu \in \mathcal{P}(\mathbb{R}^d)$ be spherically symmetric and let $\mathcal{S}^{d-1}$ be the unit sphere. Any $X \sim \mu$ can be written as $X = R_\mu U_\mu$ in terms of an angular component $U_\mu \sim \unif(\mathcal{S}^{d-1})$ and an independent radial component $R_\mu \sim r_\mu \in \mathcal{P}((0, \infty))$. Similarly, so can $Y = R_\nu U_\nu \sim \nu$. Now, $\mathbb{E} [\|R_\mu U_\mu -  R_\nu U_\nu \|^2] \ge \mathbb{E} [(R_\mu -  R_\nu)^2].$ Since the lower bound is attained by the coupling
\begin{equation*}
    (X,Y) = \big(F_{ r_\mu}^{-1}(U_1) U, F_{r_\nu}^{-1}(U_1) U\big) \sim (\mu, \nu),
\end{equation*}
where $U_1 \sim \unif([0,1])$ and $U \sim \unif(\mathcal{S}^{d-1})$, this coupling must be optimal. The optimal transport map is therefore
\begin{equation*}
    T_{\nu, \mu}(x) =  \big(F_{ r_\mu}^{-1} \circ F_{r_\nu}\big)(\|x\|) \cdot \frac{x}{\|x\|},
\end{equation*}
and so $\lip{T_{\nu, \mu}} \le 1$ if and only if $\lip{F_{ r_\mu}^{-1} \circ F_{r_\nu}} \le 1,$ as claimed.

\paragraph{Claim (iii).} Let $\mu,\nu \in \mathcal{P}(\mathbb{R}^d)$ be product measures, say $\mu = \otimes_{i=1}^d \mu^i$ and $\nu = \otimes_{i=1}^d \nu^i$. By the tensorization property of the squared Wasserstein distance, the optimal transport map is
\begin{equation*}
    T_{\nu, \mu}(x) = \left(T_{\nu^1, \mu^1}(x_1), \dots, T_{\nu^d, \mu^d}(x_d)\right)^\top,
\end{equation*}
where $T_{\nu^i, \mu^i} = F_{\mu^i}^{-1} \circ F_{\nu^i}$. Therefore, $\lip{T_{\nu, \mu}} \le 1$ if and only if $\lip{T_{\nu^i, \mu^i}} \le 1$ for all $i$, as claimed.

\paragraph{Claim (iv).} This is lifted from \citet[Theorem~13]{chewi-pooladian2023entropic}.

\subsubsection{On Example~\ref{ex:debias-n=1}} \label{app:debias-n=1}

\paragraph{Deriving the result.}

The inequality $\mathbb{E} \left[ U(\bar\mu_1, \mu_1, \nu_1)\right] \ge \wass_2^2(\mu, \nu)$ is equivalent to
\begin{equation*}
    \mathbb{E}[ \|\bar X - Y\|^2 - \|\bar X - X\|^2 ] \ge \inf_{(X,Y) \sim (\mu, \nu)} \mathbb{E}[ \|Y - X\|^2 ],
\end{equation*}
where $\bar X \sim \mu$ is independent of $(X,Y) \sim (\mu, \nu)$. Rearranging, this is equivalent to
\begin{equation*}
\sup_{(X,Y)\sim (\mu,\nu)} 2\mathbb{E}\left[ X^\top Y - \mathbb{E}[X]^\top \mathbb{E}[Y] \right] \ge 2\mathbb{E}\left[ \|X\|^2 -  \mathbb{E}[X]^\top \mathbb{E}[X] \right].
\end{equation*}
Recognizing the outer expectations as $\tr(\cov(X,Y))$ and $\tr(\var(X))$ provides the result of Example~\ref{ex:debias-n=1}.

\paragraph{Partial closure under mixtures.} Let $\nu = \sum_k p_k \nu^k$ be a mixture. By Jensen's inequality and the linearity of the expectation, it holds that
\begin{equation*}
    \sup_{(X,Y)\sim (\mu,\nu)} \tr(\cov(X,Y)) \ge \sum_{k} p_k \sup_{(X,Y_k)\sim (\mu,\nu^k)} \tr(\cov(X,Y_k)).
\end{equation*}
So, if $\sup \tr(\cov(X,Y_k)) \ge \tr(\var(X))$ for all $k,$ then $\sup \tr(\cov(X,Y)) \ge \tr(\var(X)).$ In other words, the relation of Example~\ref{ex:debias-n=1} is partially closed under mixtures.

\paragraph{Relation to convex ordering.} The convex ordering $\nu \gtrcvx \mu$ states that $\mathbb{E}_\nu[f(Y)] \ge \mathbb{E}_\mu[f(X)]$ for any convex $f$ for which the expectations are well-defined. Strassen's martingale coupling theorem \citep{strassen1965existence} states that this is equivalent to the existence of coupling $(X,Y) \sim (\mu, \nu)$ such that $\mathbb{E}[Y \mid X] = X$.

Now, suppose that that a convex ordering holds between versions of $\mu$ and $\nu$ which are centered at~$0$, i.e. that there exists a coupling of $(X,Y) \sim (\mu, \nu)$ such that $\mathbb{E}[Y - \mathbb{E}[Y] \mid X] = X - \mathbb{E}[X]$. Under this coupling, $\tr(\cov(X,Y)) = \tr(\cov(X,X)) = \tr(\var(X)),$ so the condition of Example~\ref{ex:debias-n=1} is satisfied.

\subsubsection{On Example~\ref{ex:debias-d=1}} \label{app:debias-d=1}

The asymptotic result of Example~\ref{ex:debias-d=1} is a consequence of Proposition~\ref{prop:rate-1d}. We require Lemma~\ref{lemma:w2sq-bias-1d}, which provides a tractable formula for the bias of the plug-in estimator in the one-dimensional setting.

\begin{lemma} \label{lemma:w2sq-bias-1d}
Let $(\mu,\nu)$ be one-dimensional measures with inverse-CDFs $(G, H)$, and let $U_{(1:n)}$ be the order statistics of $U_{1:n} \iid \unif(0,1)$. Then,
\begin{equation*}
\mathbb{E}[\wass_2^2(\mu_n, \nu_n)] - \wass_2^2(\mu, \nu) = \frac{2}{n} \sum_{i=1}^n \cov\left( G(U_{(i)}), H(U_{(i)}) \right).
\end{equation*}
\end{lemma}
\begin{proof}
Since $\mathbb{E}[\wass_2^2(\mu_n, \nu_n)] = \frac{1}{n} \mathbb{E}\big[\sum_{i=1}^n(X_{(i)} - Y_{(i)})^2\big]$ and since $X_{1:n}$ is independent of $Y_{1:n}$, it holds that
\begin{align*}
    \mathbb{E}[\wass_2^2(\mu_n , \nu_n)] 
    &= \mathbb{E}\left[X_1^2 + Y_1^2\right] - \frac{2}{n}\sum_{i=1}^n \mathbb{E}\left[  X_{(i)} Y_{(i)} \right] = \mathbb{E}\left[X_1^2 + Y_1^2\right] - \frac{2}{n}\sum_{i=1}^n \mathbb{E}\left[ X_{(i)}\right] \mathbb{E}\left[ Y_{(i)} \right]\\
    &= \mathbb{E}\left[X_1^2 + Y_1^2\right] - \frac{2}{n}\sum_{i=1}^n \mathbb{E}\left[ G(U_{(i)})\right] \mathbb{E}\left[ H(U_{(i)}) \right].
\end{align*}
Now, it holds that 
\begin{align*}
   \wass_2^2(\mu, \nu) 
   &= \mathbb{E}[G(U)^2 + H(U)^2] - 2\mathbb{E}[G(U) H(U)] = \mathbb{E}[X_1^2 + Y_1^2] - 2\mathbb{E}\left[ \frac{1}{n}\sum_{i=1}^n G(U_i) H(U_i) \right] \\
   &= \mathbb{E}[X_1^2 + Y_1^2] -2\mathbb{E}\left[ \frac{1}{n}\sum_{i=1}^n  G(U_{(i)}) H(U_{(i)}) \right].
\end{align*}
The claimed result follows by subtracting off the previous identities.
\end{proof}

\begin{proposition} \label{prop:rate-1d}
    Let $(\mu,\nu)$ be one-dimensional measures with inverse-CDFs $(G, H)$ that are twice differentiable with uniformly bounded second derivatives. Then,
    \begin{equation*}
        \mathbb{E}\left[\wass_2^2(\mu_n, \nu_n) - \wass_2^2(\mu, \nu)\right] = 2 J(\mu, \nu)n^{-1} + o(n^{-1}),
    \end{equation*}
    where $J(\mu, \nu) = \smallint_0^1 u(1-u) G'(u)H'(u)\mathrm{d}u.$
\end{proposition}

\begin{proof} Let $U_{(1):(n)}$ be the order statistics of $U_{1:n} \iid \unif(0,1)$. By Lemma~\ref{lemma:w2sq-bias-1d}, we have that
\begin{equation*}
	n\mathbb{E}[\wass_2^2(\bar\mu_n, \nu_n) - \wass_2^2(\mu , \nu)] = 2 \sum_{i=1}^n \cov\left( G(U_{(i)}), H(U_{(i)}) \right).
\end{equation*}
We will estimate $\cov\left( G(U_{(i)}), H(U_{(i)}) \right)$ using Taylor expansions. We require the first two moments of $U_{(i)} \sim \textup{Beta}(i, n+1-i)$,
\begin{equation} \label{eqn:beta-moments}
    a_{(i)} := \mathbb{E}[U_{(i)}] = \frac{i}{n+1} 
    \enskip \text{and} \enskip
    \sigma^2_{(i)} := \var(U_{(i)}) = \frac{i(n+1-i)}{(n+1)^2(n+2)} = \frac{\frac{i}{n+1} \big(1 - \frac{i}{n+1}\big)}{(n+1)} + O(n^{-2}).
\end{equation}
Recall that $\sup_u |G''(u)| \le G_{\max}''$ and $\sup_u |H''(u)| \le H_{\max}''$ by assumption.

By the usual Taylor expansion,
\begin{equation*}
    G(U_{(i)}) = G(a_{(i)}) + (U_{(i)} - a_{(i)})G'(a_{(i)}) + r_G(U_{(i)}), \enskip \text{where} \enskip |r_G(U_{(i)})| \le G_{\max}'' (U_{(i)} - a_{(i)})^2.
\end{equation*}
Taking expectations on both sides, $\big|\mathbb{E}[G(U_{(i)})] - G(a_{i}) \big| \le G_{\max}'' \var(U_{(i)}) = G_{\max}'' \sigma^2_{(i)}.$ So, the triangle inequality gives
\begin{equation*}
    \left|G(U_{(i)}) - \mathbb{E}[G(U_{(i)})] - (U_{(i)} - a_{i})G'(a_{(i)}) \right| \le  G_{\max}'' \sigma^2_{(i)} + G_{\max}'' (U_{(i)} - a_{(i)})^2\sigma^2_{(i)},
\end{equation*}
with a similar result for $H$. Combining these results with the elementary inequality $|g_1h_1 - g_2h_2| \le |g_1-g_2| |h_1-h_2| + |g_2||h_1-h_2| + |h_2||g_1-g_2|$, we obtain that
\begin{multline*}
    \left| \left\{ G(U_{(i)}) - \mathbb{E}[G(U_{(i)})] \right\} \left\{ H(U_{(i)}) - \mathbb{E}[H(U_{(i)})] \right\} - (U_{(i)} - a_{i})^2G'(a_{(i)})H'(a_{(i)}) \right|\le \\
    \le  G_{\max}'' H_{\max}'' \big(\sigma^2_{(i)} + (U_{(i)} - a_{(i)})^2\big)^2 + \big(G'(a_{(i)}) G_{\max}'' + H'(a_{(i)}) H_{\max}''\big)|U_{(i)} - a_{i}| \big(\sigma^2_{(i)} + (U_{(i)} - a_{(i)})^2\big).
\end{multline*}
The expectation of the right-hand side is $O(n^{-3/2})$. Therefore,
\begin{equation*}
    \cov\left( G(U_{(i)}), H(U_{(i)})\right) = G'(a_{(i)})H'(a_{(i)}) \var(U_{(i)}) + O(n^{-3/2}).
\end{equation*}
Given the definition of $a_{(i)}$ and approximation of $\var(U_{(i)})$ in equation~\eqref{eqn:beta-moments}, the result follows from the definition of the Riemann integral and the size of the remainder when it is approximated by a Riemann sum.
\end{proof}

Proposition~\ref{prop:rate-1d} requires lighter-than-Gaussian tails \citep[][Section~5.1]{bobkov2019one-dimensional} and generalizes \citet[Proposition~5.5]{solomon2022k-variance} and \citet[Theorem~5.1]{bobkov2019one-dimensional}.

\subsection{Statistical properties}

\subsubsection{Proof of Theorem~\ref{thm:concentration-bounds}}

\paragraph{Estimator $U$.}

The estimator $U(\bar\mu_n, \mu_n,\nu_n) = \wass_2^2(\bar\mu_n, \nu_n) - \wass_2^2(\bar\mu_n, \mu_n)$ satisfies the bounded difference property under the compact space Assumption~\ref{assumption:compact}. Following \cite{weed2019sharp, chizat2020faster}, since the space has diameter~$1$ by Assumption~\ref{assumption:compact}, changing any of the samples within $\nu_n$ or $\mu_n$ can only change $U$ by at most $\pm n^{-1}$, and changing any one of the samples within $\bar\mu_n$ can only change $U$ by at most $\pm 2n^{-1}$. By the bounded difference inequalities \citep{mcdiarmid1989method}, it follows that
\begin{equation} \label{eqn:concentration-u-pf}
\begin{aligned}
    &{}\mathbb{P}\left( U(\bar\mu_n, \mu_n,\nu_n) - \mathbb{E} \left[U(\bar\mu_n, \mu_n,\nu_n)\right] \ge t \right) \le \exp \left(-2t^2 / \left\{ 2n (n^{-1})^2  + n (2n^{-1})^2 \right\}  \right) =  \exp(-nt^2/3),\\
    &{}\mathbb{P}\left( U(\bar\mu_n, \mu_n,\nu_n) - \mathbb{E} \left[U(\bar\mu_n, \mu_n,\nu_n)\right] \le -t \right) \le \exp(-nt^2/3),
\end{aligned}
\end{equation}
for any $t \ge 0$. A union bound concludes the proof.

\paragraph{Estimator $\bar L$.} The proof for the estimator $\bar L(\bar\mu_n, \mu_n,\nu_n) = \wass_2(\bar\mu_n, \nu_n) - \wass_2(\bar\mu_n, \mu_n)$ is more involved. Following \citet[Appendix A]{boissard2014on} we use the transportation method, which provides concentration bounds for Lipschitz functionals. Technical details are postponed to Lemma~\ref{lemma:transport-concentration}.

The key step is to establish that, when viewed as a function of its constituent samples, the estimator $\bar L: \mathbb{R}^{3nd} \to \mathbb{R}$ is Lipschitz. We show that $\lip{\bar L} \le 2 n^{-1/2}$ in Lemma~\ref{lemma:lbar-lipschitz}. The compact support Assumption~\ref{assumption:compact} puts us in the setting of Corollary~\ref{cor:transport-concentration}, hence
\begin{equation} \label{eqn:concentration-l-pf}
\begin{aligned}
    &{}\mathbb{P}\left( \bar L(\bar\mu_n, \mu_n,\nu_n) - \mathbb{E} \left[L(\bar\mu_n, \mu_n,\nu_n)\right] \ge t \right) \le \exp( - nt^4 / 32),\\
    &{}\mathbb{P}\left( \bar L(\bar\mu_n, \mu_n,\nu_n) - \mathbb{E} \left[L(\bar\mu_n, \mu_n,\nu_n)\right] \le -t \right) \le \exp( - nt^4 / 32),
\end{aligned}
\end{equation}
for any $t \ge 0$. A union bound concludes the proof.

\subsubsection{Proof of Theorem~\ref{thm:convergence-rates}}

We first require Lemma~\ref{lemma:rates}, which recalls the exact convergence rates of $\wass_2(\bar\mu_n, \mu_n)$ and $\wass_2^2(\bar\mu_n, \mu_n)$.

\begin{lemma} \label{lemma:rates}
    Let $d \ge 5$ and consider Assumption~\ref{assumption:compact}. Then,
    \begin{equation*}
        \mathbb{E}[\wass_2(\bar\mu_n, \mu_n)]^2 \asymp \mathbb{E}[\wass_2^2(\bar\mu_n, \mu_n)] \asymp n^{-2/d}.
    \end{equation*}
\end{lemma}
\begin{proof} By Jensen's inequality, we have that
\begin{equation*}
    \mathbb{E}[\wass_1(\bar\mu_n, \mu_n)]^2 \le \mathbb{E}[\wass_2(\bar\mu_n, \mu_n)]^2 \le \mathbb{E}[\wass_2^2(\bar\mu_n, \mu_n)].
\end{equation*}
Now, \citet[Theorem~2]{chizat2020faster} provides $\mathbb{E}[\wass_2^2(\bar\mu_n, \mu_n)] \lesssim n^{-2/d}$. To see the lower asymptote, by Lemma~\ref{lemma:plugin-bias} and \citet[Section~3.3]{panaretos2019statistical} it holds that $\mathbb{E}[\wass_1(\bar\mu_n, \mu_n)] \ge \mathbb{E}[\wass_1(\mu, \mu_n)] \gtrsim n^{-1/d}.$ The claimed result follows.
\end{proof}

We turn to the proof of the main result.

\paragraph{Estimator $U$.} 

By the triangle inequality,
\begin{equation*}
    \mathbb{E} \left[ |U - \wass_2^2(\mu,\nu)| \right] \le \mathbb{E} \left[|\wass_2^2(\bar\mu_n,\nu_n) - \wass_2^2(\mu,\nu)|\right] + \mathbb{E}\left[ \wass_2^2(\bar\mu_n,\mu_n)\right] \lesssim n^{-2/d},
\end{equation*}
where we finally used \citet[Theorem~2]{chizat2020faster} and Lemma~\ref{lemma:rates}.

\paragraph{Estimator $\bar L$.} 

By the triangle inequality,
\begin{equation*}
    \left | \mathbb{E} \left[\wass_2(\mu,\nu) - \bar L \right] - \mathbb{E}[\wass_2(\bar\mu_n,\mu_n)] \right| \le \mathbb{E} \left[ \left|\wass_2(\bar\mu_n,\nu_n) - \wass_2(\mu,\nu)\right|\right] \lesssim n^{-2/d},
\end{equation*}
where we finally used \citet[Corollary~1]{chizat2020faster}. Since $\mathbb{E}[\wass_2(\bar\mu_n,\mu_n)] \asymp  n^{-1/d}$ by Lemma~\ref{lemma:rates} and since $\mathbb{E} [\wass_2(\mu,\nu) - \bar L ] \ge 0$ by Proposition~\ref{prop:l}, it follows that $\mathbb{E} [\wass_2(\mu,\nu) - \bar L] \asymp \mathbb{E}[\wass_2(\bar\mu_n,\mu_n)] \asymp n^{-1/d},$ as claimed.

\subsubsection{Proof of Corollary~\ref{cor:bounds-high-prob}}

\paragraph{Estimator $U$.} Using the lower deviation bound~\eqref{eqn:concentration-u-pf} from the proof of Theorem~\ref{thm:concentration-bounds},
\begin{equation*}
    \mathbb{P}(U \ge \mathcal{W}_2^2(\mu, \nu)) \ge 1 - \exp \left( -\frac{n}{3} \left( \mathbb{E}[U] - \wass_2^2(\mu, \nu)\right)^2 \right) \ge 1 - \exp \left( C_1 n^{1-4/d}\right)
\end{equation*}
for some constant $C_1 > 0$, since we have assumed that $\mathbb{E}[U] - \wass_2^2(\mu, \nu) \gtrsim n^{-2/d}$.

\paragraph{Estimator $L$.}

Using the upper deviation bound~\eqref{eqn:concentration-l-pf} from the proof of Theorem~\ref{thm:concentration-bounds},
\begin{equation*}
    \mathbb{P}(L \le \mathcal{W}_2^2(\mu, \nu)) = \mathbb{P}(\bar L \le \mathcal{W}_2(\mu, \nu)) \ge 1 - \exp \left( -\frac{n}{32} \left( \wass_2(\mu, \nu) - \mathbb{E}[\bar L]\right)^4 \right) \ge 1 - \exp \left( C_2 n^{1-4/d}\right)
\end{equation*}
for some constant $C_2 > 0$, since $\wass_2(\mu, \nu) - \mathbb{E}[\bar L]  \gtrsim n^{-2/d}$ by Theorem~\ref{thm:convergence-rates}.

\subsubsection{Postponed auxiliary results} \label{app:transport-concentration}

Lemma~\ref{lemma:transport-concentration} details the key ingredients of the transportation method of obtaining concentration inequalities. The idea is the following: if the fluctuations of $\mu$, measured in some function of the Wasserstein distance, can be controlled by the Kullback-Leibler divergence $\kl (Q \mid \mu) = \int (\mathrm{d}Q /\mathrm{d}\mu) \log (\mathrm{d}Q /\mathrm{d}\mu) \mathrm{d} \mu$, then Lipschitz functions of $X \sim \mu$ concentrate. We refer to \citet[Chapter~8]{boucheron2013concentration} for a pedagogical treatment.
    
\begin{lemma} \label{lemma:transport-concentration}
Let $\alpha, \beta: \mathbb{R} \to [0, \infty)$ be increasing with $\alpha(0) = \beta(0) = 0.$ Let $\Omega \subseteq \mathbb{R}^d$ and let $\mathcal{P}(\Omega)$ be the set of all $\Omega$-valued distributions. Let the ground metric be Euclidean throughout. For $\mu \in \mathcal{P}(\Omega)$ we define the following conditions 
\begin{gather*}
    \mathbf{T}_1(\alpha) : \enskip \forall Q \in \mathcal{P}(\Omega) \text{ it holds that } \alpha(\wass_1(Q, \mu)) \le \kl(Q \mid \mu), \\ 
    \mathbf{T}_2^2(\beta) : \enskip \forall Q \in \mathcal{P}(\Omega) \text{ it holds that } \beta(\wass_2^2(Q, \mu)) \le \kl(Q \mid \mu).
\end{gather*}
The following claims hold:
\begin{enumerate}[label = (\roman*)]
    \item Suppose that $\mu \in \mathcal{P}(\Omega)$ satisfies condition $\mathbf{T}_1(\alpha)$. Then, for all $f : \Omega \to \mathbb{R}$ with $\lip{f} \le L$ it holds that
    \begin{equation*}
         \forall t \ge 0: \enskip \mathbb{P}_{X \sim \mu} \left( f(X) - \mathbb{E}\left[ f(X) \right] \ge t \right) \le \exp\left( -\alpha\left(t/L \right) \right).
    \end{equation*}
    \item Suppose that $\Omega$ has diameter at most $D$ and let $\mu \in \mathcal{P}(\Omega)$. Then, $\mu$ satisfies condition $\mathbf{T}_2^2(\beta)$ with $\beta(t) = t^2 / (2 D^4)$.
    \item Suppose that $\mu_1, \dots, \mu_m \in \mathcal{P}(\Omega)$ all satisfy condition $\mathbf{T}_2^2(\beta)$. Then, $\mu = \otimes_{i=1}^m \mu_i \in \mathcal{P}(\Omega^m)$ satisfies condition $\mathbf{T}_2^2(\beta)$.
    \item Suppose that $\mu \in \mathcal{P}(\Omega)$ satisfies condition $\mathbf{T}_2^2(\beta)$. Then, $\mu$ satisfies condition $\mathbf{T}_1(\alpha)$ with $\alpha(t) = \beta(t^2)$.
\end{enumerate}
\end{lemma}
\begin{proof}
    Claim (i) is equivalent to \citet[Lemma~5]{gozlan2007large}. Claim (ii) is a particular case of \citet[Particular case~2.5]{bolley2005weighted}. Claim (iii) is a particular case of \citet[Theorem~5]{gozlan2007large}: as the squared Euclidean metric tensorizes, so must $\mathbf{T}_2^2(\beta)$. Claim (iv) uses the following argument: as $\beta \ge 0$ is an increasing function, by Jensen's inequality it holds that $\beta(\wass_2^2(Q, \mu)) \ge \beta(\wass_1^2(Q, \mu))$. Therefore, if $\mu$ satisfies $\mathbf{T}_2^2(\beta)$, we have that
    \begin{equation*}
    \enskip \forall Q \in \mathcal{P}(\Omega) : \enskip \beta(\{\wass_1(Q, \mu)\}^2) \le \kl(Q \mid \mu),
    \end{equation*}
    which is precisely condition $\mathbf{T}_1(\alpha)$ with $\alpha(t) = \beta(t^2)$.
\end{proof}

Corollary~\ref{cor:transport-concentration} uses Lemma~\ref{lemma:transport-concentration} to derive a concentration bound for Lipschitz functions of compactly-supported product measures, and is used in the proof of Theorem~\ref{thm:concentration-bounds}.

\begin{corollary} \label{cor:transport-concentration}
Let $\mu_1, \dots, \mu_m \in \mathcal{P}(\Omega),$ where $\Omega\in \mathbb{R}^d$ has diameter at most $1$. Then, $\mu = \otimes_{i=1}^m \mu_i$ satisfies inequality $\mathbf{T}_1(\alpha)$ with $\alpha = t^4/2$. Therefore, for all $t \ge 0$,
\begin{gather*}
    \mathbb{P}_{X \sim \mu} \left( f(X) - \mathbb{E}\left[ f(X) \right] \ge t \right) \le \exp\left( -t^4 / (2 \lip{f}^4)\right),\\
    \mathbb{P}_{X \sim \mu} \left( f(X) - \mathbb{E}\left[ f(X) \right] \le -t \right) \le \exp\left( -t^4 / (2 \lip{f}^4)\right).
\end{gather*}
\end{corollary}
\begin{proof}
    Lemma~\ref{lemma:transport-concentration}(ii)-(iii) implies that $\mu = \otimes_{i=1}^m \mu_i$ satisfies $\mathbf{T}_2^2(\beta)$ with $\beta(t) = t^2/2$. Lemma~\ref{lemma:transport-concentration}(iv) implies that $\mu$ also satisfies $\mathbf{T}_1(\alpha)$ with $\alpha(t) = \beta(t^2) = t^4/2$. Lemma~\ref{lemma:transport-concentration}(i) concludes, noting that both $f$ and $-f$ have Lipschitz constant $\lip{f}$.
\end{proof}

Lemma~\ref{lemma:lbar-lipschitz} establishes that $\bar{L}$ is Lipschitz, and is used in the proof of Theorem~\ref{thm:concentration-bounds}.

\begin{lemma} \label{lemma:lbar-lipschitz}
    Viewing $\bar{L}(\bar\mu_n, \mu_n, \nu_n)$ as a function of its constituent samples, it holds that $\lip{\bar L} \le 2n^{-1/2}$.
\end{lemma}
\begin{proof}
    Let $Z = [\bar X_{1:n}, X_{1:n}, Y_{1:n}] \in \mathbb{R}^{3nd}$ denote a concatenation. We define $Z' = [\bar X_{1:n}', X_{1:n}', Y_{1:n}']$ and $\bar \mu_n' = \frac{1}{n}\sum_{i=1}^n \delta_{\bar X_i'}$, $\mu_n' = \frac{1}{n} \sum_{i=1}^n \delta_{X_i'}$, $\nu_n' = \frac{1}{n} \sum_{i=1}^n \delta_{Y_i'}$. We consider a minor abuse of notation and we equivalently define $\bar L (Z) = \bar L(\bar\mu_n, \mu_n, \nu_n) = \wass_2(\bar\mu_n, \nu_n) -  \wass_2(\bar\mu_n, \nu_n).$ The function $\bar L$ is Lipschitz because
    \begin{align*}
        \bigl| \bar L \left(Z\right) - \bar L\left(Z'\right) \bigr|
            &= \left| \wass_2(\bar\mu_n, \nu_n) - \wass_2(\bar\mu_n', \nu_n') - \wass_2(\bar\mu_n, \mu_n) + \wass_2(\bar\mu_n', \mu_n')\right| \\
            &\le \wass_2(\bar\mu_n, \bar\mu_n') + \wass_2(\nu_n, \nu_n') + \wass_2(\bar\mu_n, \bar\mu_n') + \wass_2(\mu_n, \mu_n') \\
            &\le n^{-1/2}\left( \| \bar X_{1:n} - \bar X_{1:n}'\| + \| Y_{1:n} - Y_{1:n}'\| + \| \bar X_{1:n} - \bar X_{1:n}'\| + \| X_{1:n} - X_{1:n}'\|  \right)\\
            &\le 2n^{-1/2} \left\| Z - Z' \right\|,
    \end{align*}
    where we firstly used several applications of the triangle inequality, secondly the definition of the primal formulation~\eqref{eqn:k-primal}, and finally the sharp inequality $x^{1/2} + y^{1/2} \le \{2(x+y)\}^{1/2}$ twice.
\end{proof}

\section{Uncertainty quantification} \label{app:uq}

\subsection{Jackknife variance estimation} \label{app:flapjack}

The jackknife estimator of variance \citep{efron1981jackknife} for the plug-in estimator $\wass_2^2(\mu_n, \nu_n),$ based on leave-one-out empirical measures of the form $\mu_{-i} = \frac{1}{n-1}\sum_{j\in [n]\setminus i} \delta_{X_j}$, reads
\begin{equation*}
    \var(\wass_2^2(\mu_n, \nu_n)) \approx \frac{n-1}{n} \sum_{i=1}^n \Big(\wass_2^2(\mu_{-i}, \nu_{-i}) - \frac{1}{n} \sum_{j=1}^n \wass_2^2(\mu_{-j}, \nu_{-j}) \Big)^2.
\end{equation*}
Analogous jackknife estimators can be derived for e.g. $U(\bar \mu_n, \mu_n, \nu_n)$ using leave-one-out versions $U(\bar \mu_{-i}, \mu_{-i}, \nu_{-i}).$

Naively computing all $i \in [n]$ leave-one-out estimators would have complexity $O(n^4)$. Below, we present the Flapjack algorithm, which takes advantage of warm starts \citep{mills-tettey2007dynamic} to reduce the complexity to $O(n^3)$. Understanding how this saving is obtained requires some background on linear assignment problem solvers, which we next recall.

\subsubsection{Solving assignment problems}

The primal and dual formulations of the linear assignment problem are
\begin{equation}\label{eqn:assignment}
    \begin{gathered}
        \min_{\sigma \in \mathbb{S}_n} \sum_{i=1}^n C_{i \sigma(i)} = \max_{u,v\in\mathbb{R}^n} \sum_{i=1}^n (u_i + v_i) \enskip \text{subject to} \enskip \forall (i,j): \, u_i + v_j \le C_{ij},
    \end{gathered}
\end{equation}
where $\mathbb{S}_n$ is the set of permutations of $[n],$ and $C \in \mathbb{R}^{n \times n}$ is a cost matrix.

Primal-dual assignment problem solvers \citep[e.g.][]{kuhn1955hungarian, munkres1957algorithms, jonker1987shortest} have the following general structure. We initialize with a set of feasible duals $(u,v)$ and an empty partial assignment $\sigma,$ where we write $\sigma(i) = *$ if a row $i$ has not been assigned to any column $j$. Each iteration, we apply a procedure $\texttt{stage}(C,u,v,\sigma)$ that returns a new triple $(u, v, \sigma)$ and: (i) increases the number of columns in the assignment by one; (ii) maintains feasibility across all duals, i.e. $\forall (i,j):\, u_i + v_j \le C_{ij};$ (iii) ensures that there is no dual slack across the matched pairs, i.e. $\forall i: u_i + v_{\sigma(i)} = C_{i\sigma(i)}$ if $\sigma(i) \ne *.$ The complementary slackness conditions ensure that we terminate correctly after $n$ iterations of \texttt{stage}.

Efficient implementations \citep[e.g.][]{jonker1987shortest} of $\texttt{stage}$ have worst-case complexities $O(n^2),$ so the worst-case complexity of assignment problem solvers is $O(n^3).$

\subsubsection{Solving leave-one-out assignment problems}

Suppose that we wish to solve the ``leave-one-out'' assignment problem, where row~$i$ and column~$i$ of the cost matrix $C$ are removed. 
A naive solution would require solving this modified assignment problem from scratch, and thus $O(n^3)$ operations. However, by starting from the solution $(u,v,\sigma)$ to the full-data assignment problem~\eqref{eqn:assignment}, it turns out that we can reduce this complexity by an order of magnitude.

\begin{algorithm}[ht] 
    \caption{Leave-one-out assignment cost, row $i$ and column $i$ of cost matrix removed} \label{alg:loo-cost}
    \DontPrintSemicolon
    \SetAlgoLined
    \textbf{Input:} Cost matrix $C$, optimal solution $(u,v,\sigma)$ to primal-dual pair~\eqref{eqn:assignment}.
    \begin{enumerate}[nosep,leftmargin=*]
        \item Remove row $i$ from assignment: $\sigma(i) = *$.
        \item Set small cost $C_{ii} = \varepsilon$ to guarantee assignment of pair $(i,i)$, e.g. $\varepsilon < \min_{ij} C_{ij} - 2 \max_{ij} C_{ij}.$
        \item Restore feasibility: if $u_i + v_i > C_{ii}$ set $u_i = C_{ii} - v_i.$
        \item Solve for assignment: $(u, v, \sigma) \leftarrow \texttt{stage}(C, u, v, \sigma).$ 
        \item Return $\sum_{j=1, j\ne i}^n C_{j\sigma(j)}$ and reset $C_{ii}$.
    \end{enumerate}
\end{algorithm}

Algorithm~\ref{alg:loo-cost} uses the method of \cite{mills-tettey2007dynamic} to solve for the leave-one-out assignment cost. It solves an equivalent problem: $C_{ii} = \varepsilon$ is set small enough so that row $i$ is guaranteed to be assigned to column $i$; $C_{ii}$ is then discarded in the final calculation. By removing row $i$ from the assignment (line~1) and then restoring feasibility (line~3), we still obey complementary slackness with $(n-1)$ assigned rows, so one iteration of \texttt{stage} (line~4) suffices to obtain the correct solution.

Efficient implementations of Algorithm~\ref{alg:loo-cost} have $O(n^2)$ complexities, a significant saving compared to the $O(n^3)$ cost of solving the leave-one-out problem without a warm start.

\subsubsection{Flapjack algorithm}

The procedure we call ``Flapjack'' starts from an optimal solution to the assignment problem~\eqref{eqn:assignment}, then applies Algorithm~\ref{alg:loo-cost} for $i \in [n]$ to return all leave-one-out assignment costs.

Our implementation of Flapjack uses $\texttt{stage}$ from \cite{jonker1987shortest}, so has a worst-case complexity of $O(n^3).$ We also observe this scaling in practice (see Figure~\ref{fig:bench}), which relates to a tendency of the algorithm of \cite{jonker1987shortest} to perform many scans when when most of the partial assignment $\sigma$ has been filled \citep[see also][Section~3.1]{guthe2021toms1015}.

Flapjack can be used to compute the jackknife estimate of variance for the plug-in estimator $\wass_2^2(\mu_n, \nu_n)$ by fixing $C_{ij} = \|X_i - Y_j\|^2$, in which case the full-data assignment cost is $n\wass_2^2(\mu_n, \nu_n)$ and the $i$-th leave-one-out assignment cost is $(n-1)\wass_2^2(\mu_{-i}, \nu_{-i}).$

\subsection{Approximate delta method for $\bar L$} \label{app:uq-lbar}

We detail our approximate delta method for $\bar L(\bar\mu_n, \mu_n, \nu_n)  = \wass_2(\bar\mu_n, \nu_n) - \wass_2(\bar\mu_n, \nu_n).$

Let $\Delta(\alpha, \beta):= \wass_2^2(\alpha, \beta) - \mathbb{E}[(\alpha, \beta)]$. Taylor's theorem and a further approximation provide
\begin{equation*}
    \wass_2(\bar\mu_n, \nu_n) \approx \mathbb{E}[\wass_2^2(\bar\mu_n, \nu_n)]^{1/2} + \frac{\Delta(\bar\mu_n, \nu_n)}{2\mathbb{E}[\wass_2^2(\bar\mu_n, \nu_n)]^{1/2}} \approx \mathbb{E}[\wass_2(\bar\mu_n, \nu_n)] + \frac{\Delta(\bar\mu_n, \nu_n)}{2\mathbb{E}[\wass_2(\bar\mu_n, \nu_n)]},
\end{equation*}
the former of which is accurate when $\var(\wass_2^2(\bar\mu_n, \nu_n)) \ll \mathbb{E}[\wass_2^2(\bar\mu_n, \nu_n)],$ whereas the latter when $\var(\wass_2(\bar\mu_n, \nu_n)) \ll \mathbb{E}[\wass_2(\bar\mu_n, \nu_n)]^2.$ Both conditions hold as $n \to \infty$. This suggests the approximation
\begin{equation} \label{eqn:delta-lbar}
\bar L - \mathbb{E}[\bar L] \approx \frac{\Delta(\bar\mu_n, \nu_n)}{2\mathbb{E}[\wass_2(\bar\mu_n, \nu_n)]} - \frac{\Delta(\bar\mu_n, \mu_n)}{2\mathbb{E}[\wass_2(\bar\mu_n, \mu_n)]}.
\end{equation}

Using that $\Delta(\bar\mu_n, \nu_n) = \frac{1}{n} \sum_{i \in [n]} [\phi_{\bar\mu_n, \nu_n}(\bar X_{i}) + \psi_{\bar\mu_n, \nu_n}(Y_{i})] + \text{const},$ we derive the variance estimate 
\begin{equation*}
    \var(\bar L) \approx \frac{1}{n}\var\left(\left\{ \frac{\phi_{\bar\mu_n, \nu_n}(\bar X_{i}) + \psi_{\bar\mu_n, \nu_n}(Y_{i})}{2 \wass_2(\bar\mu_n, \nu_n)} - \frac{\phi_{\bar\mu_n, \mu_n}(\bar X_{i}) + \psi_{\bar\mu_n, \mu_n}(X_{i})}{2 \wass_2(\bar\mu_n, \mu_n)} \right\}_{i=1}^n \right),
\end{equation*}
based on~\eqref{eqn:delta-lbar} and the insight that the empirical Kantorovich potentials are asymptotically i.i.d. \citep[implicit in the results of][]{delbarrio2024central}. Although this is only a heuristic, experiments in a setting similar to Figure~\ref{fig:debiased-var} reveal that the variance estimate is more accurate than the jackknife, while being slightly conservative.

\subsection{Estimators that use independent blocks of correlated samples} \label{app:uq-bias}

We describe how to quantify uncertainty in the setting of Section~\ref{sec:bias-sampling}.

For simplicity, let $B_\mu=B_\nu=0$ and $T_\mu=T_\nu=1.$ Define the sum of the Kantorovich potentials $f_{\mu, \nu}(x,y) := \phi_{\mu, \nu}(x) + \psi_{\mu, \nu}(y)$. The proposed estimators are
\begin{align*}
    U(\bar\mu_n, \mu_n,\nu_n) &= \frac{1}{K} \sum_{k=1}^K \frac{1}{I} \sum_{i=0}^{I-1} \Big[ f_{\bar\mu_n, \nu_n}(X^{(i)}_{k+K}, Y^{(i)}_k) - f_{\bar\mu_n, \mu_n}(X^{(i)}_{k+K}, X^{(i)}_k) \Big],\\
    \bar L(\bar\mu_n, \mu_n,\nu_n) &= \frac{1}{K} \sum_{k=1}^K \frac{1}{I} \sum_{i=0}^{I-1} \bigg[ \frac{f_{\bar\mu_n, \nu_n}(X^{(i)}_{k+K}, Y^{(i)}_k)}{\wass_2(\bar\mu_n, \nu_n)} - \frac{f_{\bar\mu_n, \mu_n}(X^{(i)}_{k+K}, X^{(i)}_k)}{\wass_2(\bar\mu_n, \mu_n)} \bigg].
\end{align*}
To quantify the uncertainty in $\{U, \bar L\}$, we use Gaussian confidence intervals, based on the empirical variances
\begin{align*}
    \var(U) &\approx \frac{1}{K}\var\left( \left\{\frac{1}{I}\sum_{i=0}^{I-1} \Big[ f_{\bar\mu_n, \nu_n}(X^{(i)}_{k+K}, Y^{(i)}_k) - f_{\bar\mu_n, \mu_n}(X^{(i)}_{k+K}, X^{(i)}_k) \Big] \right\}_{k = 1}^K \right),\\
    \var(\bar L) &\approx \frac{1}{K}\var\left( \left\{\frac{1}{I}\sum_{i=0}^{I-1} \bigg[ \frac{f_{\bar\mu_n, \nu_n}(X^{(i)}_{k+K}, Y^{(i)}_k)}{\wass_2(\bar\mu_n, \nu_n)} - \frac{f_{\bar\mu_n, \mu_n}(X^{(i)}_{k+K}, X^{(i)}_k)}{\wass_2(\bar\mu_n, \mu_n)} \bigg] \right\}_{k = 1}^K \right),
\end{align*}
with consistency as $K \to \infty.$ These can be justified using an extension of \citet[Theorem~4.10]{delbarrio2024central} and the approximate delta method of Appendix~\ref{app:uq-lbar}.

\paragraph{Quantifying the variance reduction due the coupling.} 

When instead $(\mu_n, \nu_n)$ are correlated, we can use the estimator
\begin{multline*}
    \var(U_\text{indep}) \approx 
    \frac{1}{K} \var \left( \bigg\{\frac{1}{I} \sum_{i=0}^{I-1} \left[ \phi_{\bar\mu_n, \nu_n}(\X{i}_{k+K}) - \phi_{\bar\mu_n, \mu_n}(\X{i}_{k+K}) \right] \bigg\}_{k=1}^K  \right)\\
    + \frac{1}{K} \var \left( \bigg\{\frac{1}{I} \sum_{i=0}^{I-1} \psi_{\bar\mu_n, \nu_n}(Y^{(i)}_k) \bigg\}_{k=1}^K \right)
    + \frac{1}{K} \var \left( \bigg\{\frac{1}{I} \sum_{i=0}^{I-1} \psi_{\bar\mu_n, \mu_n}(\X{i}_k) \bigg\}_{k=1}^K \right)
\end{multline*}
to \emph{estimate the variance of $U$ as if $(\mu_n, \nu_n)$ were independent}, without actually requiring us to draw independent versions of these empirical measures. A similar estimator can be considered for $\bar L.$

When $\var(U_\text{indep}) \ge \var(U)$, since $\var(U_\text{indep})$ and $\var(U)$ are noisy overestimates of the actual variances, we expect to obtain a noisy underestimate of the factor of variance reduction $\var(U_\text{indep}) / \var(U).$

\subsection{Time-averaged estimators} \label{app:uq-conv}

We describe how to quantify uncertainty in the setting of Appendix~\ref{app:convergence-time-averaging}.

Let $\marg{t}_{n} = \frac{1}{n}\sum_{i=1}^n \delta_{\X{t}_{i}},$ based on replicates $(\X{t}_{i})_{t \ge 0}$ of a stochastic process for $i \in [n]$. Define the sum of the Kantorovich potentials $f_{\mu, \nu}(x,y) := \phi_{\mu, \nu}(x) + \psi_{\mu, \nu}(y)$. The estimators of Appendix~\ref{app:convergence-time-averaging} are
\begin{align*}
    U_{T,t} &= \frac{1}{n} \sum_{i=1}^n \Big[ f_{\marg{T}_{n}, \marg{t}_{n}}(\X{T}_i,\X{t}_i) - \frac{1}{|\mathcal{S}|}\sum_{S \in \mathcal{S}} f_{\marg{T}_{n}, \marg{S}_{n}}(\X{T}_i,\X{S}_i) \Big],\\
    \bar L_{T,t} &= \frac{1}{n} \sum_{i=1}^n \bigg[ \frac{f_{\marg{T}_{n}, \marg{t}_{n}}(\X{T}_i,\X{t}_i)}{\wass_2(\marg{T}_{n}, \marg{t}_{n})} - \frac{1}{|\mathcal{S}|}\sum_{S \in \mathcal{S}} \frac{f_{\marg{T}_{n}, \marg{S}_{n}}(\X{T}_i,\X{S}_i)}{\wass_2(\marg{T}_{n}, \marg{S}_{n})} \bigg].
\end{align*}
To quantify the uncertainty in $\{U_{T,t}, \bar L_{T,t}\}$, we use Gaussian confidence intervals, based on the empirical variances
\begin{align*}
    \var(U_{T,t}) &\approx \frac{1}{n} \var \left( \Big\{ f_{\marg{T}_{n}, \marg{t}_{n}}(\X{T}_i,\X{t}_i) - \frac{1}{|\mathcal{S}|}\sum_{S \in \mathcal{S}} f_{\marg{T}_{n}, \marg{S}_{n}}(\X{T}_i,\X{S}_i) \Big\}_{i=1}^n \right),\\
    \var(\bar L_{T,t}) &\approx \frac{1}{n} \var \left( \bigg\{ \frac{f_{\marg{T}_{n}, \marg{t}_{n}}(\X{T}_i,\X{t}_i)}{2\wass_2(\marg{T}_{n}, \marg{t}_{n})} - \frac{1}{|\mathcal{S}|}\sum_{S \in \mathcal{S}} \frac{f_{\marg{T}_{n}, \marg{S}_{n}}(\X{T}_i,\X{S}_i)}{2\wass_2(\marg{T}_{n}, \marg{S}_{n})} \bigg\}_{i=1}^n \right),
\end{align*}
where consistency is as $n \to \infty$. These can be justified using extensions of \citet[Theorem~4.10]{delbarrio2024central} and the approximate delta method of Appendix~\ref{app:uq-lbar}. For $L_{T,t} = \sgnpow{2}{\bar L_{T,t}},$ we scale up the confidence interval for $\bar L_{T,t}$ accordingly.

\section{Description of MCMC Algorithms} \label{app:mcmc-algorithms}

We describe the MCMC algorithms that are used in the analysis of Appendix~\ref{app:mcmc-theory}.

\subsection{ULA} \label{app:ula}

The unadjusted Langevin algorithm (ULA) targeting $\pi$ generates a Markov chain $(\X{t})_{t \ge  0}$ based on the recursion
\begin{equation*}
    \X{t+1} = \X{t} + \frac{h^2}{2}A\nabla\log\pi(\X{t}) + \varepsilon^{\smash{(t)}}, \enskip \varepsilon^{\smash{(t)}} \sim \mathcal{N}_d(0_d, h^2A),
\end{equation*}
where the user sets the step size $h>0$ and the preconditioner $A \succ 0.$

\subsection{OBABO} \label{app:obabo}

The OBABO discretization of the underdamped Langevin diffusion, targeting $\pi$ in the $X$-component, generates a Markov chain $(\X{t}, \Z{t})_{t \ge  0}$ based on the recursion\footnote{For simplicity, we collapsed the two partial O-steps into a full step.}
\begin{alignat*}{5}
   &\textbf{O:} \enskip&& \Z{t}_\eta = \eta \Z{t} + \sqrt{1- \eta^2}\varepsilon^{\smash{(t)}}, \enskip\varepsilon^{\smash{(t)}} \sim \mathcal{N}_d(0_d, A),\\
   &\textbf{B:} \enskip&&\Z{t+1/2} = \Z{t}_\eta + \frac{h}{2} A \nabla\log\pi(\X{t}),\\
   &\textbf{A:} \enskip&&\X{t+1} = \X{t} + h \Z{t+1/2},\\
   &\textbf{B:} \enskip&& \Z{t+1} = \Z{t+1/2} + \frac{h}{2} A \nabla\log\pi(\X{t+1}),
\end{alignat*}
where the user sets the step size $h>0,$ the preconditioner $A \succ 0,$ and the momentum persistence parameter $\eta \in [0, 1).$ When $\eta = 0,$ the process $(\X{t})_{t \ge 0}$ is an ULA chain.

\subsection{Gibbs sampler for half-t regression} \label{app:mcmc-half-t}

We consider a linear regression model with half-t($\nu$) priors,
\begin{equation} \label{eqn:half-t-model}
    \begin{gathered}
    y \mid X, \beta, \sigma^2 \sim \mathcal{N}(X\beta, \sigma^2 I_n)\\
    \beta_j | \eta, \xi, \sigma^2 \sim \mathcal{N} \Big( 0, \frac{\sigma^2}{\xi \eta_j} \Big), \enskip \eta_j^{-1/2} \sim t_+(\nu), \enskip \textup{independently for } j \in [d], \\
    \xi^{-1/2} \sim \mathcal{C}_+(0,1), \enskip \sigma^{-2} \sim \textup{Gamma} \Big( \frac{a_0}{2}, \frac{b_0}{2} \Big),
\end{gathered}
\end{equation}
where $\mathcal{C}_+(0,1)$ is the half-Cauchy distribution with density $\pi_\xi(x) \propto 1/(1+x^2)$ and
$t_+(\nu)$ is the half-t distribution with $\nu$ degrees of freedom.

Algorithm~\ref{alg:half-t} describes the approximate Gibbs sampler\footnote{The selection of the active set $\mathbb{I}_\varepsilon$ mirrors the implementations of \href{https://github.com/jamesjohndrow/horseshoe_jo/blob/master/horseshoe.jl\#L114}{Johndrow et al.} and \href{https://github.com/niloyb/BoundWasserstein/blob/main/half_t/half_t_functions.R\#L103}{Biswas and Mackey}.} of \citet[Section~4.2]{biswas2024bounding} targeting the posterior distribution $\pi(\eta,\xi,\sigma^2,\beta\mid X, y)$ of the regression model~\eqref{eqn:half-t-model}, where 
\begin{equation*}
    \begin{gathered}
        M(\xi, \eta, X) = I_n + \xi^{-1} X \diag(\eta^{-1}) X^\top, \\ 
        \log L(y,M) = - \frac{1}{2} \log \det(M) - \frac{a_0 + n}{2} \log( b_0 + y^\top M^{-1} y).
    \end{gathered}
\end{equation*}
The exact algorithm of \cite{biswas2021coupling-based} corresponds to setting $\varepsilon = 0$ in Algorithm~\ref{alg:half-t}.

\begin{algorithm}[ht] 
    \caption{Approximate Gibbs sampler for regression model~\eqref{eqn:half-t-model}} \label{alg:half-t}
    \DontPrintSemicolon
    \SetAlgoLined
    \textbf{Input:} current state $(\eta,\xi,\sigma^2,\beta)$, approximation parameter $\varepsilon \ge 0,$ step size $\sigma_\text{MH}$.
\begin{enumerate}[leftmargin=*]
\item Sample $\eta \mid \xi, \sigma^2, \beta$ component-wise. For each component $j$, target
$$\pi(\eta_{j} | \dots) \propto \eta_{j}^{\frac{\nu-1}{2}} (1+\nu \eta_{j})^{-\frac{\nu + 1}{2}} \exp({-m_{j} \eta_{j}}),$$ 
with $m_{j} = \xi \beta_{j}^2/(2 \sigma^2),$ using the slice sampler of \citet[Algorithm~4]{biswas2021coupling-based}.
\item Sample $\xi, \sigma^2, \beta \mid \eta$ as follows:
    \begin{enumerate}[(a)]
    \item Sample $\xi \mid \eta$ with approximate Metropolis-Hastings.
    
    Propose $\log\xi^*\sim$ $\mathcal{N}_1(\log\xi, \sigma_\text{MH}^2)$ and fix $\mathbb{I}_\varepsilon = \diag\big(\mathbbm{1}{\{\min(\xi^*, \xi)^{-1} \eta^{-1} > \varepsilon\}}\big).$
    
    Calculate acceptance probability
    \begin{equation*}
        q = \frac{L(y,M(\xi^*, \eta, X\mathbb{I}_\varepsilon))}{L(y,M(\xi, \eta, X\mathbb{I}_\varepsilon))} \frac{\pi_{\xi}(\xi^*)}{\pi_{\xi}(\xi)} \frac{\xi^*}{\xi}.
    \end{equation*}
    
    With probability $q$ set $\xi = \xi^*.$
    
    \item Sample $$\sigma^2 \mid \eta, \xi \sim \invgamma\Big(\frac{a_0+n}{2}, \frac{y^\top M(\xi, \eta, X\mathbb{I}_\varepsilon)^{-1} y + b_0}{2}\Big).$$
    \item Sample $$ \beta \mid \eta, \xi, \sigma^2  \sim \mathcal{N} \big( \Sigma_\varepsilon^{-1} (X\mathbb{I}_\varepsilon)^\top y, \sigma^2 \Sigma_\varepsilon^{-1} \big)$$
    with $\Sigma_\varepsilon = (X \mathbb{I}_\varepsilon)^\top (X \mathbb{I}_\varepsilon) + \xi\diag(\eta),$ using the algorithm of \cite{bhattacharya2016fast}.
    \end{enumerate}
\item Return $(\eta,\xi,\sigma^2,\beta)$.
\end{enumerate}
\end{algorithm}

\section{Analysis for Sections~\ref{sec:bias} and~\ref{sec:convergence}} \label{app:mcmc-theory}

\subsection{Proof of Proposition~\ref{prop:ula-bias-overdisp}} \label{app:obabo-bias}

Proposition~\ref{prop:ula-bias-overdisp} is an immediate consequence of the following result.

\begin{proposition} \label{prop:obab-stationary}
    Let $\pi = \mathcal{N}_d(\mu, \Sigma)$ and let the spectral radius $\rho\big(\frac{h^2}{4} A^{1/2}\Sigma^{-1}A^{1/2}\big) < 1$. The following claims hold:
    \begin{enumerate}[label=(\roman*)]
        \item The invariant distribution of the OBABO chain of Appendix~\ref{app:obabo} is $\marg{\infty} \otimes \mathcal{N}_d(0_d, A),$ where $\marg{\infty} = \mathcal{N}_d\big(\mu, (I_d - \frac{h^2}{4}A^{1/2}\Sigma^{-1}A^{1/2})^{-1}\Sigma\big)$. 
        \item The invariant distribution of the ULA chain of Appendix~\ref{app:ula} is $\marg{\infty}$.
        \item $\marg{\infty}\cotr \pi.$
    \end{enumerate}
\end{proposition}
\begin{proof} We first consider the case $A = I_d$ and $\mu = 0.$

For claim (i), the steps BAB form a velocity Verlet integrator of Hamiltonian dynamics. By e.g. \citet[Section~2.3.1]{apers2024hamiltonian}, these dynamics are an exact time-discretization of Hamiltonian dynamics that leave the Hamiltonian $ H(x, z) = \frac{1}{2}x^\top \Sigma^{-1}\big(I - \frac{h^2}{4} \Sigma^{-1}\big) x + \frac{1}{2}\|z\|^2$ invariant. The O step leaves the marginal distribution $\mathcal{N}_d(0_d, I_d)$ invariant. It follows that the invariant distribution of the OBABO chain is $\mathcal{N}_d\big(\mu, (I_d - \frac{h^2}{4}\Sigma^{-1})^{-1}\Sigma\big) \otimes \mathcal{N}_d(0_d, I_d).$

For claim (ii), we use that ULA is a particular case of OBABO with $\eta=0.$

For claim (iii), we use that $(I_d - \frac{h^2}{4}\Sigma^{-1})^{-1}\Sigma \succeq \Sigma,$ then apply Proposition~\ref{prop:cot-cases}(i).

Finally, to deal with the case of general $(A, \mu),$ we use that the process $(\smash{\bar{X}^{(t)}, \bar{Z}^{(t)}})_{t \ge 0} = (\smash{A^{-1/2}X^{(t)} - \mu, A^{-1/2} Z^{(t)}})_{t \ge 0}$ is an OBABO chain with preconditioner $\bar A = I_d.$ Transforming back to the original process provides the claimed results.
\end{proof}

\subsection{Overdispersion of approximate Gibbs sampler for half-t regression} \label{app:half-t-overdisp}

Algorithm~\ref{alg:half-t} explicitly zeroes the columns of the design matrix $X$ with \emph{a posteriori} weakest signal (via $X\mathbb{I}_\varepsilon$ in step 2). Compared to the exact algorithm of \cite{biswas2021coupling-based} (Algorithm~\ref{alg:half-t} with $(\varepsilon, X \mathbb{I}_\varepsilon) = (0, X)$), this allows for faster computation in high-dimensional settings, and causes Algorithm~\ref{alg:half-t} to sample from an overdispersed version of the exact posterior distribution of the regression coefficients $\beta$, as we now explain.

Inspecting how step 2 in Algorithm~\ref{alg:half-t} changes as the level of approximation $\varepsilon \ge 0$ increases, we see that $\mathbb{I}_\varepsilon$ becomes sparser, so the sequences $(M(\xi, \eta, X\mathbb{I}_\varepsilon)^{-1})_{\varepsilon \ge 0}$ and $(\Sigma_\varepsilon^{-1})_{\varepsilon \ge 0}$ increase in the Loewner order. Therefore, the update $\sigma^2\mid \eta, \xi$ increases in the usual stochastic order and the update $\beta \mid \eta, \xi, \sigma^2$ becomes more dispersed and the active components $\mathbb{I}_\varepsilon \beta$ become more outwardly shifted.\footnote{For the inactive components $(I_d - \mathbb{I}_\varepsilon)\beta,$ since they correspond to a weak signal, the dispersion term in the update $\beta \mid \eta, \xi, \sigma^2$ dominates.} This indicates that stationary distribution of $\beta$ spreads out as $\varepsilon$ increases.

\subsection{Proof of Proposition~\ref{prop:cot-convergence-gauss}}

Proposition~\ref{prop:cot-convergence-gauss} is an immediate consequence of the following result and Proposition~\ref{prop:cot-cases}(i).

\begin{theorem} \label{thm:ar1-characterize}
Let $(\X{t})_{t \ge 0}$ be the AR(1) process with recursion
\begin{equation*}
    \X{t+1} - \mu = B (\X{t} - \mu) + A Z^{(t)}, \enskip Z^{(t)}\sim \mathcal{N}_d(0_d, I_d).
\end{equation*}
Let the spectral radius $\rho(B) < 1$ and let $\Mu{t} = \mathbb{E}[\X{t}]$ and $\Sig{t} = \var(\X{t})$. The following claims hold:
\begin{enumerate}[label=(\roman*)]
    \item The process converges to the stationary distribution $\marg{\infty} = \mathcal{N}(\Mu{\infty}, \Sig{\infty})$, where $\Mu{\infty} = \mu$ and $\Sig{\infty} = \sum_{n \ge 0} B^n AA^\top (B^n)^\top.$
    \item $\Mu{t} - \Mu{\infty} = B^t (\Mu{0} - \Mu{\infty})$ and $\Sig{t} - \Sig{\infty} = B^t (\Sig{0} - \Sig{\infty}) (B^t)^\top$ for all $t\ge 0$.
    \item If $\Sig{0} \succeq \Sig{\infty},$ then $\Sig{t} \succeq \Sig{\infty}$ for all $t \ge 0.$
    \item If $\X{0}$ is Gaussian, then $\X{t}$ is Gaussian for all $t\ge 0$.
\end{enumerate}
\end{theorem}

\begin{proof}
Taking means and variances in the autoregression, we obtain
\begin{equation*}
\Mu{t+1} - \mu = B(\Mu{t} - \mu), \quad \Sig{t + 1} = B \Sig{t} B^\top + A A^\top.
\end{equation*}

For claim (i), the convergence part is well-known \citep{tjostheim1990nonlinear}. The stationary distribution $\marg{\infty} = \mathcal{N}(\Mu{\infty}, \Sig{\infty})$ is a fixed point of the autoregression; the solutions $\Mu{\infty} = \mu$ and $\Sig{\infty} = \sum_{n \ge 0} B^n AA^\top (B^n)^\top$ can be seen by inspection.

For claim (ii), since $\Sig{\infty}$ is a fixed point of the autoregression, it holds that $\Sig{\infty} =  B \Sig{\infty} B^\top + A A^\top$. Subtracting this off from the autoregression, we obtain that $\Sig{t+1} - \Sig{\infty} = B (\Sig{t} - \Sig{\infty}) B^\top$. Similarly, $\Mu{t+1} - \Mu{\infty} = B (\Mu{t} - \Mu{\infty})$. The claim follows by induction.

Claim (iii) follows from claim (ii).

Claim (iv) follows from the closure of Gaussians under affine transformations.
\end{proof}

\subsection{Verifying the claims of Remark~\ref{rem:ar1}} \label{app:checking-samplers-ar1}

\paragraph{Underdamped Langevin.} We consider the underdamped Langevin diffusion (ULD)
\begin{equation*}
\mathrm{d} 
\begin{bmatrix}
    \X{t}\\ \Z{t}
\end{bmatrix}
= \frac{1}{2}
\begin{bmatrix}
    A^{-1} \Z{t} \\
    \nabla \log\pi(\X{t}) \mathrm{d}t - \gamma\Z{t}
\end{bmatrix}
+
\begin{bmatrix}
    0 \\
    (\gamma A)^{1/2} \mathrm{d} W_t
\end{bmatrix}
\end{equation*}
with stationary distribution $\pi \otimes \mathcal{N}_d(0_d,A),$ where $(W_t)_{t \ge 0}$ is Brownian motion, $\gamma \in (0, \infty)$ is a friction parameter, and $A\succ 0$ is a preconditioner.

We now verify that overdispersion persists in the $X$-coordinate when the target is $\pi = \mathcal{N}_d(\mu, \Sigma).$ Suppose that $\X{0}$ is drawn independently from $\Z{0}\sim \mathcal{N}_d(0_d,A).$ 
Since the stationary and initial distributions factorize over the $X$- and $Z$-components, and furthermore since any time-discretization of the ULD is an AR(1) process, Theorem~\ref{thm:ar1-characterize}(ii) provides 
\begin{equation*}
    \begin{bmatrix}
        \Sig{t} - \Sigma & * \\
        * & *
    \end{bmatrix}
    =
    B_t   
    \begin{bmatrix}
        \Sig{0} - \Sigma & 0 \\
        0 & 0
    \end{bmatrix}
    B_t^\top, \enskip \text{for some } B_t \text{ and for all } t \ge 0,
\end{equation*}
where the blocks represent the $X$- and $Z$-components, where $\Sigma^{(t)} := \var(\X{t})$, and where $*$ denotes an arbitrary entry. Therefore, $\Sig{0} \succeq \Sig{\infty}$ implies that $ \Sig{t} \succeq \Sig{\infty}$ for all $t \ge 0,$ as desired.

\paragraph{Random scan Gibbs.} For random scan Gibbs samplers targeting Gaussians, we can prove the following result related to overdispersion over time.

\begin{proposition} \label{prop:ar1-mixture}
    Let $(\X{t})_{t \ge 0}$ be a random scan Gibbs sampler targeting $\marg{\infty} = \mathcal{N}_d(\Mu{\infty}, \Sig{\infty}).$ The following claims hold:
    \begin{enumerate}[label = (\roman*)]
    \item If $\marg{0}$ is Gaussian, then $\marg{t}$ is a mixture of Gaussian distributions for all $t \ge0$, say $\marg{t} := \sum_{k=1}^{K^{(t)}} p_k \mathcal{N}(\Mu{t}_k, \Sig{t}_k).$
    \item Let $\marg{0} = \mathcal{N}(\Mu{0}, \Sig{0}).$ If $\Sig{0} \succeq \Sig{\infty}$, then $\Sig{t}_k \succeq \Sig{\infty}$ for all $(t,k).$ Therefore, $\marg{t}\pcar \marg{\infty}$ for all $t \ge0.$
    \end{enumerate}
\end{proposition}
\begin{proof}
Representing the random scan Gibbs kernel as a mixture of Gibbs steps, we can write the evolution of the chain as \begin{equation} \label{eqn:ar1-mixture}
    \X{t+1} = \sum_{m = 1}^M \mathbbm{1}_{\{ \M{t} = m \}} \left(B_m \X{t} + A_m \Z{t}\right), \enskip \Z{t} \sim \mathcal{N}_d(0_d, I_d)
\end{equation}
where $\M{t} \sim \textup{Categorical}(p_{1:k})$ selects the mixture component, and where each of the components is a $\marg{\infty}$-invariant Gibbs step.

For claim (i), $\marg{t}$ is a Gaussian mixture for all $t \ge 0$ because linear-Gaussian mixture kernels are closed under Gaussian mixtures.

For claim (ii), we argue by induction. The base case $t = 0$ is trivial. Fixing $t \ge 0$, the recursion~\eqref{eqn:ar1-mixture} implies that for all $k$, there exist $(\ell,m)$ such that $\Sig{t+1}_k =  B_m\Sig{t}_\ell B_m^\top + A_m A_m^\top.$ Since all kernels are $\marg{\infty}$-invariant, $\Sig{\infty}$ is a fixed point of the recursion~\eqref{eqn:ar1-mixture}, hence $\Sig{t+1}_k - \Sig{\infty} = B_m(\Sig{t}_\ell - \Sig{\infty}) B_m^\top.$ Therefore, $\Sig{t}_\ell \succeq \Sig{\infty}$ implies $\Sig{t+1}_k \succeq \Sig{\infty}$. By induction, $\Sig{t}_k \succeq \Sig{\infty}$ for all $(t,k)$. Finally, because $\pcar$ is partially closed under mixtures, it follows that $\marg{t}\pcar \marg{\infty}$ for all $t \ge0.$
\end{proof}

\section{Estimating the convergence of Markov chains} \label{app:convergence-methods}

\subsection{Plug-in method with time-averaging} \label{app:convergence-time-averaging}

We present a refinement of the MCMC convergence rate estimation method of Section~\ref{sec:convergence} that applies to overdispersed initializations only. The method proceeds as follows. 

We simulate $n$ replicate Markov chains with marginals $(\marg{t})_{t \ge 0}$ up to a large time $T \gg 1$. We collect the samples from $\marg{t}$ with equal weight in $\marg{t}_n,$ for $t \ge 0$. We then estimate $$L_{T,t}\lessapprox \wass_2^2(\marg{T},\marg{t}) \lessapprox U_{T,t}$$ when $\marg{t}$ is more dispersed than $\marg{T}$, where
\begin{equation*}
    \begin{aligned}
    U_{T,t} &:= \wass_2^2(\marg{T}_{n}, \marg{t}_{n}) - \frac{1}{|\mathcal{S}|}\sum_{S \in \mathcal{S}}\wass_2^2(\marg{T}_{n}, \marg{S}_{n}),\\
    L_{T,t} &:= \bigg[ \wass_2(\marg{T}_{n}, \marg{t}_{n}) - \frac{1}{|\mathcal{S}|}\sum_{S \in \mathcal{S}} \wass_2(\marg{T}_{n}, \marg{S}_{n}) \bigg]_\pm^2 =: \sgnpow{2}{\bar L_{T,t}}.
    \end{aligned}
\end{equation*}
We quantify the uncertainty of $\{U_{T,t}, L_{T,t}\}$ as described in Appendix~\ref{app:uq-conv}. 

The estimators are valid when the MCMC algorithm has reached stationarity by time $S$ and has thereafter mixed \emph{at least once} by time $T$. In practice, we trace $\wass_2^2(\marg{T}_{n}, \marg{t}_{n})$ from $t=0$ until one integrated autocorrelation time before $T$, then choose $\mathcal{S}$ as the interval of stationarity of this trace.

If the trace does not become stationary, we increase $T.$ Pilot runs with small $n$ can help speed up the search for a large enough $T$. Another failure mode is when the trace increases towards stationarity, indicating that time-marginals are underdispersed, and that the sample splitting method of Section~\ref{sec:convergence} should be used instead.

\subsection{$p$-Wasserstein lagged coupling bound} \label{app:convergence-lagged-coupling}

We extend the coupling-based bound of \cite{biswas2019estimating} to general Wasserstein distances of arbitrary orders $p \ge 1$, as in equation~\eqref{eqn:k-primal}.

Suppose that we wish to estimate the convergence of a Markov chain with kernel $P$ and initialization $\marg{0}$ towards the stationary distribution $\marg{0}P^\infty.$
We consider a construction based on a joint Markov kernel $\tilde P((x,y), \cdot)$ with marginals $(P(x, \cdot), P(y, \cdot))$ and a lag parameter $\ell \in \mathbb{N}$: we sample a coupled pair of Markov chains $(\Xbar{t}, \X{t})_{t \ge 0}$ evolving under $\tilde P$ that is initialized at $(\Xbar{0},\X{0}) \in \Gamma (\marg{0}P^{\ell}, \marg{0}).$
Then, by the triangle and coupling inequalities, we obtain the bound
\begin{equation*}
    \wass_p(\marg{0}P^\infty, \marg{t}) 
    \le \sum_{j \ge 0} \wass_p(\marg{0}P^{t + (j+1)\ell}, \marg{0}P^{t + j\ell}) 
    \le \sum_{j \ge 0} \mathbb{E}\left[ c(\Xbar{t + j\ell}, \X{t + j\ell})^p\right]^{1/p}.
\end{equation*}

We estimate this bound by sampling i.i.d. replicates of the $\ell$-lag coupling construction, replacing expectations by empirical averages. To ensure that the estimator can be computed in finite time, an elegant solution is to design the joint Markov kernel $\tilde P$ such that the chains coalesce in finite time, see \cite{biswas2019estimating, jacob2020unbiased} for coalescive coupling strategies.

The method is appealing, as it only requires keeping track of one-dimensional summary statistics. The bound is informative when sufficiently contractive couplings $\tilde P$ can be devised. Choosing the lag $\ell$ large sharpens the bound by eliminating the inefficiency introduced by the triangle inequality, as demonstrated empirically in \cite{biswas2019estimating}.

\section{Numerical experiments} \label{app:numerics}

\subsection{Benchmark of assignment problem solvers} \label{app:bench}

\begin{figure}[ht]
    \centering
    \includegraphics[width=0.7\linewidth]{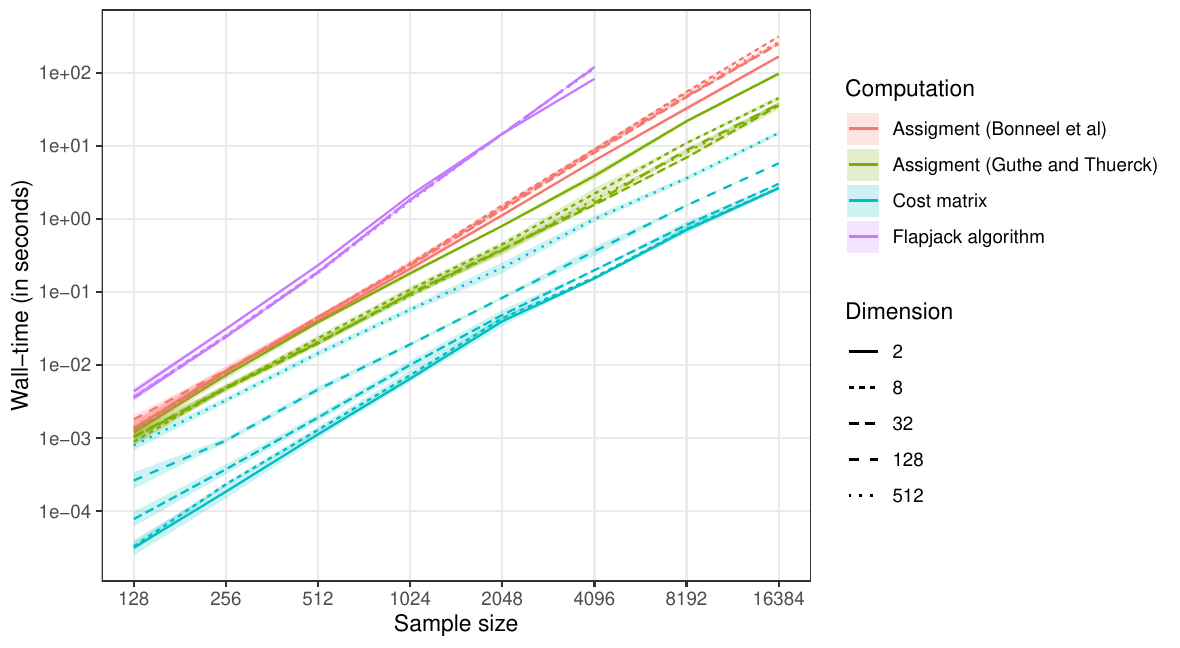}
    \caption{Benchmark of single-core assignment problem solvers. We solved for $\wass_2^2(\mu_n,\nu_n)$ with $\mu = \mathcal{N}_d(0_d,I_d)$ and $\nu = \mathcal{N}_d(0_d,4I_d)$ in various dimensions $d$ and at various sample sizes $n$. For each dimension, empirical means and standard deviations based on 8 replicates are shown.}
    \label{fig:bench}
\end{figure}

Figure~\ref{fig:bench} compares the assignment problem solvers of \cite{bonneel2011displacement} and \cite{guthe2021toms1015}, and contrasts them against the time spent computing the cost matrix using the linear algebra library Eigen \citep{cpp-eigen}. We see that both methods scale closer to $O(n^2)$ in practice, and that the method of \cite{guthe2021toms1015} outperforms that of \cite{bonneel2011displacement} and allows for problems to be solved at sample size $n=1000$ in around 0.1 seconds, and at $n=10000$ in 10 seconds.

Figure~\ref{fig:bench} also shows the wall-time of the Flapjack algorithm (Appendix~\ref{app:flapjack}). We see that this scales as $O(n^3)$, but that at sample size $n=1000$ it only takes around 2 seconds.

\subsection{Quality of approximate inference methods} \label{app:bias-experiments}

\subsubsection{Asymptotic bias of unadjusted MCMC algorithms} \label{app:bias-langevin}

For the plug-in estimators $\{U,L\}$ we used a sample size of $n = 1024,$ based on independent samples for simplicity, and we obtained empirical means and standard deviations from 256 replicates. For the coupling bound, we used $(B,I) = (1000, 2000)$ and $K = 10$.

\subsubsection{Tall data} \label{app:bias-talldata}

\paragraph{Model.} We considered the logistic regression model with likelihood
\begin{equation*}
y_i \mid x_i, \beta \sim \text{Bern}(F(x_i^\top \beta))\enskip \text{independently for observations } i \in [n],
\end{equation*}
where $x_i \in \mathbb{R}^{d}$ and $F(z) = 1/(1+ e^{-z}).$ We approximately followed the guidelines of \cite{gelman2008weakly}, centering the covariates and scaling them to scale 0.5, adding an intercept, and imposing the prior $\beta \sim \mathcal{N}_{d}(0_d, 25 I_d)$. \cite{chopin2017leave} lists the posterior log-density, score and Hessian.

\paragraph{MCMC.} The MCMC algorithms were preconditioned using the inverse-Hessian at the target mode $\beta^*$, and used the proposal covariance $d^{-1/3}[\nabla^2\log \pi(\beta^*)]^{-1},$ resulting in an $\approx 90\%$ acceptance rate for the MALA kernels. We initialized the MCMC algorithms at the mode $\beta^*$ and discarded $B = 100$ iterations as burn-in.

\paragraph{Estimators.} The parameters used in the main text can be found below. We also experimented with setting the thinning to $T=1$, and found that nearly identical point estimates $\{V,L\}$ were obtained.

\textbf{Pima dataset.} For the plug-in estimators $\{V,L\}$, we used $(K,I) = (16,100)$ with thinning $T=5$ for an overall sample size of $n = 1600$. For the coupling bound, we used $(K,I)=(32,500)$. We estimated that the coupling reduced the variance of $V$ by factors of roughly $(1.1,1.5,1.6,1.5)$ for (SGLD, SGLD-cv, Laplace, VI).

\textbf{DS1 dataset.} For the plug-in estimators $\{V,L\}$, we used $(K,I) = (16,200)$ with thinning $T=10$ for an overall sample size of $n = 3200$. For the coupling bound, we used $(K,I)=(32,2000)$. We estimated that the coupling reduced the variance of $V$ by factors of roughly $(1.0,2.2,1.6,1.2)$ for (SGLD, SGLD-cv, Laplace, VI).

\subsubsection{High-dimensional Bayesian linear regression} \label{app:bias-half-t}

The model and sampler are detailed in Appendix~\ref{app:mcmc-half-t}.

\paragraph{Model.} We set $a_0=b_0=1$. Since the model does not have an intercept, we centered the covariates and responses.

\paragraph{MCMC.} We set $\sigma_{\text{MH}}=0.8$. We initialized the MCMC algorithms from the prior and we discarded $B = 1000$ iterations as burn-in. 

\paragraph{Estimators.} For the plug-in estimators $\{U,L\}$, we used $(K,I)=(100,100)$ for an overall sample size of $n=10000$, with thinning $T=50$. For the coupling bound, we used $(K,I)=(100,5000)$. We estimated that the coupling reduced the variance of $U$ by factors of roughly $\{22, 18, 7.0, 3.4, 1.7\}$ in order of increasing $\varepsilon \in \{0.0003,0.001,0.003,0.01,0.03\}$.

\subsection{Convergence of MCMC algorithms} \label{app:convergence-experiments}

\subsubsection{Additional investigations} \label{app:mcmc-overdisp-start}

\begin{figure}[ht]
     \centering
    \includegraphics[width=\textwidth]{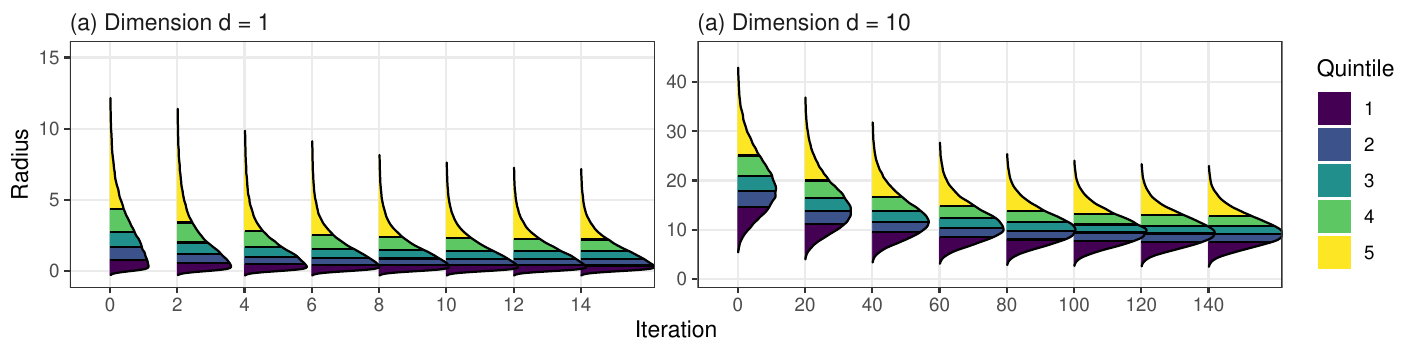}
    \caption{Density plots for the radial component of $\marg{t}$ of a RWM algorithm targeting a multivariate logistic target in various dimensions. See Appendix~\ref{app:mcmc-overdisp-start} for details.}
    \label{fig:cot-mcmc-logistic}
\end{figure}

\paragraph{Multivariate logistic target.} We consider a RWM algorithm with spherical Gaussian proposals with standard deviation $h$ targeting a multivariate logistic target with density $\marg{\infty}(x) \propto  e^{-\|x\|} / (1 + e^{-\|x\|})^2$. We initialize the sampler from $\marg{0} \cotr \marg{\infty}$ with density $\marg{0}(x) \propto \marg{\infty}(x/2).$

Our goal is to verify that overdispersion persists in the sense of $\cotr$. Since the target $\marg{\infty}$ and time-marginal $\marg{t}$ are spherically symmetric, by Proposition~\ref{prop:cot-cases}, we can verify $\marg{t} \cotr \marg{\infty}$ by checking the dispersion of their radial components.

\begin{figure}[ht]
    \centering
    \includegraphics[width=\textwidth]{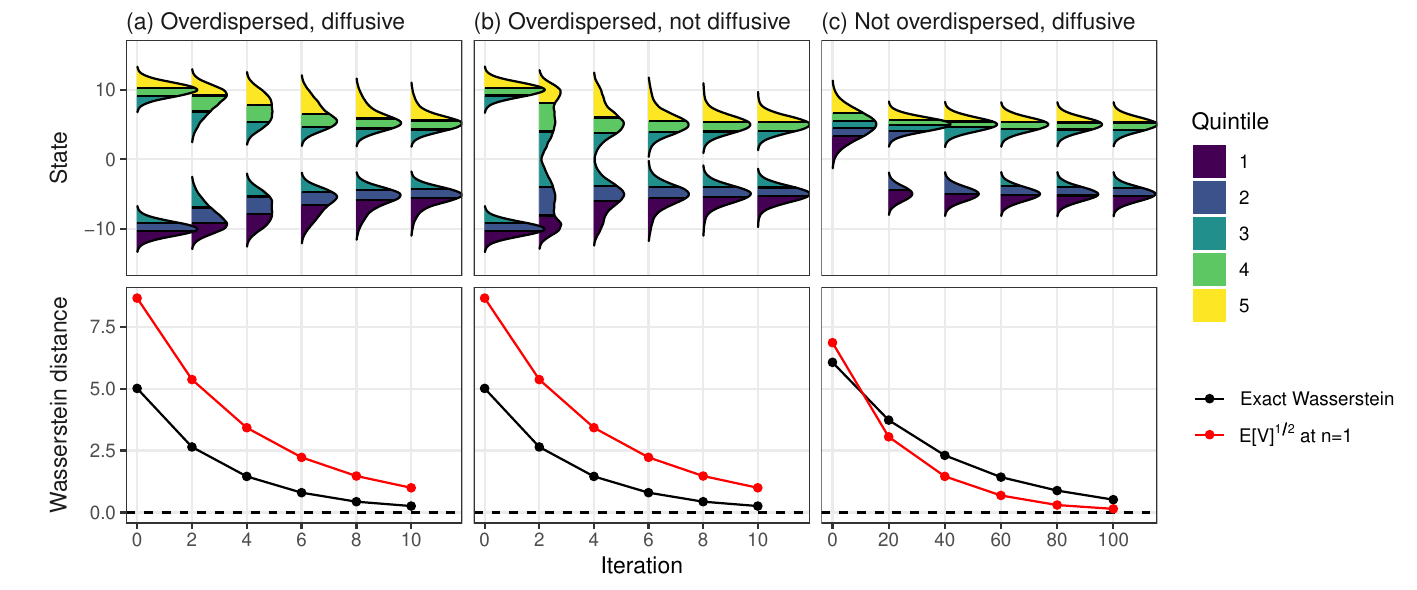}
    \caption{The effect of multimodality on the convergence of an MCMC algorithm. See Appendix~\ref{app:mcmc-overdisp-start} for details.}
    \label{fig:conv-multimodal}
\end{figure}

Figure~\ref{fig:cot-mcmc-logistic} displays the radial density of $\marg{t}$ against time $t$, where we considered dimensions, step sizes and acceptance rates of $(d,h, \alpha) \in \{(1,3,0.53), (10, 2.5, 0.24)\}$. Since the separation of any two pairs of quantiles gradually concentrates as $t \to \infty,$ we conclude that $\marg{t} \cotr \marg{\infty}$ is approximately satisfied for all $t \ge 0$.

\paragraph{Bimodal target.} 

We explore the target $\marg{\infty} = \frac{1}{2} \mathcal{N}(-5,1) + \frac{1}{2} \mathcal{N}(5,1)$ by RWM algorithms with Gaussian proposals with standard deviation $h$ and various initializations $\marg{0}$. We consider scenarios:
\begin{enumerate}[(a)]
    \item Step size $h=2$, overdispersed initialization $\marg{0} = \frac{1}{2} \mathcal{N}(-10,1) + \frac{1}{2} \mathcal{N}(10,1)$.
    \item Step size $h = 6$, overdispersed initialization .
    \item Step size $h = 4$, initialization $\marg{0} = \mathcal{N}(5,2)$ located in one of the modes.
\end{enumerate}

Figure~\ref{fig:conv-multimodal} displays marginal density plots and compares $\mathbb{E}[V]^{1/2}$ at sample size one to the true Wasserstein $\wass_2^2(\marg{\infty}, \marg{t})$. In settings (a) and (b), the marginals are overdispersed with respect to the target and the estimator $V$ is conservative. In setting (c), the marginals are not overdispersed with respect to the target and the estimator $V$ is not conservative, however $V$ is still able to distinguish the marginals from the target.

\subsubsection{Gaussian Gibbs sampler}

\paragraph{Model.} The Gaussian target has precision matrix $\Omega \in \mathbb{R}^{d \times d}$ whose only non-zero entries are $\Omega_{ii} = 1+\rho^2$ and $\Omega_{i,i\pm1} = -\rho$ for $i \in [d],$ where we identify the indices $(0,d+1)$ as $(d,1)$. 

\paragraph{Estimators.} Plug-in estimators $\{U,L\}$ were computed using the method of Appendix~\ref{app:convergence-time-averaging}, based on $n = 1024$ chains, $S \in [2000, 5000]$ and with a thinning factor of $5$. As samples from the target $\marg{\infty}$ could be drawn, we set $T = \infty$ for simplicity.

\subsubsection{Mixing time of Langevin algorithms} \label{app:conv-langevin}

\paragraph{Model.} The model and MCMC parameters are as in Appendix~\ref{app:bias-langevin}.

\paragraph{Estimators.} Plug-in estimators $\{U,L\}$ were computed using the method of Appendix~\ref{app:convergence-time-averaging}, based on $n = 1024$ chains and $S \in [300, 1000]$. As samples from the target $\marg{\infty}$ could be drawn, we set $T = \infty$ for simplicity.

We estimated the exact mixing time under the assumption that $\marg{t}$ is Gaussian for all $t \ge 0.$ This is true for ULA and OBABO, whereas for MALA and the Horowitz method this results in a very slight underestimate of the exact mixing time.

\subsubsection{Stochastic volatility model} \label{app:svm}

\paragraph{Model.} We considered the stochastic volatility model
\begin{equation*}
\begin{alignedat}{2}
    x_1 &\sim \mathcal{N}\left(0, \sigma^2/(1 - \varphi^2)\right),\\
    x_{t+1} \mid x_{t} &\sim \mathcal{N}(\varphi x_{t}, \sigma^2), \quad &&\forall t \in[d-1],\\
    y_t \mid x_t &\sim \mathcal{N}\left(0, \beta^2 \exp(x_t) \right), \quad &&\forall t \in[d].
\end{alignedat}
\end{equation*}
We fixed $(\beta, \sigma, \varphi) = (0.65, 0.15, 0.98)$ and simulated the data $y_{1:d}$ from the model. \citet[Section~9.6.2]{liu2001monte} lists the posterior log-density and score.

\paragraph{RWM.} Plug-in estimators $\{U,L\}$ were computed using the method of Appendix~\ref{app:convergence-time-averaging}, based on $n = 1024$ chains, $T = 1.5 \times 10^6$, $S \in [5 \times 10^5, 1.25 \times10^6]$ and thinning every $500$ iterations.

\paragraph{MALA.} Plug-in estimators $\{U,L\}$ were computed were computed using the method of Appendix~\ref{app:convergence-time-averaging}, based on $n = 1024$ chains, $T = 3 \times 10^4$, $S \in [10^4, 2.5 \times10^4]$ and thinning every $15$ iterations.

\paragraph{Fisher-MALA.} Plug-in estimators $\{U,L\}$ were computed were computed using the method of Appendix~\ref{app:convergence-time-averaging}, based on $n = 1024$ chains, $T = 1.25 \times 10^4$, $S \in [7.5 \times 10^3, 9 \times 10^3]$ and thinning every $5$ iterations.

The covariance structure of Fisher-MALA was adapted using the default recursion of \cite{titsias2023optimal}, diminishing the adaptation at the rate $t^{-1}$ with the iteration $t.$ The global scale parameter $h_t^2$ was updated with adaptation diminishing at a rate $t^{-2/3}$ after 1000 iterations, using the recursion $h_{t+1}^2 = h_t^2 + \ell (\alpha_t - \alpha^*) \cdot \min{(1, 100 t^{-2/3})}$ based on the current acceptance probability $\alpha_t$, the target acceptance probability $\alpha^* = 0.574$ \citep{roberts1998optimal}, and the default learning rate $\ell = 0.015$ of \cite{titsias2023optimal}.

\begin{figure}
    \centering
    \includegraphics[width=\linewidth]{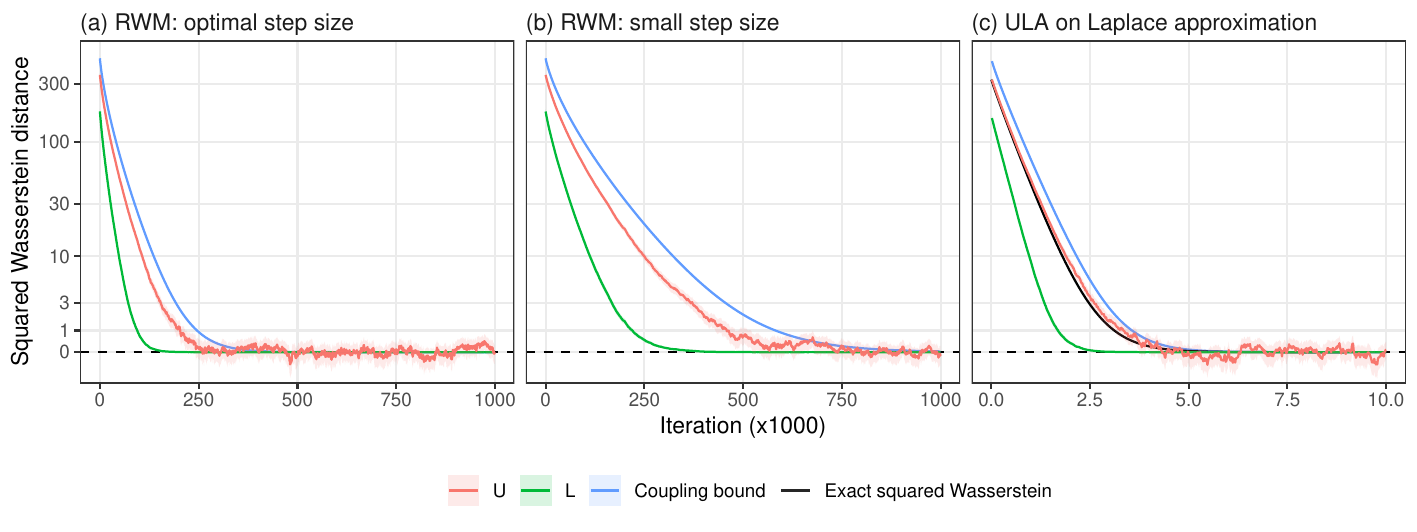}
    \caption{Additional experiments with samplers targeting the stochastic volatility model or its Laplace approximation. See Appendix~\ref{app:svm} for details.}
    \label{fig:svm-extra}
\end{figure}

\paragraph{Additional experiments.} Figure~\ref{fig:svm-extra} displays the results of additional experiments. We repeated the RWM experiments in the main text, replacing the coupling with the contractive GCRN coupling of \cite{papp2024scalable}, finding that the coupling bound became effective but that the proposed estimator $U$ was even sharper. We also considered an ULA targeting a Laplace approximation to the SVM (same parameters as MALA; we used a CRN coupling), finding that $U$ was remarkably close to the exact squared Wasserstein distance.

%% file: main.bbl
\begin{thebibliography}{86}
\providecommand{\natexlab}[1]{#1}
\providecommand{\url}[1]{\texttt{#1}}
\expandafter\ifx\csname urlstyle\endcsname\relax
  \providecommand{\doi}[1]{doi: #1}\else
  \providecommand{\doi}{doi: \begingroup \urlstyle{rm}\Url}\fi

\bibitem[Andrieu and Thoms(2008)]{andrieu2008tutorial}
C.~Andrieu and J.~Thoms.
\newblock {A tutorial on adaptive MCMC}.
\newblock \emph{Statistics and Computing}, 18\penalty0 (4):\penalty0 343--373,
  2008.

\bibitem[Baker et~al.(2019)Baker, Fearnhead, Fox, and Nemeth]{baker2019control}
J.~Baker, P.~Fearnhead, E.~B. Fox, and C.~Nemeth.
\newblock {Control variates for stochastic gradient MCMC}.
\newblock \emph{Statistics and Computing}, 29\penalty0 (3):\penalty0 599--615,
  2019.

\bibitem[Bardenet et~al.(2017)Bardenet, Doucet, and Holmes]{bardenet2017tall}
R.~Bardenet, A.~Doucet, and C.~Holmes.
\newblock {On Markov chain Monte Carlo methods for tall data}.
\newblock \emph{Journal of Machine Learning Research}, 18\penalty0
  (47):\penalty0 1--43, 2017.

\bibitem[Besag(1994)]{besag1994mala}
J.~Besag.
\newblock {Comments on ``Representations of knowledge in complex systems" by U.
  Grenander and M.I. Miller}.
\newblock \emph{Journal of the Royal Statistical Society: Series B},
  56\penalty0 (4):\penalty0 549--581, 1994.

\bibitem[Bhattacharya et~al.(2024)Bhattacharya, Linero, and
  Oates]{bhattacharya2024grand}
A.~Bhattacharya, A.~Linero, and C.~J. Oates.
\newblock {Grand Challenges in Bayesian Computation}.
\newblock \emph{Bulletin of the International Society for Bayesian Analysis
  (ISBA)}, 31\penalty0 (3), 2024.

\bibitem[Biswas and Mackey(2024)]{biswas2024bounding}
N.~Biswas and L.~Mackey.
\newblock {Bounding Wasserstein Distance with Couplings}.
\newblock \emph{Journal of the American Statistical Association}, 119\penalty0
  (548):\penalty0 2947--2958, 2024.

\bibitem[Biswas et~al.(2019)Biswas, Jacob, and Vanetti]{biswas2019estimating}
N.~Biswas, P.~E. Jacob, and P.~Vanetti.
\newblock {Estimating convergence of Markov chains with \mbox{L-lag}
  couplings}.
\newblock In \emph{{Advances in Neural Information Processing Systems}},
  volume~32, pages 7391--7401, 2019.

\bibitem[Biswas et~al.(2022)Biswas, Bhattacharya, Jacob, and
  Johndrow]{biswas2021coupling-based}
N.~Biswas, A.~Bhattacharya, P.~E. Jacob, and J.~E. Johndrow.
\newblock {Coupling-based convergence assessment of some Gibbs samplers for
  high-dimensional Bayesian regression with shrinkage priors}.
\newblock \emph{Journal of the Royal Statistical Society: Series B (Statistical
  Methodology)}, 84\penalty0 (3):\penalty0 973--996, 2022.

\bibitem[Blei et~al.(2017)Blei, Kucukelbir, and McAuliffe]{blei2017variational}
D.~M. Blei, A.~Kucukelbir, and J.~D. McAuliffe.
\newblock {Variational Inference: A Review for Statisticians}.
\newblock \emph{Journal of the American Statistical Association}, 112\penalty0
  (518):\penalty0 859--877, 2017.

\bibitem[Bobkov and Ledoux(2019)]{bobkov2019one-dimensional}
S.~Bobkov and M.~Ledoux.
\newblock One-dimensional empirical measures, order statistics, and
  {K}antorovich transport distances.
\newblock \emph{Memoirs of the American Mathematical Society}, 261\penalty0
  (1259), 2019.

\bibitem[Bonneel et~al.(2011)Bonneel, van~de Panne, Paris, and
  Heidrich]{bonneel2011displacement}
N.~Bonneel, M.~van~de Panne, S.~Paris, and W.~Heidrich.
\newblock {Displacement Interpolation Using {L}agrangian Mass Transport}.
\newblock \emph{ACM Transactions on Graphics}, 30\penalty0 (6):\penalty0 1--12,
  2011.

\bibitem[Bou-Rabee and Vanden-Eijnden(2010)]{bou-rabee2010pathwise}
N.~Bou-Rabee and E.~Vanden-Eijnden.
\newblock {Pathwise Accuracy and Ergodicity of Metropolized Integrators for
  SDEs}.
\newblock \emph{Communications on Pure and Applied Mathematics: A Journal
  Issued by the Courant Institute of Mathematical Sciences}, 63\penalty0
  (5):\penalty0 655--696, 2010.

\bibitem[Brenier(1991)]{brenier1991polar}
Y.~Brenier.
\newblock {Polar factorization and monotone rearrangement of vector-valued
  functions}.
\newblock \emph{Communications on Pure and Applied Mathematics}, 44\penalty0
  (4):\penalty0 375--417, 1991.

\bibitem[B{\"u}hlmann et~al.(2014)B{\"u}hlmann, Kalisch, and
  Meier]{buhlmann2014high}
P.~B{\"u}hlmann, M.~Kalisch, and L.~Meier.
\newblock {High-dimensional statistics with a view toward applications in
  biology}.
\newblock \emph{Annual Review of Statistics and Its Application}, 1\penalty0
  (1):\penalty0 255--278, 2014.

\bibitem[Carpenter et~al.(2017)Carpenter, Gelman, Hoffman, Lee, Goodrich,
  Betancourt, Brubaker, Guo, Li, and Riddell]{carpenter2017stan}
B.~Carpenter, A.~Gelman, M.~D. Hoffman, D.~Lee, B.~Goodrich, M.~Betancourt,
  M.~Brubaker, J.~Guo, P.~Li, and A.~Riddell.
\newblock {Stan: A Probabilistic Programming Language}.
\newblock \emph{Journal of Statistical Software}, 76\penalty0 (1):\penalty0
  1--32, 2017.

\bibitem[Carvalho et~al.(2010)Carvalho, Polson, and
  Scott]{carvalho2010horseshoe}
C.~M. Carvalho, N.~G. Polson, and J.~G. Scott.
\newblock {The horseshoe estimator for sparse signals}.
\newblock \emph{Biometrika}, 97\penalty0 (2):\penalty0 465--480, 2010.

\bibitem[Charlier et~al.(2021)Charlier, Feydy, Glaunès, Collin, and
  Durif]{charlier2021kernel}
B.~Charlier, J.~Feydy, J.~A. Glaunès, F.-D. Collin, and G.~Durif.
\newblock {Kernel Operations on the GPU, with Autodiff, without Memory
  Overflows}.
\newblock \emph{Journal of Machine Learning Research}, 22\penalty0
  (74):\penalty0 1--6, 2021.

\bibitem[Chizat et~al.(2020)Chizat, Roussillon, L\'eger, Vialard, and
  Peyr\'e]{chizat2020faster}
L.~Chizat, P.~Roussillon, F.~L\'eger, F.-X. Vialard, and G.~Peyr\'e.
\newblock {Faster Wasserstein distance estimation with the Sinkhorn
  divergence}.
\newblock In \emph{{Advances in Neural Information Processing Systems}},
  volume~33, 2020.

\bibitem[Chopin and Papaspiliopoulos(2020)]{chopin2020introduction}
N.~Chopin and O.~Papaspiliopoulos.
\newblock \emph{{An introduction to sequential Monte Carlo}}.
\newblock Springer Cham, 2020.

\bibitem[Christensen et~al.(2005)Christensen, Roberts, and
  Rosenthal]{christensen2005scaling}
O.~F. Christensen, G.~O. Roberts, and J.~S. Rosenthal.
\newblock {Scaling limits for the transient phase of local Metropolis--Hastings
  algorithms}.
\newblock \emph{Journal of the Royal Statistical Society Series B: Statistical
  Methodology}, 67\penalty0 (2):\penalty0 253--268, 2005.

\bibitem[Cuturi(2013)]{cuturi2013sinkhorn}
M.~Cuturi.
\newblock {Sinkhorn Distances: Lightspeed Computation of Optimal Transport}.
\newblock In \emph{{Advances in Neural Information Processing Systems}},
  volume~26, 2013.

\bibitem[Deb et~al.(2021)Deb, Ghosal, and Sen]{deb2021rates}
N.~Deb, P.~Ghosal, and B.~Sen.
\newblock {Rates of Estimation of Optimal Transport Maps using Plug-in
  Estimators via Barycentric Projections}.
\newblock In \emph{{Advances in Neural Information Processing Systems}}, pages
  29736--29753, 2021.

\bibitem[Dehaene and Barthelm{\'e}(2018)]{dehaene2018expectation}
G.~Dehaene and S.~Barthelm{\'e}.
\newblock {Expectation propagation in the large data limit}.
\newblock \emph{Journal of the Royal Statistical Society Series B: Statistical
  Methodology}, 80\penalty0 (1):\penalty0 199--217, 2018.

\bibitem[del Barrio and Loubes(2019)]{delbarrio2019central}
E.~del Barrio and J.-M. Loubes.
\newblock {Central limit theorems for empirical transportation cost in general
  dimension}.
\newblock \emph{Annals of Probability}, 47\penalty0 (2):\penalty0 926--951,
  2019.

\bibitem[del Barrio et~al.(2024)del Barrio, Gonz{\'a}lez-Sanz, and
  Loubes]{delbarrio2024central}
E.~del Barrio, A.~Gonz{\'a}lez-Sanz, and J.-M. Loubes.
\newblock {Central limit theorems for general transportation costs}.
\newblock \emph{Annales de l'Institut Henri Poincaré, Probabilités et
  Statistiques}, 60\penalty0 (2):\penalty0 847--873, 2024.

\bibitem[Dobson et~al.(2021)Dobson, Li, and Zhai]{dobson2021using}
M.~Dobson, Y.~Li, and J.~Zhai.
\newblock {Using coupling methods to estimate sample quality of stochastic
  differential equations}.
\newblock \emph{SIAM/ASA Journal on Uncertainty Quantification}, 9\penalty0
  (1):\penalty0 135--162, 2021.

\bibitem[Durmus and Moulines(2019)]{durmus2019high}
A.~Durmus and E.~Moulines.
\newblock {High-dimensional Bayesian inference via the unadjusted Langevin
  algorithm}.
\newblock \emph{Bernoulli}, 25\penalty0 (4A):\penalty0 2854--2882, 2019.

\bibitem[Dvurechensky et~al.(2018)Dvurechensky, Gasnikov, and
  Kroshnin]{dvurechensky2018computational}
P.~Dvurechensky, A.~Gasnikov, and A.~Kroshnin.
\newblock {Computational Optimal Transport: Complexity by Accelerated Gradient
  Descent Is Better Than by Sinkhorn’s Algorithm}.
\newblock In \emph{{Proceedings of the 35th International Conference on Machine
  Learning}}, volume~80, pages 1367--1376, 2018.

\bibitem[Efron and Stein(1981)]{efron1981jackknife}
B.~Efron and C.~Stein.
\newblock {The Jackknife Estimate of Variance}.
\newblock \emph{The Annals of Statistics}, 9\penalty0 (3):\penalty0 586--596,
  1981.

\bibitem[Fearnhead et~al.(2018)Fearnhead, Bierkens, Pollock, and
  Roberts]{fearnhead2018piecewise}
P.~Fearnhead, J.~Bierkens, M.~Pollock, and G.~O. Roberts.
\newblock {Piecewise Deterministic Markov Processes for Continuous-Time Monte
  Carlo}.
\newblock \emph{Statistical Science}, 33\penalty0 (3):\penalty0 386--412, 2018.

\bibitem[Fournier and Guillin(2015)]{fournier2015rate}
N.~Fournier and A.~Guillin.
\newblock {On the rate of convergence in Wasserstein distance of the empirical
  measure}.
\newblock \emph{Probability Theory and Related Fields}, 162:\penalty0 707--738,
  2015.

\bibitem[Gelbrich(1990)]{gelbrich1990formula}
M.~Gelbrich.
\newblock {On a formula for the {L2} {W}asserstein metric between measures on
  {E}uclidean and {H}ilbert spaces}.
\newblock \emph{Mathematische Nachrichten}, 147\penalty0 (1):\penalty0
  185--203, 1990.

\bibitem[Gelman and Rubin(1992)]{gelman1992inference}
A.~Gelman and D.~B. Rubin.
\newblock {Inference from Iterative Simulation Using Multiple Sequences}.
\newblock \emph{Statistical Science}, 7\penalty0 (4):\penalty0 457--472, 1992.

\bibitem[Gelman et~al.(2013)Gelman, Carlin, Stern, Dunson, Vehtari, and
  Rubin]{gelman2013bayesian}
A.~Gelman, J.~B. Carlin, H.~S. Stern, D.~B. Dunson, A.~Vehtari, and D.~B.
  Rubin.
\newblock \emph{{Bayesian Data Analysis}}.
\newblock CRC Press, 3 edition, 2013.

\bibitem[Genevay et~al.(2018)Genevay, Peyre, and Cuturi]{genevay2018learning}
A.~Genevay, G.~Peyre, and M.~Cuturi.
\newblock {Learning Generative Models with Sinkhorn Divergences}.
\newblock In \emph{{Proceedings of the Twenty-First International Conference on
  Artificial Intelligence and Statistics}}, volume~84, pages 1608--1617, 2018.

\bibitem[Giovagnoli and Wynn(1995)]{giovagnoli1995multivariate}
A.~Giovagnoli and H.~P. Wynn.
\newblock {Multivariate dispersion orderings}.
\newblock \emph{Statistics \& Probability Letters}, 22\penalty0 (4):\penalty0
  325--332, 1995.

\bibitem[Gorham and Mackey(2017)]{gorham2017measuring}
J.~Gorham and L.~Mackey.
\newblock {Measuring Sample Quality with Kernels}.
\newblock In \emph{{Proceedings of the 34th International Conference on Machine
  Learning}}, volume~70, pages 1292--1301. PMLR, 2017.

\bibitem[Guthe and Thuerck(2021)]{guthe2021toms1015}
S.~Guthe and D.~Thuerck.
\newblock {Algorithm 1015: A Fast Scalable Solver for the Dense Linear (Sum)
  Assignment Problem}.
\newblock \emph{ACM Trans. Math. Softw.}, 47\penalty0 (2), 2021.

\bibitem[Heng and Jacob(2019)]{heng2019unbiased}
J.~Heng and P.~E. Jacob.
\newblock {Unbiased Hamiltonian Monte Carlo with couplings}.
\newblock \emph{Biometrika}, 106\penalty0 (2):\penalty0 287--302, 2019.

\bibitem[Horowitz(1991)]{horowitz1991generalized}
A.~M. Horowitz.
\newblock {A generalized guided Monte Carlo algorithm}.
\newblock \emph{Physics Letters B}, 268\penalty0 (2):\penalty0 247--252, 1991.

\bibitem[Huggins et~al.(2020)Huggins, Kasprzak, Campbell, and
  Broderick]{huggins2020validated}
J.~Huggins, M.~Kasprzak, T.~Campbell, and T.~Broderick.
\newblock {Validated variational inference via practical posterior error
  bounds}.
\newblock In \emph{{International Conference on Artificial Intelligence and
  Statistics}}, pages 1792--1802. PMLR, 2020.

\bibitem[H\"utter and Rigollet(2021)]{hutter2021minimax}
J.-C. H\"utter and P.~Rigollet.
\newblock {Minimax estimation of smooth optimal transport maps}.
\newblock \emph{The Annals of Statistics}, 49\penalty0 (2):\penalty0
  1166--1194, 2021.

\bibitem[Jacob et~al.(2020)Jacob, O’Leary, and Atchadé]{jacob2020unbiased}
P.~E. Jacob, J.~O’Leary, and Y.~F. Atchadé.
\newblock {Unbiased Markov chain Monte Carlo methods with couplings}.
\newblock \emph{Journal of the Royal Statistical Society: Series B (Statistical
  Methodology)}, 82\penalty0 (3):\penalty0 543--600, 2020.

\bibitem[Johndrow et~al.(2020)Johndrow, Orenstein, and
  Bhattacharya]{johndrow2020scalable}
J.~Johndrow, P.~Orenstein, and A.~Bhattacharya.
\newblock {Scalable Approximate MCMC Algorithms for the Horseshoe Prior}.
\newblock \emph{Journal of Machine Learning Research}, 21\penalty0
  (73):\penalty0 1--61, 2020.

\bibitem[Johnson(1996)]{johnson1996studying}
V.~E. Johnson.
\newblock {Studying Convergence of Markov Chain Monte Carlo Algorithms Using
  Coupled Sample Paths}.
\newblock \emph{Journal of the American Statistical Association}, 91\penalty0
  (433):\penalty0 154--166, 1996.

\bibitem[Kassraie et~al.(2024)Kassraie, Pooladian, Klein, Thornton, Niles-Weed,
  and Cuturi]{kassraie2024progressive}
P.~Kassraie, A.-A. Pooladian, M.~Klein, J.~Thornton, J.~Niles-Weed, and
  M.~Cuturi.
\newblock {Progressive Entropic Optimal Transport Solvers}.
\newblock In \emph{{The Thirty-eighth Annual Conference on Neural Information
  Processing Systems}}, 2024.

\bibitem[Komarek and Moore(2003)]{komarek2003fast}
P.~Komarek and A.~W. Moore.
\newblock {Fast Robust Logistic Regression for Large Sparse Datasets with
  Binary Outputs}.
\newblock In \emph{{Proceedings of the Ninth International Workshop on
  Artificial Intelligence and Statistics}}, volume~R4, pages 163--170, 2003.

\bibitem[Kong(1992)]{kong1992note}
A.~Kong.
\newblock {A note on importance sampling using standardized weights}.
\newblock Technical report, University of Chicago, Deptartment of Statistics,
  1992.

\bibitem[Kucukelbir et~al.(2017)Kucukelbir, Tran, Ranganath, Gelman, and
  Blei]{kucukelbir2017automatic}
A.~Kucukelbir, D.~Tran, R.~Ranganath, A.~Gelman, and D.~M. Blei.
\newblock {Automatic Differentiation Variational Inference}.
\newblock \emph{Journal of Machine Learning Research}, 18\penalty0
  (14):\penalty0 1--45, 2017.

\bibitem[Liu(2001)]{liu2001monte}
J.~S. Liu.
\newblock \emph{{Monte Carlo Strategies in Scientific Computing}}.
\newblock Springer, 2001.

\bibitem[Ma et~al.(2015)Ma, Chen, and Fox]{ma2015complete}
Y.-A. Ma, T.~Chen, and E.~Fox.
\newblock {A complete recipe for stochastic gradient MCMC}.
\newblock \emph{Advances in neural information processing systems}, 28, 2015.

\bibitem[Manole et~al.(2024)Manole, Balakrishnan, Niles-Weed, and
  Wasserman]{manole2024plugin}
T.~Manole, S.~Balakrishnan, J.~Niles-Weed, and L.~Wasserman.
\newblock {Plugin estimation of smooth optimal transport maps}.
\newblock \emph{The Annals of Statistics}, 52\penalty0 (3):\penalty0 966--998,
  2024.

\bibitem[Margossian et~al.(2024)Margossian, Hoffman, Sountsov, Riou-Durand,
  Vehtari, and Gelman]{margossian2024nestedRhat}
C.~C. Margossian, M.~D. Hoffman, P.~Sountsov, L.~Riou-Durand, A.~Vehtari, and
  A.~Gelman.
\newblock {Nested {$\widehat{R}$}: Assessing the Convergence of Markov Chain
  Monte Carlo When Running Many Short Chains}.
\newblock \emph{Bayesian Analysis}, 2024.

\bibitem[Miller(1974)]{miller1974jackknife}
R.~G. Miller.
\newblock {The Jackknife--A Review}.
\newblock \emph{Biometrika}, 61\penalty0 (1):\penalty0 1--15, 1974.

\bibitem[Mills-Tettey et~al.(2007)Mills-Tettey, Stentz, and
  Dias]{mills-tettey2007dynamic}
G.~A. Mills-Tettey, A.~Stentz, and M.~B. Dias.
\newblock {The Dynamic Hungarian Algorithm for the Assignment Problem with
  Changing Costs}.
\newblock Technical Report CMU-RI-TR-07-27, Robotics Institute, Carnegie Mellon
  University, 2007.

\bibitem[Minka(2001)]{minka2001expectation}
T.~P. Minka.
\newblock {Expectation Propagation for approximate Bayesian inference}.
\newblock In \emph{{Proceedings of the 17th Conference in Uncertainty in
  Artificial Intelligence}}, pages 362--369, 2001.

\bibitem[Monmarch{\'e}(2021)]{monmarche2021high}
P.~Monmarch{\'e}.
\newblock {High-dimensional MCMC with a standard splitting scheme for the
  underdamped Langevin diffusion}.
\newblock \emph{Electronic Journal of Statistics}, 15\penalty0 (2):\penalty0
  4117--4166, 2021.

\bibitem[Nemeth and Fearnhead(2021)]{nemeth2021stochastic}
C.~Nemeth and P.~Fearnhead.
\newblock {Stochastic Gradient Markov Chain Monte Carlo}.
\newblock \emph{Journal of the American Statistical Association}, 116\penalty0
  (533):\penalty0 433--450, 2021.

\bibitem[Panaretos and Zemel(2019)]{panaretos2019statistical}
V.~M. Panaretos and Y.~Zemel.
\newblock {Statistical aspects of Wasserstein distances}.
\newblock \emph{Annual Review of Statistics and Its Application}, 6\penalty0
  (1):\penalty0 405--431, 2019.

\bibitem[Papp and Sherlock(2024)]{papp2024scalable}
T.~P. Papp and C.~Sherlock.
\newblock {Scalable couplings for the random walk Metropolis algorithm}.
\newblock \emph{Journal of the Royal Statistical Society Series B: Statistical
  Methodology}, 2024.

\bibitem[Paty et~al.(2020)Paty, d'Aspremont, and Cuturi]{paty2020regularity}
F.-P. Paty, A.~d'Aspremont, and M.~Cuturi.
\newblock {Regularity as regularization: smooth and strongly convex {B}renier
  potentials in optimal transport}.
\newblock In \emph{{Proceedings of the Twenty Third International Conference on
  Artificial Intelligence and Statistics}}, pages 1222--1232, 2020.

\bibitem[Peyré and Cuturi(2019)]{peyre2019computational}
G.~Peyré and M.~Cuturi.
\newblock {Computational Optimal Transport: With Applications to Data Science}.
\newblock \emph{Foundations and Trends® in Machine Learning}, 11\penalty0
  (5--6):\penalty0 355--607, 2019.

\bibitem[Prado et~al.(2024)Prado, Nemeth, and Sherlock]{prado2024mhss}
E.~Prado, C.~Nemeth, and C.~Sherlock.
\newblock {Metropolis--Hastings with Scalable Subsampling}.
\newblock \emph{arXiv preprint arXiv:2407.19602}, 2024.

\bibitem[{R Core Team}(2025)]{r}
{R Core Team}.
\newblock \emph{{R}: A Language and Environment for Statistical Computing}.
\newblock R Foundation for Statistical Computing, 2025.

\bibitem[Rippl et~al.(2016)Rippl, Munk, and Sturm]{rippl2016limit}
T.~Rippl, A.~Munk, and A.~Sturm.
\newblock {Limit laws of the empirical {W}asserstein distance: {G}aussian
  distributions}.
\newblock \emph{Journal of Multivariate Analysis}, 151:\penalty0 90--109, 2016.

\bibitem[Roberts and Rosenthal(1998)]{roberts1998optimal}
G.~O. Roberts and J.~S. Rosenthal.
\newblock {Optimal scaling of discrete approximations to Langevin diffusions}.
\newblock \emph{Journal of the Royal Statistical Society: Series B (Statistical
  Methodology)}, 60\penalty0 (1):\penalty0 255--268, 1998.

\bibitem[Roberts and Sahu(1997)]{roberts1997updating}
G.~O. Roberts and S.~K. Sahu.
\newblock {Updating Schemes, Correlation Structure, Blocking and
  Parameterization for the {G}ibbs Sampler}.
\newblock \emph{Journal of the Royal Statistical Society: Series B (Statistical
  Methodology)}, 59\penalty0 (2):\penalty0 291--317, 1997.

\bibitem[Roberts and Tweedie(1996)]{roberts1996exponential}
G.~O. Roberts and R.~L. Tweedie.
\newblock Exponential convergence of {L}angevin distributions and their
  discrete approximations.
\newblock \emph{Bernoulli}, 2\penalty0 (4):\penalty0 341--363, 1996.

\bibitem[Roberts et~al.(1997)Roberts, Gelman, and Gilks]{roberts1997weak}
G.~O. Roberts, A.~Gelman, and W.~R. Gilks.
\newblock {Weak convergence and optimal scaling of random walk Metropolis
  algorithms}.
\newblock \emph{The Annals of Applied Probability}, 7\penalty0 (1):\penalty0
  110--120, 1997.

\bibitem[Sejourne et~al.(2022)Sejourne, Vialard, and
  Peyr\'e]{sejourne2022faster}
T.~Sejourne, F.-X. Vialard, and G.~Peyr\'e.
\newblock {Faster Unbalanced Optimal Transport: Translation invariant Sinkhorn
  and 1-D Frank-Wolfe}.
\newblock In \emph{{Proceedings of The 25th International Conference on
  Artificial Intelligence and Statistics}}, volume 151, pages 4995--5021. PMLR,
  2022.

\bibitem[Shaked(1982)]{shaked1982dispersive}
M.~Shaked.
\newblock {Dispersive ordering of distributions}.
\newblock \emph{Journal of Applied Probability}, 19\penalty0 (2):\penalty0
  310--320, 1982.

\bibitem[Shaked and Shanthikumar(2007)]{shaked2007stochastic}
M.~Shaked and J.~G. Shanthikumar.
\newblock \emph{{Stochastic Orders}}.
\newblock Springer, 2007.

\bibitem[Sisson et~al.(2018)Sisson, Fan, and Beaumont]{sisson2018handbook}
S.~A. Sisson, Y.~Fan, and M.~Beaumont.
\newblock \emph{{Handbook of approximate Bayesian computation}}.
\newblock CRC press, 2018.

\bibitem[Sixta et~al.(2024)Sixta, Rosenthal, and Brown]{sixta2024bounding}
S.~Sixta, J.~S. Rosenthal, and A.~Brown.
\newblock {Bounding and estimating MCMC convergence rates using common random
  number simulations}.
\newblock \emph{arXiv preprint arXiv:2309.15735}, 2024.

\bibitem[Smith et~al.(1988)Smith, Everhart, Dickson, Knowler, and
  Johannes]{smith1988using}
J.~W. Smith, J.~E. Everhart, W.~Dickson, W.~C. Knowler, and R.~S. Johannes.
\newblock {Using the ADAP Learning Algorithm to Forecast the Onset of Diabetes
  Mellitus}.
\newblock In \emph{{Proceedings of the Annual Symposium on Computer Application
  in Medical Care}}, pages 261--265. American Medical Informatics Association,
  1988.

\bibitem[Staudt and Hundrieser(2024)]{staudt2024convergence}
T.~Staudt and S.~Hundrieser.
\newblock {Convergence of Empirical Optimal Transport in Unbounded Settings}.
\newblock \emph{Bernoulli}, advance publication, 2024.

\bibitem[Strassen(1965)]{strassen1965existence}
V.~Strassen.
\newblock {The Existence of Probability Measures with Given Marginals}.
\newblock \emph{The Annals of Mathematical Statistics}, 36\penalty0
  (2):\penalty0 423--439, 1965.

\bibitem[Tierney(1994)]{tierney1994markov}
L.~Tierney.
\newblock {Markov Chains for Exploring Posterior Distributions}.
\newblock \emph{The Annals of Statistics}, 22\penalty0 (4):\penalty0
  1701--1728, 1994.

\bibitem[Titsias(2023)]{titsias2023optimal}
M.~Titsias.
\newblock {Optimal Preconditioning and Fisher Adaptive Langevin Sampling}.
\newblock In \emph{{Advances in Neural Information Processing Systems}},
  volume~36, pages 29449--29460, 2023.

\bibitem[Vats et~al.(2019)Vats, Flegal, and Jones]{vats2019multivariate}
D.~Vats, J.~M. Flegal, and G.~L. Jones.
\newblock {Multivariate output analysis for Markov chain Monte Carlo}.
\newblock \emph{Biometrika}, 106\penalty0 (2):\penalty0 321--337, 2019.

\bibitem[Villani(2003)]{villani2003topics}
C.~Villani.
\newblock \emph{{Topics in Optimal Transportation}}.
\newblock American Mathematical Society, 2003.

\bibitem[Villani(2009)]{villani2009optimal}
C.~Villani.
\newblock \emph{{Optimal Transport: Old and New}}.
\newblock Springer, 2009.

\bibitem[Wang and Titterington(2005)]{wang2005inadequacy}
B.~Wang and D.~M. Titterington.
\newblock {Inadequacy of interval estimates corresponding to variational
  Bayesian approximations}.
\newblock In \emph{{International workshop on artificial intelligence and
  statistics}}, pages 373--380, 2005.

\bibitem[Weed and Bach(2019)]{weed2019sharp}
J.~Weed and F.~Bach.
\newblock {Sharp asymptotic and finite-sample rates of convergence of empirical
  measures in Wasserstein distance}.
\newblock \emph{Bernoulli}, 25\penalty0 (4A):\penalty0 2620--2648, 2019.

\bibitem[Welling and Teh(2011)]{welling2011sgld}
M.~Welling and Y.~W. Teh.
\newblock {Bayesian Learning via Stochastic Gradient Langevin Dynamics}.
\newblock In \emph{{Proceedings of the 28th International Conference on Machine
  Learning (ICML-11)}}, pages 681--688, 2011.

\bibitem[Wu et~al.(2022)Wu, Schmidler, and Chen]{wu2022minimax}
K.~Wu, S.~Schmidler, and Y.~Chen.
\newblock {Minimax Mixing Time of the Metropolis-Adjusted Langevin Algorithm
  for Log-Concave Sampling}.
\newblock \emph{Journal of Machine Learning Research}, 23\penalty0
  (270):\penalty0 1--63, 2022.

\end{thebibliography}


\begin{thebibliography}{39}
\providecommand{\natexlab}[1]{#1}
\providecommand{\url}[1]{\texttt{#1}}
\expandafter\ifx\csname urlstyle\endcsname\relax
  \providecommand{\doi}[1]{doi: #1}\else
  \providecommand{\doi}{doi: \begingroup \urlstyle{rm}\Url}\fi

\bibitem[Apers et~al.(2024)Apers, Gribling, and
  Szil{{\'a}}gyi]{apers2024hamiltonian}
S.~Apers, S.~Gribling, and D.~Szil{{\'a}}gyi.
\newblock {Hamiltonian Monte Carlo for efficient Gaussian sampling: long and
  random steps}.
\newblock \emph{Journal of Machine Learning Research}, 25\penalty0
  (348):\penalty0 1--30, 2024.

\bibitem[Bhattacharya et~al.(2016)Bhattacharya, Chakraborty, and
  Mallick]{bhattacharya2016fast}
A.~Bhattacharya, A.~Chakraborty, and B.~K. Mallick.
\newblock {Fast sampling with Gaussian scale mixture priors in high-dimensional
  regression}.
\newblock \emph{Biometrika}, 103\penalty0 (4):\penalty0 985--991, 2016.

\bibitem[Biswas and Mackey(2024)]{biswas2024bounding}
N.~Biswas and L.~Mackey.
\newblock {Bounding Wasserstein Distance with Couplings}.
\newblock \emph{Journal of the American Statistical Association}, 119\penalty0
  (548):\penalty0 2947--2958, 2024.

\bibitem[Biswas et~al.(2019)Biswas, Jacob, and Vanetti]{biswas2019estimating}
N.~Biswas, P.~E. Jacob, and P.~Vanetti.
\newblock {Estimating convergence of Markov chains with \mbox{L-lag}
  couplings}.
\newblock In \emph{{Advances in Neural Information Processing Systems}},
  volume~32, pages 7391--7401, 2019.

\bibitem[Biswas et~al.(2022)Biswas, Bhattacharya, Jacob, and
  Johndrow]{biswas2021coupling-based}
N.~Biswas, A.~Bhattacharya, P.~E. Jacob, and J.~E. Johndrow.
\newblock {Coupling-based convergence assessment of some Gibbs samplers for
  high-dimensional Bayesian regression with shrinkage priors}.
\newblock \emph{Journal of the Royal Statistical Society: Series B (Statistical
  Methodology)}, 84\penalty0 (3):\penalty0 973--996, 2022.

\bibitem[Bobkov and Ledoux(2019)]{bobkov2019one-dimensional}
S.~Bobkov and M.~Ledoux.
\newblock One-dimensional empirical measures, order statistics, and
  {K}antorovich transport distances.
\newblock \emph{Memoirs of the American Mathematical Society}, 261\penalty0
  (1259), 2019.

\bibitem[Boissard and Le~Gouic(2014)]{boissard2014on}
E.~Boissard and T.~Le~Gouic.
\newblock {On the mean speed of convergence of empirical and occupation
  measures in {W}asserstein distance}.
\newblock \emph{Annales de l'Institut Henri Poincar\'{e}, Probabilit\'{e}s et
  Statistiques}, 50\penalty0 (2):\penalty0 539--563, 2014.

\bibitem[Bolley and Villani(2005)]{bolley2005weighted}
F.~Bolley and C.~Villani.
\newblock {Weighted {Csisz\'ar-Kullback-Pinsker} inequalities and applications
  to transportation inequalities}.
\newblock \emph{Annales de la Facult\'e des Sciences de Toulouse}, 14\penalty0
  (3):\penalty0 331--352, 2005.

\bibitem[Bonneel et~al.(2011)Bonneel, van~de Panne, Paris, and
  Heidrich]{bonneel2011displacement}
N.~Bonneel, M.~van~de Panne, S.~Paris, and W.~Heidrich.
\newblock {Displacement Interpolation Using {L}agrangian Mass Transport}.
\newblock \emph{ACM Transactions on Graphics}, 30\penalty0 (6):\penalty0 1--12,
  2011.

\bibitem[Boucheron et~al.(2013)Boucheron, Lugosi, and
  Massart]{boucheron2013concentration}
S.~Boucheron, G.~Lugosi, and P.~Massart.
\newblock \emph{{Concentration Inequalities: A Nonasymptotic Theory of
  Independence}}.
\newblock Oxford University Press, 2013.

\bibitem[Chewi and Pooladian(2023)]{chewi-pooladian2023entropic}
S.~Chewi and A.-A. Pooladian.
\newblock {An entropic generalization of {Caffarelli's} contraction theorem via
  covariance inequalities}.
\newblock \emph{Comptes Rendus. Math\'ematique}, 361:\penalty0 1471--1482,
  2023.

\bibitem[Chizat et~al.(2020)Chizat, Roussillon, L\'eger, Vialard, and
  Peyr\'e]{chizat2020faster}
L.~Chizat, P.~Roussillon, F.~L\'eger, F.-X. Vialard, and G.~Peyr\'e.
\newblock {Faster Wasserstein distance estimation with the Sinkhorn
  divergence}.
\newblock In \emph{{Advances in Neural Information Processing Systems}},
  volume~33, 2020.

\bibitem[Chopin and Ridgway(2017)]{chopin2017leave}
N.~Chopin and J.~Ridgway.
\newblock {Leave Pima Indians Alone: Binary Regression as a Benchmark for
  Bayesian Computation}.
\newblock \emph{Statistical Science}, 32\penalty0 (1):\penalty0 64--87, 2017.

\bibitem[del Barrio et~al.(2024)del Barrio, Gonz{\'a}lez-Sanz, and
  Loubes]{delbarrio2024central}
E.~del Barrio, A.~Gonz{\'a}lez-Sanz, and J.-M. Loubes.
\newblock {Central limit theorems for general transportation costs}.
\newblock \emph{Annales de l'Institut Henri Poincaré, Probabilités et
  Statistiques}, 60\penalty0 (2):\penalty0 847--873, 2024.

\bibitem[Efron and Stein(1981)]{efron1981jackknife}
B.~Efron and C.~Stein.
\newblock {The Jackknife Estimate of Variance}.
\newblock \emph{The Annals of Statistics}, 9\penalty0 (3):\penalty0 586--596,
  1981.

\bibitem[Gelman et~al.(2008)Gelman, Jakulin, Pittau, and Su]{gelman2008weakly}
A.~Gelman, A.~Jakulin, M.~G. Pittau, and Y.-S. Su.
\newblock {A weakly informative default prior distribution for logistic and
  other regression models}.
\newblock \emph{The Annals of Applied Statistics}, 2\penalty0 (4):\penalty0
  1360--1383, 2008.

\bibitem[Gozlan and L\'eonard(2007)]{gozlan2007large}
N.~Gozlan and C.~L\'eonard.
\newblock {A large deviation approach to some transportation cost
  inequalities}.
\newblock \emph{Probability Theory and Related Fields}, 139:\penalty0 235--283,
  2007.

\bibitem[Guennebaud et~al.(2010)Guennebaud, Jacob, et~al.]{cpp-eigen}
G.~Guennebaud, B.~Jacob, et~al.
\newblock {Eigen v3}.
\newblock http://eigen.tuxfamily.org, 2010.

\bibitem[Guthe and Thuerck(2021)]{guthe2021toms1015}
S.~Guthe and D.~Thuerck.
\newblock {Algorithm 1015: A Fast Scalable Solver for the Dense Linear (Sum)
  Assignment Problem}.
\newblock \emph{ACM Trans. Math. Softw.}, 47\penalty0 (2), 2021.

\bibitem[Jacob et~al.(2020)Jacob, O’Leary, and Atchadé]{jacob2020unbiased}
P.~E. Jacob, J.~O’Leary, and Y.~F. Atchadé.
\newblock {Unbiased Markov chain Monte Carlo methods with couplings}.
\newblock \emph{Journal of the Royal Statistical Society: Series B (Statistical
  Methodology)}, 82\penalty0 (3):\penalty0 543--600, 2020.

\bibitem[Jonker and Volgenant(1987)]{jonker1987shortest}
R.~Jonker and A.~Volgenant.
\newblock {A shortest augmenting path algorithm for dense and sparse linear
  assignment problems}.
\newblock \emph{Computing}, 38\penalty0 (4):\penalty0 325--340, 1987.

\bibitem[Kuhn(1955)]{kuhn1955hungarian}
H.~W. Kuhn.
\newblock {The Hungarian method for the assignment problem}.
\newblock \emph{Naval Research Logistics Quarterly}, 2\penalty0
  (1--2):\penalty0 83--97, 1955.

\bibitem[Lawson and Lim(2001)]{lawson2001geometric}
J.~D. Lawson and Y.~Lim.
\newblock {The Geometric Mean, Matrices, Metrics, and More}.
\newblock \emph{The American Mathematical Monthly}, 108\penalty0 (9):\penalty0
  797--812, 2001.

\bibitem[Liu(2001)]{liu2001monte}
J.~S. Liu.
\newblock \emph{{Monte Carlo Strategies in Scientific Computing}}.
\newblock Springer, 2001.

\bibitem[McCann(1995)]{mccann1995existence}
R.~J. McCann.
\newblock {Existence and uniqueness of monotone measure-preserving maps}.
\newblock \emph{Duke Mathematical Journal}, 80\penalty0 (2):\penalty0 309--323,
  1995.

\bibitem[McDiarmid(1989)]{mcdiarmid1989method}
C.~McDiarmid.
\newblock {On the method of bounded differences}.
\newblock In J.~Siemons, editor, \emph{{Surveys in Combinatorics, 1989: Invited
  Papers at the Twelfth British Combinatorial Conference}}, pages 148--188.
  Cambridge University Press, 1989.

\bibitem[Mills-Tettey et~al.(2007)Mills-Tettey, Stentz, and
  Dias]{mills-tettey2007dynamic}
G.~A. Mills-Tettey, A.~Stentz, and M.~B. Dias.
\newblock {The Dynamic Hungarian Algorithm for the Assignment Problem with
  Changing Costs}.
\newblock Technical Report CMU-RI-TR-07-27, Robotics Institute, Carnegie Mellon
  University, 2007.

\bibitem[Munkres(1957)]{munkres1957algorithms}
J.~Munkres.
\newblock {Algorithms for the Assignment and Transportation Problems}.
\newblock \emph{Journal of the Society for Industrial and Applied Mathematics},
  5\penalty0 (1):\penalty0 32--38, 1957.

\bibitem[Nesterov(2004)]{nesterov2004introductory}
Y.~Nesterov.
\newblock \emph{{Introductory Lectures on Convex Optimization}}.
\newblock Springer, 2004.

\bibitem[Panaretos and Zemel(2019)]{panaretos2019statistical}
V.~M. Panaretos and Y.~Zemel.
\newblock {Statistical aspects of Wasserstein distances}.
\newblock \emph{Annual Review of Statistics and Its Application}, 6\penalty0
  (1):\penalty0 405--431, 2019.

\bibitem[Papp and Sherlock(2024)]{papp2024scalable}
T.~P. Papp and C.~Sherlock.
\newblock {Scalable couplings for the random walk Metropolis algorithm}.
\newblock \emph{Journal of the Royal Statistical Society Series B: Statistical
  Methodology}, 2024.

\bibitem[Peyré and Cuturi(2019)]{peyre2019computational}
G.~Peyré and M.~Cuturi.
\newblock {Computational Optimal Transport: With Applications to Data Science}.
\newblock \emph{Foundations and Trends® in Machine Learning}, 11\penalty0
  (5--6):\penalty0 355--607, 2019.

\bibitem[Roberts and Rosenthal(1998)]{roberts1998optimal}
G.~O. Roberts and J.~S. Rosenthal.
\newblock {Optimal scaling of discrete approximations to Langevin diffusions}.
\newblock \emph{Journal of the Royal Statistical Society: Series B (Statistical
  Methodology)}, 60\penalty0 (1):\penalty0 255--268, 1998.

\bibitem[Solomon et~al.(2022)Solomon, Greenewald, and
  Nagaraja]{solomon2022k-variance}
J.~Solomon, K.~Greenewald, and H.~Nagaraja.
\newblock {$k$}-variance: A clustered notion of variance.
\newblock \emph{SIAM Journal on Mathematics of Data Science}, 4\penalty0
  (3):\penalty0 957--978, 2022.

\bibitem[Strassen(1965)]{strassen1965existence}
V.~Strassen.
\newblock {The Existence of Probability Measures with Given Marginals}.
\newblock \emph{The Annals of Mathematical Statistics}, 36\penalty0
  (2):\penalty0 423--439, 1965.

\bibitem[Titsias(2023)]{titsias2023optimal}
M.~Titsias.
\newblock {Optimal Preconditioning and Fisher Adaptive Langevin Sampling}.
\newblock In \emph{{Advances in Neural Information Processing Systems}},
  volume~36, pages 29449--29460, 2023.

\bibitem[Tjøstheim(1990)]{tjostheim1990nonlinear}
D.~Tjøstheim.
\newblock Non-linear time series and markov chains.
\newblock \emph{Advances in Applied Probability}, 22\penalty0 (3):\penalty0
  587--611, 1990.

\bibitem[Weed and Bach(2019)]{weed2019sharp}
J.~Weed and F.~Bach.
\newblock {Sharp asymptotic and finite-sample rates of convergence of empirical
  measures in Wasserstein distance}.
\newblock \emph{Bernoulli}, 25\penalty0 (4A):\penalty0 2620--2648, 2019.

\bibitem[Zhou(2018)]{zhou2018fenchel}
X.~Zhou.
\newblock {On the Fenchel Duality between Strong Convexity and Lipschitz
  Continuous Gradient}.
\newblock \emph{arXiv preprint arXiv:1803.06573}, 2018.

\end{thebibliography}
